\documentclass[a4paper,11pt, oneside,titlepage, binder=0.5cm]{book}
\usepackage[T1]{fontenc}		
\usepackage[utf8]{inputenc}		
\usepackage[english]{babel}		
\usepackage{geometry}			
\usepackage{setspace}			
\usepackage{fancyhdr}			
\usepackage{afterpage} 			
\usepackage{hyperref}			
\usepackage{url}
\usepackage{color}			
\usepackage{xcolor}			
\usepackage{enumerate}			
\usepackage{enumitem}			
\usepackage{graphicx}			
\usepackage{epstopdf}			
\usepackage{pstool}			
\usepackage{psfrag}
\usepackage{subfig}		
\usepackage{array}			
\usepackage{tabularx}			
\usepackage[bottom]{footmisc}		
\usepackage{booktabs}			
\usepackage{longtable}			
\usepackage{caption}			
\usepackage{float}			
\usepackage{subfloat}			
\usepackage{rotating}			
\usepackage{rotfloat}			
\usepackage{amsmath} 			
\usepackage{amssymb}			
\usepackage{cancel}                     
\usepackage{mleftright}
\usepackage{listings}                   
\usepackage[swapnames,norules,nouppercase]{frontespizio}
\usepackage[intoc]{nomencl}
\usepackage[acronym,toc,nomain,nopostdot,nonumberlist]{glossaries}
\usepackage[round]{natbib}
\newcommand{\bianca}{\renewcommand\NAT@open{[}\renewcommand\NAT@close{]}}
\makeatother

\setcitestyle{citesep={,}}
\usepackage{verbatim}                   
\usepackage{wrapfig}                 
\usepackage{blindtext}
\renewcommand{\chaptermark}[1]{\markright{\chaptername\ \thechapter.\ #1}{}}    
\renewcommand{\sectionmark}[1]{\markright{\sectionname\ \thesection.\ #1}}      
\renewcommand{\sectionmark}[1]{\markright{\thesection.\ #1}}
\fancypagestyle{plain}{
                        \fancyhf{}                              
                        \cfoot{\thepage}                 
                        }                       
\hypersetup{colorlinks=true, linkcolor=black, citecolor=black, filecolor=black, urlcolor=black}
\definecolor{sapienza}{RGB}{130,36,51} 
\definecolor{cust1}{RGB}{85,85,85}
\definecolor{cust2}{RGB}{212,212,212}
\makenomenclature 
\makeglossaries 

\usepackage{calligra}
\usepackage{fancyhdr}

\usepackage{geometry}
\usepackage{color,soul}
\usepackage{mathtools}
\usepackage{graphicx}
\usepackage{rotating}
\usepackage{pdflscape}
\usepackage{amssymb}
\usepackage{amsfonts}
\usepackage{mathrsfs}
\usepackage{caption}
\usepackage{amsmath}
\usepackage{lscape}
\usepackage{booktabs}
\usepackage{xcolor} 
\usepackage{float}
\usepackage{adjustbox}
\usepackage{subcaption}
\usepackage{bm}
\usepackage{lipsum}
\usepackage{amsmath, amssymb, amsthm, nccmath}
\usepackage{algpseudocode}
\usepackage[linesnumbered,ruled,vlined]{algorithm2e}
\usepackage{subfig}
\usepackage{makecell}
\usepackage{dirtytalk}
\usepackage{xcolor,colortbl}
\usepackage{multicol}
\usepackage{multirow}
\usepackage{array}
\usepackage{float}
\usepackage{makecell}
\usepackage{tabularx}
\restylefloat{table}
\usepackage{subcaption}
\usepackage{afterpage}
\usepackage{array}
\usepackage{csquotes}
\usepackage{bm}
\usepackage{xspace}
\usepackage{amsfonts}
\usepackage{tikz}
\usepackage{hhline}
\usetikzlibrary{trees}
\usetikzlibrary{calc}
\usetikzlibrary{arrows}
\usepackage{pgfplots}
\usepackage{gensymb}
\usetikzlibrary{shapes}
\usetikzlibrary{shapes.geometric}
\usetikzlibrary{arrows,decorations.pathmorphing,backgrounds,positioning,fit}
\usetikzlibrary{arrows,shapes,positioning,shadows,trees}
\usetikzlibrary{shapes,shadows,trees}
\usepackage{makeidx}  
\usepackage{graphicx}        
\usepackage{multicol}        
\usepackage[bottom]{footmisc}
\usepackage{url}

\usepackage{tabularx}
\usepackage[round]{natbib}
\setcitestyle{citesep={,}}
\usepackage{amsmath, amssymb, amsthm, nccmath}
\usepackage{algpseudocode}
\usepackage[linesnumbered,ruled,vlined]{algorithm2e}
\usepackage{setspace}
\usepackage{microtype}
\usepackage{booktabs}
\usepackage{longtable}
\usepackage{array}
\usepackage{multirow}
\usepackage{wrapfig}
\usepackage{float}
\usepackage{colortbl}
\usepackage{setspace}
\usepackage{pdflscape}
\usepackage{tabu}
\usepackage{adjustbox}
\usepackage{arydshln}
\usepackage{threeparttable}
\usepackage[thinlines]{easytable}
\usepackage[width=0.87\textwidth, font=small,labelfont=bf]{caption}
\usepackage{dcolumn,booktabs}
\newcolumntype{d}[1]{D{.}{.}{#1}}

\usepackage[normalem]{ulem}
\usepackage{makecell}
\usepackage{xcolor}
\usepackage{float}
\usepackage{rotating}
\usepackage{amssymb}
\newcommand{\bbeta}{\boldsymbol{\beta}}
\usepackage{hyperref}
\usepackage{multirow}
\usepackage{bm}
\usepackage{bbm}
\usepackage[ruled]{algorithm2e}
\pgfplotsset{compat=1.18}

\usepackage{bm}
\usepackage{bbm}
\usepackage{adjustbox}
\usepackage{arydshln}
\usepackage{threeparttable}
\usepackage{dcolumn,booktabs}
\newcolumntype{d}[1]{D{.}{.}{#1}}

\begin{document}

\def\papertitle{On Quantile Regression Forests for Modelling Mixed-Frequency and Longitudinal Data}

\begin{titlepage}
\centering
\includegraphics[width=0.33\textwidth]{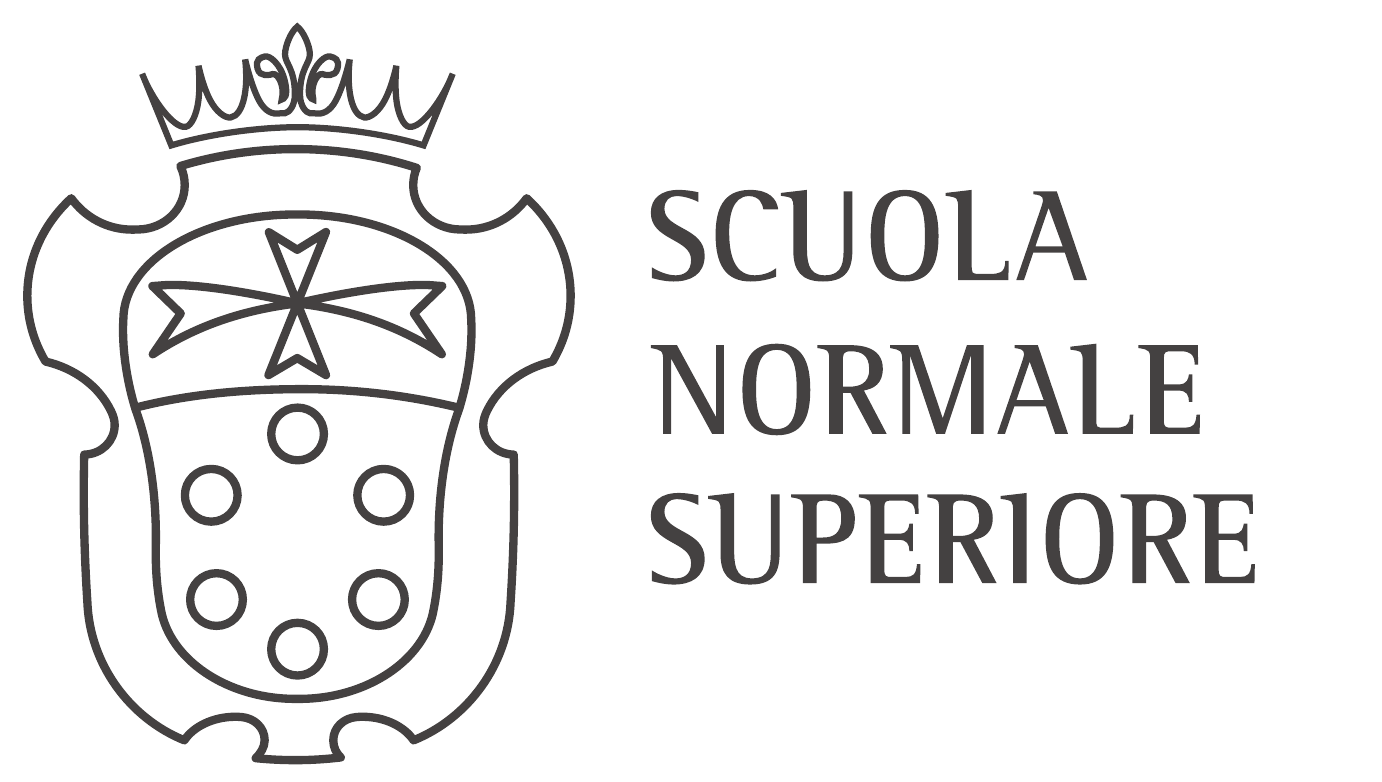}

\vspace{1cm}

{ Class of Sciences \par}
\vspace{0.3cm}

{ \large Ph.D. in Data Science \par}
\vspace{0.3cm}
{ 36\degree cycle \par}

\vspace{2cm}

{\bfseries\boldmath\Large
\begin{center}
    \papertitle
\end{center}
} 

\vspace{2cm}

{
Scientific Disciplinary Area SECS-S/01}

\vspace{2.3cm}
\hspace{0.5cm}
\begin{minipage}[t]{0.53\textwidth}
    Candidate\par \vspace{0.3cm}
  Mila Andreani\par
\end{minipage}
\begin{minipage}[t]{0.39\textwidth}
    Supervisors \par
    \vspace{0.3cm}
     Prof. Francesca Chiaromonte \par
    Prof. Lea Petrella \par
    Prof. Nicola Salvati \par
\end{minipage}

\vspace{3cm}
{
Academic Year 2023-2024\par}
\end{titlepage}

\setstretch{1.1}
\pagenumbering{Roman}











\newpage

\thispagestyle{fancy}			
\setlength{\parskip}{0pt plus 3.0pt}
\section*{Abstract}
The aim of this thesis is to extend the applications of the Quantile Regression Forest (QRF) algorithm to handle mixed-frequency and longitudinal data. To this end, standard statistical approaches have been exploited to build two novel algorithms: the Mixed- Frequency Quantile Regression Forest (MIDAS-QRF) and the Finite Mixture Quantile Regression Forest (FM-QRF).
\vspace{0.15in}

\noindent The MIDAS-QRF combines the flexibility of QRF with the Mixed Data Sampling (MIDAS) approach, enabling non-parametric quantile estimation with variables observed at different frequencies. FM-QRF, on the other hand, extends random effects machine learning algorithms to a QR framework, allowing for conditional quantile estimation in a longitudinal data setting.
The contributions of this dissertation lie both methodologically and empirically. 

\vspace{0.15in}

\noindent Methodologically, the MIDAS-QRF and the FM-QRF represent two novel approaches for handling mixed-frequency and longitudinal data in QR machine learning framework. Empirically, the application of the proposed models in financial risk management and climate-change impact evaluation demonstrates their validity as accurate and flexible models to be applied in complex empirical settings.


\begingroup
\let\cleardoublepage\clearpage
\tableofcontents
\endgroup


 
\begingroup
\let\cleardoublepage\clearpage
\listoffigures
\endgroup


\begingroup
\let\cleardoublepage\clearpage
\listoftables
\endgroup



\pagenumbering{arabic}

\chapter{Introduction and Overview}
\thispagestyle{plain}
Quantile Regression (QR) has been introduced in \cite{koenker1978regression} as a powerful technique to model the entire conditional distribution of a response variable given a set of covariates. This approach is particularly useful when the standard regression models fail at correctly modeling the relationship between the response and the covariates, or when the gaussianity assumption on the outcome's distribution is untanable. In such scenarios, QR allows to obtain more robust and reliable results by modeling location parameters beyond the conditional mean, and a variety of fields, such as economics, finance, healthcare, and environmental science \citep{koenker2005quantile, koenker2017handbook}, have reaped the benefits of this approach. 
\vspace{0.15in}

\noindent The recent development of non-parametric QR models have further extended the applications of QR, including the machine learning realm. In this context, QR machine learning algorithms represent one of the main advancements in overcoming the limits of the parametric formulation of standard QR models. As a matter of fact, machine learning algorithms do not require any a-priori assumption on the functional form of the relationship between the outcome and the covariates, resulting more flexible and reliable than standard QR in empirical applications involving unknown and highly complex relationships among variables.

\vspace{0.15in}

\noindent Few examples of QR machine learning algorithms are the QR neural network model of \cite{white1992nonparametric}, QR Support Vector Machines \citep{hwang2005simple, xu2015weighted} and QR Random Forests \citep{meinshausen2006quantile, athey2019generalized}.
\vspace{0.15in}

\noindent Despite the benefits of these algorithms, their application is constrained to standard experimental designs, as they often falter in non-standard empirical settings, such as those involving mixed-frequency or longitudinal data.
\vspace{0.15in}

\noindent The former case is particularly relevant in time series analysis, in which information is often available at different temporal resolutions. Standard statistical and econometrics models, including QR, usually require the use of variables observed at the same frequency, causing potentially useful predictors to be excluded due to the temporal mismatch. One of the main contributions to address this issue is the Mixed Data Sampling (MIDAS) approach proposed in \cite{ghysels2007midas}. This model allows to include variables observed at different frequencies, allowing to expand the research methodology beyond the standard statistical setting. 

\vspace{0.15in}

\noindent Given its innovativity, the MIDAS approach has been extended to QR and applied in a variety of fields, such as finance and economics \citep{kuzin2009midas, andreani2021multivariate, candila2023mixed}, environmental sciences \citep{oloko2022climate, jiang2023carbon} and tourism \citep{bangwayo2015can, wen2021forecasting}.

\vspace{0.15in}

\noindent In the longitudinal data setting instead, previous contributions to the QR literature have exploited mixed-effects or random effects models to account for the potential association between dependent observations \citep{farcomeni2012quantile, smith2015multilevel, alfo2017finite, marino2018mixed, merlo2021forecasting,merlo2022quantile,merlo2022quantilets}. Although these models incorporate an individual-specific random intercept to model unobserved heterogeneity, their parametric formulation may not be suitable in various empirical applications, leading to potentially inaccurate inferences.

\vspace{0.15in}

\noindent Non-parametric approaches have been proposed, but in the machine learning realm, the algorithms that have been extended to handle mixed-frequency data \citep{xu2019artificial} and longitudinal data \citep{xiong2019mixed, luts2012mixed, hajjem2014mixed, hajjem2011mixed, sela2012re}  lack the capability to estimate conditional quantiles, as they have been developed only in a standard regression setting.

\vspace{0.15in}

\noindent Thus, the aim of this dissertation is to bridge this gap in the literature by introducing two novel machine learning algorithms that allow to estimate conditional quantiles in the mixed-frequency and longitudinal data frameworks.

\vspace{0.15in}

\noindent Both proposed models build upon the Quantile Regression Forest (QRF) algorithm, ensuring a high level of accuracy and computational efficiency. The choice of the QRF algorithm stems from its inherent flexibility, allowing for easy comparison with standard econometric models in terms of statistical adequacy, accuracy, interpretability, and computational effort. Empirical results presented in this thesis demonstrate that the proposed extensions outperform commonly used econometric and statistical models in the QR literature.

\vspace{0.15in}

\noindent The main contributions of this dissertation are twofold. Methodologically, it introduces two algorithms: the Mixed-Frequency Quantile Regression Forest (MIDAS-QRF, detailed in Chapter \ref{ch:MIDAS-QRF}) and the Finite Mixture Quantile Regression Forest (FM-QRF, presented in Chapter \ref{ch:FM-QRF}). The MIDAS-QRF is based on a novel methodology that merges the MIDAS approach and the QRF algorithm, enabling non-parametric estimation of quantiles using data observed at different frequencies. The FM-QRF, on the other hand, builds upon random effects machine learning algorithms and it is based on leaving the random effects distribution unspecified and on estimating the fixed part of the model with QRF. Quantile estimates are obtained using an iterative procedure based on the Expectation Maximization-type algorithm with the Asymmetric Laplace distribution as the working likelihood. This methodology extends the work of \cite{geraci2007quantile, geraci2014linear, alfo2017finite} to a non-linear and non-parametric framework and adapts the mixed-modeling approach presented in \cite{hajjem2014mixed} to a QR framework.

\vspace{0.15in}

\noindent The validity of the proposed models has been tested empirically in the financial and economics settings. As a matter of fact, the recent emerging risks concerning financial crises and climate change have required financial institutions, policymakers, and researchers to develop novel methodological approaches to capture complex relationships and manage such risks.

\vspace{0.15in}

\noindent The two innovative methodological approaches result particularly useful in this settings, where non-Gaussian characteristics and mixed-frequency or longitudinal data are common. The MIDAS-QRF is empirically applied in a financial risk management setting for computing the well-known financial risk measure Value-at-Risk. Empirical findings demonstrate that the MIDAS-QRF delivers statistically adequate forecasts that outperform popular models in terms of accuracy (refer to Section \ref{sec:MIDAS-QRF-empirical}).
\vspace{0.15in}

\noindent The FM-QRF is applied in a climate-change impact evaluation setting to predict the Growth-at-Risk (GaR) of GDP growth for 210 countries. Climate-related variables are used as covariates, revealing heterogeneous effects of unsustainable climate practices among countries. 
\vspace{0.15in}

\noindent In order to test the flexibility of the proposed FM-QRF, the model is applied on an additional longitudinal dataset concerning the effects of the COVID-19 pandemic on children's mental health to extend previous findings based on standard linear models (refer to Section \ref{sec:FM-QRF-empirical}).



\chapter{Mixed-Frequency Quantile Regression Forests} \label{ch:MIDAS-QRF}
\chaptermark{Mixed-Frequency QRF}
\section{Introduction}

Standard regression models infer the effects of a set of covariates on the conditional expected value of the response variable. However, in real-world scenarios the effects of the covariates may vary across different parts of the outcome variable's conditional distribution. In this case, standard regression models may provide misleading results, and a more complete picture of the response variable's distribution would allow to obtain more robust results, especially if the distribution of the outcome exhibits non-Gaussian characteristics.

\vspace{0.15in}

\noindent To this end, a variety of models has been developed to estimate location parameters beyond the expected value, with Quantile Regression (QR) being one of the most important ones. Introduced in \cite{koenker1978regression} as a generalization of median regression, QR represents a flexible methodology to model data that violate the gaussianity assumptions of standard regression models. As a result, QR has become widely popular among scholars and practitioners in several fields, such as environmental science \citep{reich2012spatiotemporal, vasseur2021comparing, coronese2019evidence}, healthcare and medicine \citep{wei2006quantile,
merlo2022two, borgoni2018modelling}, finance and economics \citep{taylor1999quantile, merlo2022quantile, bernardi2018bayesian, petrella2019joint, daouia2018estimation, daouia2021expecthill}. 
\vspace{0.15in}

\noindent Although being widely applied in several empirical studies, the standard QR model may be affected by two main limitations. 
The first one arises when dealing with data collected at mixed-frequencies, as in time series modelling. In this domain, it is often necessary to incorporate information with different temporal resolution to uncover meaningful relations among the phenomena of interest. The application of standard statistical and econometrics models, including QR, usually requires the use of variables observed at the same frequency, causing potentially useful predictors to be excluded due to the temporal mismatch. 
One of the most relevant approaches developed to overcome this issue is the Mixed Data Sampling (MIDAS) model proposed by \cite{ghysels2007midas}. This approach includes mixed-frequency variables as covariates in linear models, allowing to obtain more accurate predictions. For this reason, it has been applied to a variety of fields, such as finance and economics \citep{kuzin2009midas, andreani2021multivariate, candila2023mixed}, environmental sciences \citep{oloko2022climate, jiang2023carbon} and tourism \citep{bangwayo2015can, wen2021forecasting}.
\vspace{0.15in}

\noindent Another possible limitation of the standard QR model is its reliance on the a-priori specification of the functional form of the relation between the outcome and the covariates. In many empirical applications, this relationship is often unknown and highly complex, and more robust results could be otbtained employing a non-parametric approach.
To address this issue, the standard QR approach has been extended to the machine learning realm: starting from \cite{white1992nonparametric}, which applies neural networks to QR, other contributions have incorporated QR in the most common machine learning algorithms, such as Support Vector Machines \citep{hwang2005simple, xu2015weighted} and Random Forests \citep{meinshausen2006quantile, athey2019generalized}. However, the majority of these models cannot handle mixed-frequency data. For this reason, recently \cite{xu2019artificial, xu2021qrnn} extended QR neural networks to the MIDAS framework, although they showed the presence of some drawbacks related to the low interpretability, high computational effort and high number of observations needed to train the model.\vspace{0.15in}

\noindent Thus, the aim of this chapter is to propose a comprehensive methodology to estimate quantiles that addresses the limitations of both standard QR and complex machine learning models. To this end, the MIDAS-QRF is introduced as a novel machine learning algorithm able to embed the MIDAS component into the Quantile Regression Forest algorithm (QRF) proposed by \cite{meinshausen2006quantile}. 
\vspace{0.15in}

\noindent Being based on the QRF algorithm, the proposed model offers several advantages with respect to deep learning algorithms. The MIDAS-QRF is easier to train and it also allows to extract the so called "variable importance" of each covariate,  giving insight into the relative importance of different covatiates in forecasting the response variable. These features make MIDAS-QRF particularly flexible and useful in a variety of domains and applications.
Moreover, the MIDAS components introduced in the QRF allow to model non-linear relationships among variables sampled at different frequencies without specifying a-priori any particular functional form and without making any assumption on the dependent variable's distribution. The proposed model also expands the applicability of the well known QRF, since it handles mixed-frequency data often involved in real empirical applications. Last but not least,
the proposed MIDAS-QRF model is particularly suitable for empirical applications involving variables with skewed and fat-tailed distributions.

\vspace{0.15in}

\noindent Moreover, the MIDAS-QRF offer advantages also over standard statistical methods, such as multivariate splines on the covariate space. This is due to the MIDAS-QRF (and Random Forest algorithms in general) nature as an ensemble method. By combining the predictions of multiple decision trees, the MIDAS-QRF allows to reduce overfitting and improve accuracy with respect to individual models. Moreover, Contributions to the literature \citep{genuer2012variance, breiman1996bagging, breiman2001random, hastie2009random} show that this approach based on aggregation also helps in reducing the variance of the model, leading to increased stability in forecasts. Moreover, algorithms based on Random Forests are particularly effective in handling high-dimensional data, thanks to the random selection of features at each split, which mitigates the curse of dimensionality and provides insights into feature importance. Additionally, they are flexible, scalable, and can naturally handle missing values \citep{breiman2001random}. 

\vspace{0.15in}

\noindent It is quite common to encounter dependent variables in the financial domain that are not normally distributed, and are further influenced by additional covariates observed at lower frequencies, often exhibiting non-linear relationships.
\vspace{0.15in}

\noindent For this reason, the MIDAS-QRF is employed to estimate the well-known financial risk measure Value-at-Risk (VaR), which is one of key risk metrics employed for capital calculation, decision-making, and risk management within the Basel III banking framework. \citep{Jorion:1997}. From a statistical point of view, VaR represents the conditional quantile of a financial variable's distribution, and a variety of models have been applied to its estimation, such as QR, linear ARCH models \citep{koenker1996conditional, taylor1999quantile}, GARCH models \citep{xiao2009conditional, lee2013quantile, zheng2018hybrid}, penalized QR \citep{bayer2018combining}, models based on the Asymmetric Laplace distribution \citep{merlo2021forecasting, taylor2019forecasting}, as well as extensions to multivariate settings \citep{petrella2019joint, bernardi2017multiple, merlo2022quantile}. Some of these models have been also extended to account for mixed-frequency data, see for example \cite{engle2013stock, candila2023mixed} and \cite{mo2018macroeconomic}.
\vspace{0.15in}

\noindent In empirical applications involving time series, it may be useful to estimate quantiles in a dynamic framework to model the time-varying distribution of the variable of interest. In the financial literature, the Conditional Autoregressive Value-at-Risk (CAViaR) model of \cite{engle2004caviar} has been proposed to accurately model he time-changing distribution of portfolio returns. This approach is based on directly estimating the conditional quantile via a linear autoregressive model, and it has been recently extended to the mixed-frequency framework by \cite{xu2021quantile}.
\vspace{0.15in}

\noindent In the spirit of the CaViaR, this chapter proposes also an extension of the MIDAS-QRF to a dynamic framework. The resulting Dynamic MIDAS-QRF allows to estimate quantiles in an autoregressive framework by introducing the lagged quantile predictions as additional covariate similarly to the CaViaR model.

\vspace{0.15in}

\noindent Given that financial data are usually observed at different frequencies and are often characterized by the well-known stylized facts \citep{cont2001empirical}, they are rarely well-fitted by linear models. In this context, the proposed MIDAS-QRF may be more appropriate than others to obtain accurate quantile estimates. 
\vspace{0.15in}

\noindent For this reason, the proposed model is applied in an empirical setting related to the financial field. In particular, this chapter focuses on the emerging topic of the financialization of energy commodities. This products have been widely employed over the two past decades as  hedging and speculative assets, especially during
periods of financial and economic downturns. This phenomenon, along with the deregulation of over-the-
counter markets, led to a significant increase of the volatility of energy commodities returns.
Thus, the study of the risks linked to such commodities is particularly relevant under a risk management framework.
The empirical application is focused on three daily energy commodities indexes, the West Texas Intermediate (WTI) Crude Oil, the Brent Crude Oil and the Heating Oil. The aim of the empirical application is to employ the proposed MIDAS-QRF and its dynamic version to predict VaR, defined as the maximum loss that a financial operator can incur over a defined time horizon for a given confidence level. The time series spans from July 2014 to March 2022, including observations collected during the COVID-19 pandemic and at the beginning of Russian-Ukrainian conflict. In order to measure the risks associated with energy commodities, three different low-frequency covariates are considered: the monthly Real Broad Dollar Index, whose effects on oil prices have been investigated in \cite{LIN201659, AKRAM2009838}, the quarterly Natural Gas returns and the quarterly Saudi Arabia Crude Oil Production, whose link to oil prices has been extensively studied by the U.S. Energy Information Administration (EIA) \footnote{https://www.eia.gov/finance/markets/crudeoil/spot\_prices.php}.

\vspace{0.15in}

\noindent The statistical validity of the out-of-sample VaR forecasts is tested by means of backtesting procedures \citep{christoffersen1998evaluating, christoffersen2004backtesting, engle2004caviar}. The empirical results show that the proposed models outperform several well-known statistical and machine learning models often considered in the literature. A variable importance analysis is also reported to show which variables may be considered the most relevant in predicting the VaR.
The rest of the chapter is organized as follows: Section \ref{sec:MIDAS-QRF-notation} gives preliminary information on QRF, Section \ref{sec:MIDAS-QRF-methodology} describes the proposed methodology, Section \ref{sec:MIDAS-QRF-empirical} presents the empirical application of MIDAS-QRF and Section \ref{sec:MIDAS-QRF-conclusions} concludes.


\section{Quantile Regression Forest: Notation and Preliminary Results}
\label{sec:MIDAS-QRF-notation}

This section reports the notation and some preliminars on QRF useful for the rest of the chapter. 
\vspace{0.15in}

 \noindent Let $\mathcal{S}=\{(Y_i, \mathbf{X}_i)\}_{i=1}^N$ be a random sample of dimension $N$ of random variables drawn from the unknown joint distribution of the random variables $(Y, \mathbf{X})\in \mathbb{R} \times  \mathbb{R}^p$ with realisation $\mathbf{s}=\{(y_i, \mathbf{x}_i)\}_{i=1}^N$, where $Y$ is the response variable and $\mathbf{X}$ the vector of $p$ covariates.
 \vspace{0.15in}

\noindent The QRF algorithm \citep{meinshausen2006quantile} has been developed as an extension to the Random Forests (RF) algorithm introduced by \cite{breiman2001random}. Both models rely on the Classification and Regression Trees (CART) \citep{breiman1984classification} algorithm. Differently from the the Random Forest approach, which estimates the conditional expected value, the QRF estimates the quantile of the conditional distribution of $Y$.
\vspace{0.15in}

\noindent Specifically, each decision tree in the QRF is trained with the CART algorithm, that consists in recursively partitioning the training sample $\mathcal{S}$ into $M$ disjoint sub-samples denoted with $R_m, m=1, \dots, M$ according to a splitting rule. In this setting, the splitting rule is based on minimizing the the Sum of Squared Errors (SSE) in each sub-sample of $R_m$:

\begin{equation}
SSE_{R_m}= \; \; 
\sum_{y_i \in R_m} (\bar{y}_m - y_i)^2,
\end{equation} 

\noindent where $\bar{y}_m$ is the mean of the observations in $R_m$ \footnote{Other splitting rules can be considered. For instance, \cite{athey2019generalized} proposes to use a quantile loss-base splitting rule. The choice of the splitting rule depends mainly on the empirical application of interest and the final forecast accuracy, as no contribution in the literature offers clear and robust evidence for choosing one approach over another in quantile regression settings.
In this thesis, the standard QRF approach of \cite{meinshausen2006quantile} has been chosen since it is more established in the machine learning literature. However, the proposed model can be easily adapted to consider other QRF models with different splitting rules.}. 
\vspace{0.15in}

\noindent At the end of the tree training, each sub-sample that is no further splitted is denoted as "terminal node" and indicated with $R^{*}_m$. 

\vspace{0.15in}

\noindent The QRF prediction of the conditional quantile $Q_{\tau}(Y|\mathbf{X}=x)$ is obtained from the estimated conditional distribution $F(y |\mathbf{X}=x)$, defined as follows:



\begin{equation}
	\label{eq:qrfconddistr}
	\begin{aligned}
		F(y|\mathbf{X}=x)
		&:=P(Y\leq y|\mathbf{X}=x )\\
		&:=\mathbb{E}[\mathbf{1}_{\{Y\leq y\}}|\mathbf{X}=x]
	\end{aligned}
\end{equation}

\noindent In particular, given a new set of observations, $\hat{F}(y|\mathbf{X}=x)$ is estimated by individuating one terminal node $R^{*}_m$ in each tree and by averaging the estimations of the $B$ trees:

\begin{equation}
		\hat{F}(y|\mathbf{X}=x )=\frac{1}{B}\sum_{b=1}^{B}\sum_{Y_i \in {R^{*}_{m,b}}} \frac{\mathbf{1}_{\{Y_{i}\leq y\}}}{|{R^{*}_{m,b}}|} ,
	\end{equation}

\noindent Subsequently, the quantile at probability level $\tau \in (0,1)$ is estimated as follows:

\begin{equation}
\hat{Q}_{\tau}(Y|\mathbf{X}=x) := \inf \, \{y : \widehat{F} \; (y|\mathbf{X}=x ) \leq \tau\}
\end{equation}

\noindent The QRF also computes the Variable Importance of each covariate, i.e. the influence of each covariate on the model's performance. The bigger the importance, the greater the positive effect of the variable on the model's accuracy. This feature improves the interpretability of the phenomena of interest and it can be considered as a nice property held by Random Forests with respect, for instance, to Neural Networks. 

\vspace{0.15in}

\noindent As mentioned in the introduction, the standard QRF algorithm cannot handle directly mixed-frequency data as many other models, and the aim of this chapter is to fill this gap by introducing mixed frequency variables in a QRF framework exploiting the MIDAS approach. 

\vspace{0.15in}

\noindent The methodology of the resulting MIDAS-QRF is presented in the next section.

\section{Methodology}
\label{sec:MIDAS-QRF-methodology}

This section describes the MIDAS-QRF methodological approach, developed to extend the QRF algorithm of \cite{meinshausen2006quantile} via the MIDAS approach of \cite{ghysels2007midas} in order to handle mixed-frequency data in quantiles estimation via Random Forests. 

\subsection{The MIDAS-QRF Model}
Let $Y_{i,t}$ and $\mathbf{X}_{i,t}$, $i=1, \dots, N_t, t=1, \dots, T$ be the high-frequency response variable and the set of high-frequency covariates observed at time $i$ of the $t-th$ period of the year. Moreover, $\textbf{Z}_t=\{Z_t^h\}^{H}_{h=1}$ is the vector of H low-frequency covariates observed in the $t-th$ period of the year. In this sense, $T$ represents the overall number of low-frequency periods. 

\vspace{0.15in}

\noindent For instance, if $Z_t$ is observed monthly, $T=12$ and the value $N_t$ represents the total number of days in the $t-th$ month. In a financial setting, the variable $Y_{i,t}$ may represent the daily financial returns sampled at day $i$ while the low-frequency variables might be monthly economic variables measuring the general state of the economy, sampled at the $t-th$ month of the year.
\vspace{0.15in}

\noindent The MIDAS approach proposed by \cite{ghysels2007midas} allows to include mixed-frequency variables in the QRF model, where the dependent variable is observed at a higher-frequency than the covariates.
\vspace{0.15in}

\noindent The simplest MIDAS linear regression is:

\begin{equation}
	\label{eq:MIDAS}
	Y_{i,t}=\beta_0+\beta_1MC_{i-1,t}^1+\dots+\beta_H MC_{i-1,t}^H+ \varepsilon_{i,t}
\end{equation}

where the covariate $MC_{i,t}^h=(\sum_{j=1}^{K}\phi_k(\bm{\omega})Z_{t-j}^h)_{i,t}$ is the MIDAS component, a filter of the last $K$ observations of $Z^h$ up to time $i-1$ of the $t$-th period. The number of lags $k$ can be chosen arbitrarily or via grid search. For interpretability purposes, in this chapter the number $k$ is chosen so that to consider meaningful fraction of the year (for instance, for monthly variables it would makes sense to consider three months, that is $k=3$).

The function $\phi_k(\boldsymbol{\omega})$ is the Beta weighting function (see \cite{candila2023mixed} and references therein) in which $\boldsymbol{\omega}=(\omega_1, \omega_2)$:

\begin{equation}
	\label{eq:beta}
	\phi_k(\omega_1, \omega_2)=\frac{(k/K)^{\omega_1-1}(1-k/K)^{\omega_2-1}}{\sum_{j=1}^{K}(j/K)^{\omega_1-1}(1-j/K)^{\omega_2-1}}.
\end{equation}

This function allows to impute a greater weight to more recent observations by setting $\omega_1=1$ and optimising $\omega_2$ with respect to the model's likelihood or a proper loss function. 
Other weighting functions are discussed in \cite{ghysels2019estimating}. 

\vspace{0.15in}

\noindent In this chapter, the linear specification of \eqref{eq:MIDAS} is relaxed by considering:
\begin{equation}
\label{eq:RF2}
Y_{i,t}=f(\mathbf{X}_{i-1,t}, \mathbf{MC}_{i-1,t})+\varepsilon_{i,t}
\end{equation}
where $f(\cdot)$ is a non-parametric function and $\mathbf{MC}_{i-1,t}=\{MC_{i-1,t}^h\}_{h=1}^H$  is the matrix of MIDAS components related to the set of covariates $\mathbf{Z}_t$.

\vspace{0.15in}

\noindent Being interested in the VaR of $Y_{i,t}$, the MIDAS-QRF estimates the conditional quantile:

\begin{equation}
\hat{Q}_{\tau}(Y_{i,t}|\mathbf{X}_{i-1,t}, \mathbf{MC}_{i-1,t})=f_{\tau}(\mathbf{X}_{i-1,t}, \mathbf{MC}_{i-1,t})
\end{equation}
by including the MIDAS covariates in the QRF model. This is achieved by training the QRF with the training set $\mathbf{s}^*=\{(y_{i,t}, \mathbf{x}_{i-1,t}, \mathbf{MC}_{i-1,t})\}_{i=1, t=1}^{N_t,T}$, which includes the observations of the MIDAS component of each low-frequency covariate.


\vspace{0.15in}

\noindent In this non-parametric context, the likelihood of the MIDAS-QRF model cannot be computed, and consequently $\omega_2$ cannot be optimized via maximum likelihood as in the standard MIDAS model.
To overcome this issue, the optimal $\omega_2$ could be found via grid search as the one delivering the higher forecast accuracy. A similar approach is also discussed in \cite{candila2023mixed}. However, this procedure can be particularly burdersome if the vector $\textbf{Z}_t$ is large. \vspace{0.15in}

\noindent Thus, proposed methodology reduces the computational effort of the MIDAS-QRF training as follows. For each covariate $Z_h$, a set of $MC_{i-1,t}^h$ is computed by using different values of $\omega_2$, obtaining a matrix of MIDAS components. Then, the Principal Component Analysis (PCA) is applied on the resulting matrix to reduce its dimensionality, and the first component of the PCA is used in $\mathbf{s}^*$ as MIDAS component related to $Z_h$.
This procedure is performed separately for each low-frequency covariate in the training set.
\vspace{0.15in}

\noindent The main benefit of this approach is that it retains the most important information in $Z_h$ while reducing computational effort and training time for the MIDAS-QRF. Additionally, like the standard QRF, the MIDAS-QRF assesses the relevance of each covariate in predicting the response variable through the Variable Importance measure. Although this measure cannot be interpreted as the coefficients in parametric models, it allows to enhance the interpretability of Random Forests-based models compared to other machine learning algorithms.
\vspace{0.15in}

\noindent A potential limitation of the MIDAS-QRF is that the estimation procedure to obtain MC values can increase variability in MIDAS-QRF estimates, a common issue in MIDAS models. However, being based on Random Forests, the MIDAS-QRF algorithm allows to reduce estimates variability due to its ensemble nature. Additionally, the MIDAS-QRF retains all statistical properties of the standard QRF algorithm since the MIDAS component is considered as an additional covariate in the training set. For more details on the MIDAS-QRF and QRF statistical properties, refer to \cite{meinshausen2006quantile}, in which the model's consistency is shown along with numerical examples.

\subsection{Variable Importance computation}

Variable Importance is usually computed permuting the observations of the generic $p$-th covariate and measuring the effect on the model's forecast accuracy. 
The idea behind this procedure is that if a covariate significantly affects the model's performance, permuting its values would result in a decrease in forecast accuracy. On the contrary, if the covariate is less important, permuting its values should have minimal influence on the model's performance.
\vspace{0.15in}

\noindent More in detail, the Variable Importance  is computed in two steps. In the first step, Out-Of-Bag (OOB) observations of the training set are used to compute the Sum of Squared Residuals (SSR) of the QRF, denoted with $m$:

\begin{equation}
    m=\sum_{i=1}^S (y_i^{OOB}-\hat{y}_i)^2
\end{equation}

\noindent where $S$ denotes the total number of OOB observations selected in the procedure (in empirical setting, this number is usually pre-determined by the function used to implement the algorithm) and $y_i^{OOB}$ is the $i-th$ OOB observation of the outcome variable.

\noindent Subsequently, the observations of the $p$-th covariate are permuted, and the SSR is re-calculated. The resulting SSR is denoted with $m^*$. 
The importance of the $p$-th variable at each quantile level $\tau$, denoted with $I_{p, \tau}$, is measured as:
\begin{center}
\begin{equation}
I_{p, \tau}=m-m^*
\end{equation}
\end{center}
The bigger the decrease of the SSR after permutation, the greater the variable importance.
The ability to extract the Variable Importance measure allows the MIDAS-QRF to retain the grade of interpretability of standard Random Forest-based algorithms.
\vspace{0.15in}

\noindent In this chapter, the Variable Importance of the covariates included in the training set is extracted to gain a deeper insight into the results obtained in Section \ref{sec:MIDAS-QRF-empirical}.

\subsection{Dynamic MIDAS-QRF}

In order to model the quantile in a dynamic framework, the MIDAS-QRF approach is extended to an autoregressive framework. A dynamic approach to quantile estimation has already been introduced in a parametric setting with the CaViaR model of \cite{engle2004caviar}. The aim of this study is to apply this well-established autoregressive approach to the MIDAS-QRF. Thus, the resulting Dynamic MIDAS-QRF relies on considering lagged values of the quantile predictions as additional covariate used to train the model.

\vspace{0.15in}

\noindent The iterative algorithm of the Dynamic MIDAS-QRF consists in an initialisation phase and a two-step procedure. Denoting with $R= \sum_{t=1}^T N_t$ the total number of observations in $\mathbf{s}^*$, in the initialization consists in computing a vector of quantile forecasts $\mathbf{\hat{Q}}^{\tau}_0=\{\hat{Q}_{r}^{\tau}\}_{r=1}^{V-1}$ with $V<R$ using any suitable autoregressive model, such as CaViaR. 

\vspace{0.15in}

\noindent Then, the two-step procedure consists in:
\begin{enumerate}

\item \textbf{First step}: a MIDAS-QRF is trained considering the training set 
\newline $\mathbf{s}^*=\{y_r, \mathbf{x}_{r-1}, \mathbf{MC}_{r-1}, \hat{Q}_{r-1}^{\tau}\}_{r=1}^{V}$ and used to compute the quantile prediction $\hat{Q}_{V}^{\tau}$.

\item \textbf{Second step:} the training set is updated with the additional quantile prediction $\hat{Q}_{V}^{\tau}$ and the MIDAS-QRF is trained again.
\end{enumerate}

\vspace{0.15in}

\noindent The algorithm iterates between these two steps until the entire dataset 
\newline $\mathbf{s}^*=\{y_r, \mathbf{x}_{r-1}, \mathbf{MC}_{r-1}, \hat{Q}_{r-1}^{\tau}\}_{r=1}^{R}$ is included in the training set with their respective quantile predictions. Finally, a vector of quantile predictions $\hat{\mathbf{Q}}^\tau=\{\hat{Q}^{\tau}_{r}\}_{r=1}^{R}$ is obtained and can be used also to evaluate the forecast accuracy of the model.

\section{Empirical Application} \label{sec:MIDAS-QRF-empirical}

Over the past two decades, there has been a growing interest among investors in using energy commodities as hedging and speculative assets, especially during periods of financial and economic downturns. This phenomenon, known as the financialization of energy commodities, along with the deregulation of over-the-counter markets, led to a significant increase of the volatility of energy commodities returns. 
\vspace{0.15in}

\noindent Thus, the study of the risks linked to such commodities is particularly relevant under a risk management framework. 
For this reason, this section shows the empirical application of the static and dynamic versions of the MIDAS-QRF to forecast the well-known financial risk measure VaR of three energy commodities: WTI Crude Oil, Brent Crude Oil and Heating Oil.
The performance of the proposed models is measured in terms of statistical adequacy by means of backtesting procedures, and in terms of forecast accuracy, measured using the quantile loss function, i.e. the check function, of \cite{koenker1978regression} $\rho_\tau(u)=u(\tau-\mathbf{1}_{\{u<0\}})$. The data summary statistics of the log-returns of each index are reported in Table \ref{tab:sum-stats} along with their graphs in Figures \ref{fig:wti}-\ref{fig:brent}-\ref{fig:heat}.

\vspace{0.15in}

\noindent The MIDAS-QRF and its dynamic specification are employed to estimate one-step-ahead VaR forecasts at three different probability levels ($\tau=0.01, 0.025, 0.05$) with an expanding-window approach by refitting the models every ten days.
The training set considers time series spanning from September 2014 to October 2019. The forecasts are made on an out-of-sample test set composed of 700 observations from November 2019 to April 2022, covering the pre-pandemic, pandemic and post-pandemic period, including the beginning of the Russian-Ukrainian conflict. The richness of information contained in this specific time span allows to train the model in "standard" settings and testing it in setting in which low volatility periods are alternated by periods of high volatility. This approach allows to test the ability of the model to obtain reliable forecast both in standard settings and in unseen and unexpected situations.

\vspace{0.15in}

\noindent The covariates set includes both low-frequency and daily variables. The low-frequency variables are the monthly Real Broad Dollar Index (DOLL), the quarterly Natural Gas returns (NATGAS) and the quarterly Saudi Arabia Crude Oil Production (SAUDI-PROD). Daily variables are the daily lag 1 and lag 2 of the indexes of interest along with the daily Standard and Poor's 500 Index (SP500). The number of lags was selected to create a dataset of manageable size, enabling a comparison between the proposed MIDAS-QRF model and simpler models that consider only a single covariate. All the daily, monthly and quarterly time series have been by computing their log-returns.
The dynamic MIDAS-QRF is trained with the same dataset used for the static version, but the lagged vector of quantile forecasts is introduced as additional covariate, denoted with $lag\_quant$.

\noindent Thus, the MIDAS-QRF equation in this empirical application is:

\begin{equation}
\footnotesize
\hat{Q}_{\tau}(Y_{i,t}|X_{i,t}, \mathbf{MC}_{i-1,t})^{MIDAS-QRF}=f_{\tau}(SP500_{i,t}, MC_{i-1,t}^{DOLL}, MC_{i-1,t}^{NATGAS}, MC_{i-1,t}^{SAUDI})
\end{equation}
 And the dynamic MIDAS-QRF model equation is:

\begin{equation}
\footnotesize
\begin{aligned}
    \hat{Q}_{\tau}(Y_{i,t}|\mathbf{X}_{i,t},\mathbf{MC}_{i-1,t})^{DYN} = f_{\tau}(lag\_quant_{i-1,t},
    SP500_{i,t},     MC_{i-1,t}^{DOLL}, MC_{i-1,t}^{NATGAS}, MC_{i-1,t}^{SAUDI})
\end{aligned}
\end{equation}

\vspace{0.15in}

\noindent The benchmark models set includes the parametric models GARCH and GARCH-MIDAS \citep{engle2013stock}  with Gaussian and Student's-t distributions of the errors, semi-parametric models, that are the four different specifications of the CaViaR model, namely Asymmetric Slope (AS), Symmetric Absolute Value (SAV), Indirect GARCH (IG) and Adaptive (AD), and the standard Quantile Regression Forest of \cite{meinshausen2006quantile}. The functional form of the parametric and semi-parametric models is reported in Table \ref{tab:models_eq}.

\begin{landscape}
    \begin{table}[b]

  \centering
  \caption{Models specifications \label{tab:models_eq}}
  \begin{adjustbox}{max width=2\textwidth}
    \begin{threeparttable}
      \begin{tabular}{l c c}
        \toprule  
        \textit{Model}      &   \textit{Functional Form}     & \textit{Err. Distr.}\\
        \midrule                                                                                                          
            \multirow{2}{*}{GARCH--norm}              & $Y_{i,t}|\mathcal{F}_{i-1,t}  = \sqrt{h_{i,t}} \eta_{i,t}$ & $\eta_{i,t}\overset{i.i.d}{\sim} \mathcal{N}\left(0, 1\right)$\\  
           &      $h_{i,t} = \omega + \alpha_{}  Y_{i-1, t}^2 + \beta h_{i-1, t}$& \\
        \addlinespace
        \hdashline
        \addlinespace
           \multirow{2}{*}{GARCH--t}              & $Y_{i,t}|\mathcal{F}_{i-1,t}  = \sqrt{h_{i,t}} \eta_{i,t}$ & $\eta_{i,t}\overset{i.i.d}{\sim} t_{\nu}$\\  
           &      $h_{i,t} = \omega + \alpha  Y_{i-1, t}^2 + \beta h_{i-1, t}$& \\     
               \addlinespace
        \hdashline
       
        \addlinespace  
               
        GM  &      $\xi_{i,t}=(1-\alpha_{}-\beta_{}-\gamma_{}/2) + \left(\alpha_{} +  \gamma_{} \cdot \mathbbm{1}_{\left(Y_{i-1,t}  < 0 \right)}\right) \frac{Y_{i-1,t}^2}{\pi_t} + \beta_{} \xi_{i-1,t}$& \\
        & $\pi_t=\exp\left\lbrace m + \zeta \sum_{k=1}^K \delta_k(\omega) Z_{t-k}\right\rbrace$\\
        
           \addlinespace
        \hdashline
        \addlinespace

  CAVIAR-SAV       &     $Q^{\tau}_{i,t} = \beta_0+ \beta_1 Q^{\tau}_{i-1,t}+ \beta_2 |Y_{i-1,t}| $&\\
        \addlinespace
        \hdashline
        \addlinespace

CAVIAR-AD & $
Q^{\tau}_{i,t}(\boldsymbol{\beta})=Q_{i-1,t}^{\tau}(\beta_1)+\beta_1 \{[1+exp(G[Y_{i-1,t}-Q^{\tau}_{i-1,t}(\beta_1)])]^{-1}-\tau\}
$&\\
        \addlinespace
        \hdashline
        \addlinespace

          CAVIAR-AS       &    $Q^{\tau}_{i,t} =  \beta_0+  \beta_1 Q^{\tau}_{i-1,t}+ (\beta_{2}\mathbbm{1}_{(Y_{i-1,t}>0)}+\beta_{3} \mathbbm{1}_{(Y_{i-1,t}<0)}) |Y_{i-1,t}|$&\\
        \addlinespace
        \hdashline
        \addlinespace
         CAVIAR-IG      &    $Q^{\tau}_{i,t} =  -\sqrt {\beta_0+ \beta_1 (Q^{\tau}_{i-1,t})^2+ \beta_2 Y_{i-1,t}^2 }$&\\ 
           \addlinespace
        \hdashline
    
        \bottomrule
      \end{tabular}
      \begin{tablenotes}[flushleft]
        \setlength\labelsep{0pt}
        \footnotesize
        \item Functional forms of the parametric models, that is GARCH,  with Gaussian and Student's t distributions for the errors (GARCH-norm, GARCH-std, respectively) and GARCH-MIDAS (GM) models with the four different low-frequency variables. The semi-parametric models are the Aymmetric Absolute Value (SAV), Adaptive (AD), Asymmetric Slope (AS)and Indirect GARCH (IG) specifications of the CaViaR model. We denote with $\mathcal{F}_{i-1,t}$ the information available up to time $i-1,t$.
      \end{tablenotes}
    \end{threeparttable}
  \end{adjustbox}
\end{table}

\end{landscape}

\noindent The computational time to fit the MIDAS-QRF and its dynamic specification is equal, on average, to 471 seconds on an ordinary multi-CPU server Intel Xeon with 24 cores.

\subsection{Backtesting Procedures}

Backtesting procedures represent statistical tests employed in VaR analysis to assess the accuracy and reliability of the the models used to forecast VaR.

\noindent The main backtesting procedures commonly used in the VaR literature and used in this thesis are:

\begin{itemize}
    \item \textbf{Unconditional Coverage Test} \citep{kupiec1995techniques}: tests whether the actual frequency of VaR violations (instances where actual losses exceed the predicted VaR) matches the expected violation frequency. For accurate models, the proportion of VaR violations should align with the VaR confidence level (e.g., for a 5\% VaR, breaches should occur about 5\% of the time).

    \item \textbf{Conditional Coverage Test} \citep{christoffersen1998evaluating}: combines the Unconditional Coverage Test with an independence test to evaluate whether the VaR violations are randomly distributed over time and are, thus, independent. As a matter of fact, the presence of clustered violations might indicate a not reliable and robust model. 

\item \textbf{Dynamic Quantile Test} \citep{Manganelli:2004}: tests whether the VaR violations are serially uncorrelated conditional on previous quantile estimates. 
\end{itemize}

\subsection{MIDAS-QRF Results}
The results in terms of quantile loss and backtesting procedures for the MIDAS-QRF model are presented in Tables \ref{tab:brent}-\ref{tab:wti}-\ref{tab:heat}. The columns UC\_pval, CC\_pval and DQ\_pval indicate the p-value results of the Unconditional Coverage, Conditional Coverage and Dynamic Quantile tests, respectively. The AE column reports the Actual over Expected exceedance ratio. The column $\%Loss$ indicates the ratio between the loss of the static version of the MIDAS-QRF model with respect to the other benchmark models:

$$
\textit{\% Loss}=\frac{Loss_{MIDAS-QRF}}{Loss_{Benchmark}}
$$

\vspace{0.15in}

\noindent The results of the backtesting procedures show that, differently from the benchmark models, the MIDAS-QRF consistently delivers adequate forecasts at all quantile levels for each index. For instance, for the Brent index the MIDAS-QRF is the only model passing all the backtesting procedures at quantile level $0.01$. In terms of forecast accuracy, the MIDAS-QRF outperforms every benchmark model for all index, with a consistent increase in accuracy at the lower quantile levels of the Brent and WTI index.

\begin{table}[H]
\renewcommand{\arraystretch}{0.72}
\centering
\begin{tabular}{ccccccc}
\hline
\textbf{BRENT}                                                                   & \multicolumn{6}{c}{$\tau=0.01$}                               \\ \hline
                               & \textit{Loss}   & \textit{UC\_pval} & \textit{CC\_pval} & \textit{DQ\_pval }& \textit{AE }  & \textit{\%Loss}\\ \hline
\rowcolor[HTML]{D9D9D9} MIDAS-QRF                                            & 16.927 & 0.013    & 0.025    & 0.302    & 2.167 &        \\
GARCH-norm                                                                                        & 32.498 & 0.000    & 0.000    & 0.001    & 6.000 & 52\%   \\
GARCH-std                                                                                         & 21.138 & 0.000    & 0.000    & 0.000    & 3.333 & 80\%   \\
CAViAR-SAV                                                                                        & 22.302 & 0.005    & 0.012    & 0.863    & 2.333 & 76\%   \\
CAViAR-AD                                                                                         & 30.301 & 0.001    & 0.001    & 0.495    & 2.667 & 56\%   \\
CAViAR-AS                                                                                         & 17.763 & 0.000    & 0.000    & 0.476    & 3.000 & 95\%   \\
CAViAR-IG                                                                                         & 17.672 & 0.005    & 0.012    & 0.743    & 2.333 & 96\%   \\
STD-RF                                                                                            & 26.989 & 0.000    & 0.000    & 0.000    & 4.333 & 53\%   \\
GM-DOLL                                                                                           & 21.986 & 0.000    & 0.000    & 0.063    & 3.500 & 77\%   \\
GM-NATGAS                                                                                         & 23.698 & 0.000    & 0.000    & 0.000    & 5.500 & 72\%   \\
GM-SAUDIPROD                                                                                    & 23.040 & 0.000    & 0.000    & 0.001    & 4.167 & 74\%   \\ \hline
                                                                                                  & \multicolumn{6}{c}{$\tau=0.025$}                         \\ \hline
                                                                                                  & \textit{Loss}   & \textit{UC\_pval} & \textit{CC\_pval} & \textit{DQ\_pva}l & \textit{AE}    & \textit{\%Loss} \\ \hline
\rowcolor[HTML]{D9D9D9} MIDAS-QRF                                            & 27.956 & 0.052    & 0.150    & 0.661    & 1.533 &        \\
GARCH-norm                                                                                        & 43.322 & 0.000    & 0.000    & 0.448    & 2.867 & 65\%   \\
GARCH-t                                                                                           & 35.528 & 0.000    & 0.001    & 0.004    & 2.067 & 79\%   \\
\rowcolor[HTML]{D9D9D9} CAViAR-SAV                                           & 31.118 & 0.213    & 0.426    & 0.999    & 1.333 & 90\%   \\
\rowcolor[HTML]{D9D9D9}CAViAR-AD                                                                                         & 41.668 & 0.087    & 0.028    & 0.621    & 1.467 & 67\%   \\
CAViAR-AS                                                                                         & 30.157 & 0.001    & 0.002    & 0.981    & 1.933 & 93\%   \\
\rowcolor[HTML]{D9D9D9} CAViAR-IG                                            & 29.786 & 0.213    & 0.426    & 1.000    & 1.333 & 94\%   \\
QRF                                                                                               & 36.670 & 0.000    & 0.000    & 0.000    & 2.333 & 76\%   \\
GM-DOLL                                                                                           & 33.193 & 0.001    & 0.001    & 0.567    & 2.000 & 84\%   \\
GM-NATGAS                                                                                         & 33.926 & 0.000    & 0.000    & 0.007    & 3.133 & 82\%   \\
GM-SAUDIPROD                                                                                      & 33.301 & 0.000    & 0.000    & 0.251    & 2.333 & 84\%   \\ \hline
                                                                                                  & \multicolumn{6}{c}{$\tau=0.05$}                          \\ \hline
                                                                                                  & \textit{Loss  } & \textit{UC\_pval} & \textit{CC\_pval} & \textit{DQ\_pval} & \textit{AE}    & \%Loss \\ \hline
\rowcolor[HTML]{D9D9D9} MIDAS-QRF                                            & 41.602 & 1.000    & 0.517    & 0.990    & 1.000 &        \\
GARCH-norm                                                                                        & 56.708 & 0.000    & 0.000    & 1.000    & 1.733 & 73\%   \\
\rowcolor[HTML]{D9D9D9}GARCH-t                                                                                           & 50.389 & 0.022    & 0.061    & 0.543    & 1.433 & 82\%   \\
\rowcolor[HTML]{D9D9D9} CAViAR-SAV                                           & 44.244 & 1.000    & 0.920    & 1.000    & 1.000 & 94\%   \\
CAViAR-AD                                                                                         & 54.762 & 0.361    & 0.006    & 0.331    & 1.167 & 76\%   \\
\rowcolor[HTML]{D9D9D9} CAViAR-AS                                                                                         & 42.719 & 0.074    & 0.186    & 0.930    & 1.333 & 97\%   \\
\rowcolor[HTML]{D9D9D9} CAViAR-IG                                            & 44.391 & 0.580    & 0.850    & 1.000    & 1.100 & 94\%   \\
QRF                                                                                               & 49.901 & 0.005    & 0.020    & 0.224    & 1.533 & 83\%   \\
\rowcolor[HTML]{D9D9D9}GM-DOLL & 47.451 & 0.074    & 0.196    & 0.999    & 1.333 & 88\%   \\
GM-NATGAS                                                                                         & 47.105 & 0.000    & 0.000    & 0.212    & 2.167 & 88\%   \\
GM-SAUDIPROD                                                                                      & 46.437 & 0.001    & 0.003    & 0.914    & 1.633 & 89\%  
\\ \hline
\end{tabular}
\caption{Loss and Backtesting results of the MIDAS-QRF for the Brent Index. The shade of grey indicate models for which the p-value of the test in greater than the $1\%$ significance level.}
\label{tab:brent}
\end{table}

\begin{table}[H]
\renewcommand{\arraystretch}{0.72}
\centering
\begin{tabular}{ccccccc}
\hline
\textbf{WTI}                           & \multicolumn{6}{c}{$\tau=0.01$}                         \\ \hline
                                                        & \textit{Loss}   & \textit{UC\_pval} & \textit{CC\_pval} & \textit{DQ\_pval }& \textit{AE }  & \textit{\%Loss} \\ \hline
\rowcolor[HTML]{D9D9D9} MIDAS-QRF  & 16.539 & 0.066    & 0.078    & 0.908    & 1.83 &        \\
GARCH-norm                                              & 34.805 & 0.000    & 0.000    & 0.682    & 3.00 & 47\%   \\
GARCH-t                                                 & 24.362 & 0.001    & 0.002    & 0.035    & 2.67 & 68\%   \\
\rowcolor[HTML]{D9D9D9}CAViaR-SAV                                              & 24.996 & 0.013    & 0.025    & 0.403    & 2.17 & 66\%   \\
CAViaR-AD                                               & 33.682 & 0.005    & 0.000    & 0.177    & 2.33 & 49\%   \\
CAViaR-AS                                               & 24.817 & 0.000    & 0.000    & 0.347    & 3.00 & 67\%   \\
\rowcolor[HTML]{D9D9D9} CAViaR-IG  & 25.017 & 0.066    & 0.078    & 0.921    & 1.83 & 66\%   \\
QRF                                                     & 31.076 & 0.000    & 0.000    & 0.003    & 3.83 & 53\%   \\
GM-DOLL                                                 & 17.640 & 0.005    & 0.002    & 0.056    & 2.33 & 93\%   \\
\rowcolor[HTML]{D9D9D9}GM-NATGAS                                               & 17.773 & 0.013    & 0.025    & 0.034    & 2.17 & 93\%   \\
GM-SAUDIPROD                                            & 18.845 & 0.000    & 0.000    & 0.000    & 3.33 & 88\%   \\ \hline
                                                        & \multicolumn{6}{c}{$\tau=0.025$}                        \\ \hline
                                                        & \textit{Loss}   & \textit{UC\_pval} & \textit{CC\_pval} & \textit{DQ\_pval }& \textit{AE }  & \textit{\%Loss}\\ \hline
\rowcolor[HTML]{D9D9D9} MIDAS-QRF  & 28.277 & 0.609    & 0.201    & 1.000    & 1.13 &        \\
\rowcolor[HTML]{D9D9D9}GARCH-norm                                              & 48.663 & 0.030    & 0.095    & 0.874    & 1.60 & 58\%   \\
GARCH-t                                                 & 37.256 & 0.002    & 0.010    & 0.068    & 1.87 & 76\%   \\
\rowcolor[HTML]{D9D9D9} CAViaR-SAV & 31.218 & 0.087    & 0.032    & 0.680    & 1.47 & 90\%   \\
CAViaR-AD                                               & 45.144 & 0.139    & 0.000    & 0.673    & 1.40 & 62\%   \\
\rowcolor[HTML]{D9D9D9}CAViaR-AS                                               & 29.696 & 0.052    & 0.026    & 0.987    & 1.53 & 95\%   \\
\rowcolor[HTML]{D9D9D9} CAViaR-IG  & 30.968 & 0.087    & 0.032    & 1.000    & 1.47 & 91\%   \\
QRF                                                     & 41.786 & 0.005    & 0.002    & 0.807    & 1.80 & 67\%   \\
\rowcolor[HTML]{D9D9D9}GM-DOLL                                                 & 29.507 & 0.087    & 0.032    & 0.859    & 1.47 & 95\%   \\
\rowcolor[HTML]{D9D9D9}GM-NATGAS                                               & 29.092 & 0.017    & 0.039    & 0.286    & 1.67 & 97\%   \\
GM-SAUDIPROD                                            & 30.501 & 0.000    & 0.000    & 0.047    & 2.27 & 92\%   \\ \hline
                                                        & \multicolumn{6}{c}{$\tau=0.05$}                         \\ \hline
                                                        & \textit{Loss}   & \textit{UC\_pval} & \textit{CC\_pval} & \textit{DQ\_pval }& \textit{AE }  & \textit{\%Loss}\\
\rowcolor[HTML]{D9D9D9} MIDAS-QRF  & 43.150 & 0.050    & 0.113    & 0.997    & 1.37 &        \\
\rowcolor[HTML]{D9D9D9} GARCH-norm & 66.613 & 0.149    & 0.202    & 0.703    & 1.27 & 62\%   \\
GARCH-t                                                 & 53.153 & 0.002    & 0.004    & 0.132    & 1.60 & 81\%   \\
\rowcolor[HTML]{D9D9D9} CAViaR-SAV & 46.644 & 0.361    & 0.510    & 0.997    & 1.17 & 92\%   \\
CAViaR-AD                                               & 62.262 & 0.205    & 0.000    & 0.564    & 1.23 & 69\%   \\
\rowcolor[HTML]{D9D9D9}CAViaR-AS                                               & 49.768 & 0.022    & 0.040    & 0.985    & 1.43 & 87\%   \\
\rowcolor[HTML]{D9D9D9} CAViaR-IG  & 46.762 & 0.149    & 0.212    & 1.000    & 1.27 & 92\%   \\
\rowcolor[HTML]{D9D9D9} QRF        & 56.984 & 0.149    & 0.202    & 0.996    & 1.27 & 76\%   \\
\rowcolor[HTML]{D9D9D9} GM-DOLL    & 45.487 & 0.580    & 0.093    & 0.996    & 1.10 & 95\%   \\
\rowcolor[HTML]{D9D9D9}GM-NATGAS                                               & 44.453 & 0.022    & 0.020    & 0.520    & 1.43 & 97\%   \\
GM-SAUDIPROD                                            & 45.606 & 0.005    & 0.015    & 0.459    & 1.53 & 95\%  
\\ \hline
\end{tabular}
\caption{Loss and Backtesting results of the MIDAS-QRF for the WTI Index.  The shade of grey indicate models for which the p-value of the test in greater than the $1\%$ significance level.}
\label{tab:wti}
\end{table}

\begin{table}[H]
\renewcommand{\arraystretch}{0.72}
\centering
\begin{tabular}{ccccccc}
\hline
\textbf{HEATING OIL}                   & \multicolumn{6}{c}{$\tau=0.01$}                         \\ \hline
                                                        & \textit{Loss}   & \textit{UC\_pval} & \textit{CC\_pval} & \textit{DQ\_pval }& \textit{AE }  & \textit{\%Loss}\\ \hline
\rowcolor[HTML]{D9D9D9} MIDAS-QRF  & 12.120 & 0.134    & 0.116    & 0.11     & 1.67 &        \\
\rowcolor[HTML]{D9D9D9}GARCH-norm                                              & 12.948 & 0.030    & 0.047    & 0.266    & 2.00 & 93\%   \\
\rowcolor[HTML]{D9D9D9} GARCH-t    & 12.519 & 0.252    & 0.152    & 0.298    & 1.50 & 97\%   \\
\rowcolor[HTML]{D9D9D9}CAViaR-SAV                                              & 12.646 & 0.134    & 0.010    & 0.302    & 1.67 & 96\%   \\
CAViaR-AD                                               & 20.670 & 0.001    & 0.000    & 0.000    & 2.67 & 59\%   \\
\rowcolor[HTML]{D9D9D9} CAViaR-AS  & 13.217 & 0.134    & 0.116    & 0.294    & 1.67 & 92\%   \\
\rowcolor[HTML]{D9D9D9} CAViaR-IG  & 51.123 & 0.252    & 0.452    & 0.356    & 1.50 & 24\%   \\
QRF                                                     & 14.545 & 0.001    & 0.002    & 0.009    & 2.67 & 83\%   \\
GM-DOLL                                                 & 12.946 & 0.005    & 0.002    & 0.063    & 2.33 & 94\%   \\
\rowcolor[HTML]{D9D9D9}GM-NATGAS                                               & 12.992 & 0.134    & 0.010    & 0.292    & 1.67 & 93\%   \\
GM-SAUDIPROD                                            & 14.354 & 0.000    & 0.000    & 0.000    & 3.83 & 84\%   \\ \hline
                                                        & \multicolumn{6}{c}{$\tau=0.025$}                        \\ \hline
                                                        & \textit{Loss}   & \textit{UC\_pval} & \textit{CC\_pval} & \textit{DQ\_pval }& \textit{AE }  & \textit{\%Loss}\\
\rowcolor[HTML]{D9D9D9} MIDAS-QRF  & 22.461 & 0.052    & 0.085    & 0.247    & 1.53 &        \\
\rowcolor[HTML]{D9D9D9}GARCH-norm                                              & 22.756 & 0.213    & 0.039    & 0.974    & 1.33 & 99\%   \\
\rowcolor[HTML]{D9D9D9}GARCH-t                                                 & 22.808 & 0.213    & 0.039    & 0.974    & 1.33 & 98\%   \\
\rowcolor[HTML]{D9D9D9} CAViaR-SAV & 22.638 & 0.796    & 0.180    & 0.967    & 1.07 & 99\%   \\
CAViaR-AD                                               & 31.125 & 0.001    & 0.000    & 0.000    & 1.93 & 73\%   \\
\rowcolor[HTML]{D9D9D9} CAViaR-AS  & 23.138 & 0.213    & 0.426    & 0.978    & 1.33 & 97\%   \\
 CAViaR-IG  & 23.137 & 0.009    & 0.010    & 0.966    & 1.73 & 97\%   \\
QRF                                                     & 26.242 & 0.000    & 0.000    & 0.000    & 2.20 & 86\%   \\
\rowcolor[HTML]{D9D9D9} GM-DOLL    & 22.730 & 0.05     & 0.09     & 0.65     & 1.53 & 99\%   \\
\rowcolor[HTML]{D9D9D9} GM-NATGAS  & 22.610 & 0.14     & 0.15     & 0.78     & 1.40 & 99\%   \\
GM-SAUDIPROD                                            & 24.230 & 0.000    & 0.000    & 0.14     & 2.20 & 93\%   \\ \hline
                                                        & \multicolumn{6}{c}{$\tau=0.05$}                         \\ \hline
                                                        & \textit{Loss}   & \textit{UC\_pval} & \textit{CC\_pval} & \textit{DQ\_pval }& \textit{AE }  & \textit{\%Loss}\\ \hline
\rowcolor[HTML]{D9D9D9} MIDAS-QRF  & 35.515 & 0.034    & 0.024    & 0.014    & 1.40 &        \\
\rowcolor[HTML]{D9D9D9} GARCH-norm & 35.763 & 0.852    & 0.566    & 0.995    & 1.03 & 99\%   \\
\rowcolor[HTML]{D9D9D9} GARCH-t    & 35.893 & 0.580    & 0.276    & 0.997    & 1.10 & 98\%   \\
\rowcolor[HTML]{D9D9D9} CAViaR-SAV & 35.399 & 0.275    & 0.467    & 0.997    & 1.20 & 100\%  \\
CAViaR-AD                                               & 42.767 & 0.034    & 0.000    & 0.000    & 1.40 & 83\%   \\
\rowcolor[HTML]{D9D9D9} CAViaR-AS  & 35.540 & 0.205    & 0.108    & 0.998    & 1.23 & 99\%   \\
 CAViaR-IG  & 37.269 & 0.009    & 0.031    & 0.985    & 1.50 & 95\%   \\
 QRF        & 39.255 & 0.005    & 0.015    & 0.022    & 1.53 & 90\%   \\
\rowcolor[HTML]{D9D9D9} GM-DOLL    & 35.315 & 0.463    & 0.763    & 0.999    & 1.13 & 99\%   \\
\rowcolor[HTML]{D9D9D9} GM-NATGAS  & 35.095 & 0.361    & 0.281    & 0.999    & 1.17 & 100\%  \\
\rowcolor[HTML]{D9D9D9}GM-SAUDIPROD                                            & 36.531 & 0.022    & 0.040    & 0.980    & 1.43 & 97\%  \\ \hline
\end{tabular}
\caption{Loss and Backtesting results of the MIDAS-QRF for the Heating Oil Index.  The shade of grey indicate models for which the p-value of the test in greater than the $1\%$ significance level.}
\label{tab:heat}

\end{table}

\newpage

\subsection{Dynamic MIDAS-QRF Results}
The forecast accuracy and the results of the  backtesting procedures of the dynamic MIDAS-QRF model are presented in Tables \ref{tab:brent-dyn}-\ref{tab:wti-dyn}-\ref{tab:heat-dyn}.

\vspace{0.15in}

\noindent These results highlight that, similarly to the MIDAS-QRF, the Dynamic MIDAS-QRF passess all of the backtesting procedures at all quantile levels for each index. In terms of forecast accuracy, the dynamic specification of the MIDAS-QRF outperforms all benchmark models at all quantile level of every index, and the most relevant increase in forecast accuracy is achieved for the WTI and Brent index.
Moreover, the comparison in terms of forecast accuracy with the static MIDAS-QRF highlights that the autoregressive structure of the dynamic MIDAS-QRF  allows to gain a higher degree of accuracy especially at the lower quantile level $0.01$ of each index. This result suggests that introducing a dynamic element into the MIDAS-QRF allows to model tail risk more accurately, especially when the distribution of the variable changes over time.

\begin{table}[H]
\renewcommand{\arraystretch}{0.7}
\centering
\begin{tabular}{ccccccc}
\hline
\textbf{BRENT} & \multicolumn{6}{c}{$\tau=0.01$}                          \\ \hline
                                & \textit{Loss}   & \textit{UC\_pval} & \textit{CC\_pval} & \textit{DQ\_pval} & \textit{AE }   & \textit{\%Loss} \\ \hline
\rowcolor[HTML]{D9D9D9} 
DYN MIDAS-QRF                       & 15.922 & 0.013    & 0.025    & 0.231    & 2.167 &        \\
\rowcolor[HTML]{D9D9D9} 
MIDAS-QRF                   & 16.927 & 0.013    & 0.025    & 0.302    & 2.167 & 94\%   \\
GARCH-norm                      & 32.498 & 0.000    & 0.000    & 0.001    & 6.000 & 49\%   \\
GARCH-std                       & 21.138 & 0.000    & 0.000    & 0.000    & 3.333 & 75\%   \\
CAViAR-SAV                      & 22.302 & 0.005    & 0.012    & 0.863    & 2.333 & 71\%   \\
CAViAR-AD                       & 30.301 & 0.001    & 0.001    & 0.495    & 2.667 & 53\%   \\
CAViAR-AS                       & 17.763 & 0.000    & 0.000    & 0.476    & 3.000 & 90\%   \\
CAViAR-IG                       & 17.672 & 0.005    & 0.012    & 0.743    & 2.333 & 90\%   \\
STD-RF                          & 26.989 & 0.000    & 0.000    & 0.000    & 4.333 & 59\%   \\
GM-DOLL                         & 21.986 & 0.000    & 0.000    & 0.063    & 3.500 & 72\%   \\
GM-NATGAS                       & 23.698 & 0.000    & 0.000    & 0.000    & 5.500 & 67\%   \\
GM-SAUDIPROD                  & 23.040 & 0.000    & 0.000    & 0.001    & 4.167 & 69\%   \\ \hline
                                & \multicolumn{6}{c}{$\tau=0.025$}                         \\ \hline
                                & \textit{Loss}   & \textit{UC\_pval} & \textit{CC\_pval} & \textit{DQ\_pval} & \textit{AE }   & \textit{\%Loss} \\ \hline
\rowcolor[HTML]{D9D9D9} 
DYN MIDAS-QRF                        & 27.288 & 1.000    & 0.682    & 0.997    & 1.000 &        \\
\rowcolor[HTML]{D9D9D9} 
MIDAS-QRF                             & 27.956 & 0.052    & 0.150    & 0.661    & 1.533 & 98\%   \\
GARCH-norm                      & 43.322 & 0.000    & 0.000    & 0.448    & 2.867 & 63\%   \\
GARCH-t                         & 35.528 & 0.000    & 0.001    & 0.004    & 2.067 & 77\%   \\
\rowcolor[HTML]{D9D9D9} 
CAViAR-SAV                      & 31.118 & 0.213    & 0.426    & 0.999    & 1.333 & 88\%   \\
\rowcolor[HTML]{D9D9D9} CAViAR-AD                       & 41.668 & 0.087    & 0.028    & 0.621    & 1.467 & 65\%   \\
CAViAR-AS                       & 30.157 & 0.001    & 0.002    & 0.981    & 1.933 & 90\%   \\
\rowcolor[HTML]{D9D9D9} 
CAViAR-IG                       & 29.786 & 0.213    & 0.426    & 1.000    & 1.333 & 92\%   \\
QRF                             & 36.670 & 0.000    & 0.000    & 0.000    & 2.333 & 74\%   \\
GM-DOLL                         & 33.193 & 0.001    & 0.001    & 0.567    & 2.000 & 82\%   \\
GM-NATGAS                       & 33.926 & 0.000    & 0.000    & 0.007    & 3.133 & 80\%   \\
GM-SAUDIPROD                   & 33.301 & 0.000    & 0.000    & 0.251    & 2.333 & 82\%   \\ \hline
                                & \multicolumn{6}{c}{$\tau=0.05$}                          \\ \hline
                                & \textit{Loss}   & \textit{UC\_pval} & \textit{CC\_pval} & \textit{DQ\_pval} & \textit{AE }   & \textit{\%Loss} \\ \hline
\rowcolor[HTML]{D9D9D9} 
DYN MIDAS-QRF                        & 41.967 & 0.711    & 0.593    & 0.999    & 1.067 &        \\
\rowcolor[HTML]{D9D9D9} 
QRF                             & 41.602 & 1.000    & 0.517    & 0.990    & 1.000 & 101\%  \\
GARCH-norm                      & 56.708 & 0.000    & 0.000    & 1.000    & 1.733 & 74\%   \\
\rowcolor[HTML]{D9D9D9} GARCH-t                         & 50.389 & 0.022    & 0.061    & 0.543    & 1.433 & 83\%   \\
\rowcolor[HTML]{D9D9D9} 
CAViAR-SAV                      & 44.244 & 1.000    & 0.920    & 1.000    & 1.000 & 95\%   \\
CAViAR-AD                       & 54.762 & 0.361    & 0.006    & 0.331    & 1.167 & 77\%   \\
\rowcolor[HTML]{D9D9D9} CAViAR-AS                       & 42.719 & 0.074    & 0.186    & 0.930    & 1.333 & 98\%   \\
\rowcolor[HTML]{D9D9D9} 
CAViAR-IG                       & 44.391 & 0.580    & 0.850    & 1.000    & 1.100 & 95\%   \\
QRF                             & 49.901 & 0.005    & 0.020    & 0.224    & 1.533 & 84\%   \\
\rowcolor[HTML]{D9D9D9} 
\cellcolor[HTML]{D9D9D9}GM-DOLL & 47.451 & 0.074    & 0.196    & 0.999    & 1.333 & 88\%   \\
GM-NATGAS                       & 47.105 & 0.000    & 0.000    & 0.212    & 2.167 & 89\%   \\
GM-SAUDIPROD                   & 46.437 & 0.001    & 0.003    & 0.914    & 1.633 & 90\%   \\ \hline
\end{tabular}
\caption{Loss and Backtesting results of the Dynamic MIDAS-QRF for the Brent Index.  The shade of grey indicate models for which the p-value of the test in greater than the $1\%$ significance level.}
\label{tab:brent-dyn}
\end{table}

\begin{table}[H]
\renewcommand{\arraystretch}{0.7}
\centering
\begin{tabular}{ccccccc}
\hline
\textbf{WTI} & \multicolumn{6}{c}{$\tau=0.01$}                         \\ \hline
                              & \textit{Loss}   & \textit{UC\_pval} & \textit{CC\_pval} & \textit{DQ\_pval }& \textit{AE }  & \textit{\%Loss}\\ \hline
\rowcolor[HTML]{D9D9D9} 
DYN MIDAS-QRF                       & 14.848 & 0.689    & 0.166    & 1.000    & 1.17 &        \\
\rowcolor[HTML]{D9D9D9} 
MIDAS-QRF                           & 16.539 & 0.066    & 0.078    & 0.908    & 1.83 & 90\%   \\
GARCH-norm                    & 34.805 & 0.000    & 0.000    & 0.682    & 3.00 & 43\%   \\
GARCH-t                       & 24.362 & 0.001    & 0.002    & 0.035    & 2.67 & 61\%   \\
\rowcolor[HTML]{D9D9D9} CAViaR-SAV                    & 24.996 & 0.013    & 0.025    & 0.403    & 2.17 & 59\%   \\
CAViaR-AD                     & 33.682 & 0.005    & 0.000    & 0.177    & 2.33 & 44\%   \\
CAViaR-AS                     & 24.817 & 0.000    & 0.000    & 0.347    & 3.00 & 60\%   \\
\rowcolor[HTML]{D9D9D9} 
CAViaR-IG                     & 25.017 & 0.066    & 0.078    & 0.921    & 1.83 & 59\%   \\
QRF                           & 31.076 & 0.000    & 0.000    & 0.003    & 3.83 & 48\%   \\
GM-DOLL                       & 17.640 & 0.005    & 0.002    & 0.056    & 2.33 & 84\%   \\
\rowcolor[HTML]{D9D9D9} GM-NATGAS                     & 17.773 & 0.013    & 0.025    & 0.034    & 2.17 & 84\%   \\
GM-SAUDIPROD                 & 18.845 & 0.000    & 0.000    & 0.000    & 3.33 & 79\%   \\ \hline
                              & \multicolumn{6}{c}{$\tau=0.025$}                        \\ \hline
                              & \textit{Loss}   & \textit{UC\_pval} & \textit{CC\_pval} & \textit{DQ\_pval }& \textit{AE }  & \textit{\%Loss}\\ \hline
\rowcolor[HTML]{D9D9D9} 
DYN MIDAS-QRF                       & 28.160 & 0.609    & 0.201    & 1.000    & 1.13 &        \\
\rowcolor[HTML]{D9D9D9} 
MIDAS-QRF                           & 28.277 & 0.609    & 0.201    & 1.000    & 1.13 & 100\%  \\
\rowcolor[HTML]{D9D9D9} GARCH-norm                    & 48.663 & 0.030    & 0.095    & 0.874    & 1.60 & 58\%   \\
GARCH-t                       & 37.256 & 0.002    & 0.010    & 0.068    & 1.87 & 76\%   \\
\rowcolor[HTML]{D9D9D9} 
CAViaR-SAV                    & 31.218 & 0.087    & 0.032    & 0.680    & 1.47 & 90\%   \\
CAViaR-AD                     & 45.144 & 0.139    & 0.000    & 0.673    & 1.40 & 62\%   \\
\rowcolor[HTML]{D9D9D9} CAViaR-AS                     & 29.696 & 0.052    & 0.026    & 0.987    & 1.53 & 95\%   \\
\rowcolor[HTML]{D9D9D9} 
CAViaR-IG                     & 30.968 & 0.087    & 0.032    & 1.000    & 1.47 & 91\%   \\
QRF                           & 41.786 & 0.005    & 0.002    & 0.807    & 1.80 & 67\%   \\
GM-DOLL                       & 29.507 & 0.087    & 0.032    & 0.859    & 1.47 & 95\%   \\
GM-NATGAS                     & 29.092 & 0.017    & 0.039    & 0.286    & 1.67 & 97\%   \\
GM-SAUDIPROD                 & 30.501 & 0.000    & 0.000    & 0.047    & 2.27 & 92\%   \\ \hline
                              & \multicolumn{6}{c}{$\tau=0.05$}                         \\ \hline
                              & \textit{Loss}   & \textit{UC\_pval} & \textit{CC\_pval} & \textit{DQ\_pval }& \textit{AE }  & \textit{\%Loss}\\ \hline
\rowcolor[HTML]{D9D9D9} 
DYN MIDAS-QRF                       & 43.800 & 0.568    & 0.672    & 0.991    & 0.90 &        \\
\rowcolor[HTML]{D9D9D9} 
MIDAS-QRF                           & 43.150 & 0.050    & 0.113    & 0.997    & 1.37 & 102\%  \\
\rowcolor[HTML]{D9D9D9} 
GARCH-norm                    & 66.613 & 0.149    & 0.202    & 0.703    & 1.27 & 66\%   \\
GARCH-t                       & 53.153 & 0.002    & 0.004    & 0.132    & 1.60 & 82\%   \\
\rowcolor[HTML]{D9D9D9} 
CAViaR-SAV                    & 46.644 & 0.361    & 0.510    & 0.997    & 1.17 & 94\%   \\
CAViaR-AD                     & 62.262 & 0.205    & 0.000    & 0.564    & 1.23 & 70\%   \\
CAViaR-AS                     & 49.768 & 0.022    & 0.040    & 0.985    & 1.43 & 88\%   \\
\rowcolor[HTML]{D9D9D9} 
CAViaR-IG                     & 46.762 & 0.149    & 0.212    & 1.000    & 1.27 & 94\%   \\
\rowcolor[HTML]{D9D9D9} 
QRF                           & 56.984 & 0.149    & 0.202    & 0.996    & 1.27 & 77\%   \\
\rowcolor[HTML]{D9D9D9} 
GM-DOLL                       & 45.487 & 0.580    & 0.093    & 0.996    & 1.10 & 96\%   \\
\rowcolor[HTML]{D9D9D9} GM-NATGAS                     & 44.453 & 0.022    & 0.020    & 0.520    & 1.43 & 99\%   \\
GM-SAUDIPROD                 & 45.606 & 0.005    & 0.015    & 0.459    & 1.53 & 96\%   \\ \hline
\end{tabular}
\caption{Loss and Backtesting results of the Dynamic MIDAS-QRF for the WTI Index.  The shade of grey indicate models for which the p-value of the test in greater than the $1\%$ significance level.}
\label{tab:wti-dyn}
\end{table}

\begin{table}[H]
\renewcommand{\arraystretch}{0.7}
\centering
\begin{tabular}{ccccccc}
\hline
\textbf{HEATING OIL} & \multicolumn{6}{c}{$\tau=0.01$}                         \\ \hline
                               & \textit{Loss}   & \textit{UC\_pval} & \textit{CC\_pval} & \textit{DQ\_pval }& \textit{AE }  & \textit{\%Loss}\\ \hline
\rowcolor[HTML]{D9D9D9} 
DYN MIDAS-QRF                        & 11.827 & 0.252    & 0.153    & 0.178    & 1.50 &        \\
\rowcolor[HTML]{D9D9D9} 
MIDAS-QRF                            & 12.120 & 0.134    & 0.116    & 0.11     & 1.67 & 98\%   \\
\rowcolor[HTML]{D9D9D9} GARCH-norm                     & 12.948 & 0.030    & 0.047    & 0.266    & 2.00 & 91\%   \\
\rowcolor[HTML]{D9D9D9} 
GARCH-t                        & 12.519 & 0.252    & 0.152    & 0.298    & 1.50 & 94\%   \\
\rowcolor[HTML]{D9D9D9} CAViaR-SAV                     & 12.646 & 0.134    & 0.010    & 0.302    & 1.67 & 94\%   \\
CAViaR-AD                      & 20.670 & 0.001    & 0.000    & 0.000    & 2.67 & 57\%   \\
\rowcolor[HTML]{D9D9D9} 
CAViaR-AS                      & 13.217 & 0.134    & 0.116    & 0.294    & 1.67 & 89\%   \\
\rowcolor[HTML]{D9D9D9} 
CAViaR-IG                      & 51.123 & 0.252    & 0.452    & 0.356    & 1.50 & 23\%   \\
QRF                            & 14.545 & 0.001    & 0.002    & 0.009    & 2.67 & 81\%   \\
GM-DOLL                        & 12.946 & 0.005    & 0.002    & 0.063    & 2.33 & 91\%   \\
\rowcolor[HTML]{D9D9D9} GM-NATGAS                      & 12.992 & 0.134    & 0.010    & 0.292    & 1.67 & 91\%   \\
GM-SAUDIPROD                  & 14.354 & 0.000    & 0.000    & 0.000    & 3.83 & 82\%   \\ \hline
                               & \multicolumn{6}{c}{$\tau=0.025$}                        \\ \hline
                               & \textit{Loss}   & \textit{UC\_pval} & \textit{CC\_pval} & \textit{DQ\_pval }& \textit{AE }  & \textit{\%Loss}\\ \hline
\rowcolor[HTML]{D9D9D9} 
DYN MIDAS-QRF                        & 20.975 & 0.213    & 0.177    & 0.777    & 1.33 &        \\
\rowcolor[HTML]{D9D9D9} 
MIDAS-QRF                            & 22.461 & 0.052    & 0.085    & 0.247    & 1.53 & 93\%   \\
\rowcolor[HTML]{D9D9D9} GARCH-norm                     & 22.756 & 0.213    & 0.039    & 0.974    & 1.33 & 92\%   \\
\rowcolor[HTML]{D9D9D9} GARCH-t                        & 22.808 & 0.213    & 0.039    & 0.974    & 1.33 & 92\%   \\
\rowcolor[HTML]{D9D9D9} 
CAViaR-SAV                     & 22.638 & 0.796    & 0.180    & 0.967    & 1.07 & 93\%   \\
CAViaR-AD                      & 31.125 & 0.001    & 0.000    & 0.000    & 1.93 & 67\%   \\
\rowcolor[HTML]{D9D9D9} 
CAViaR-AS                      & 23.138 & 0.213    & 0.426    & 0.978    & 1.33 & 91\%   \\

CAViaR-IG                      & 23.137 & 0.009    & 0.010    & 0.966    & 1.73 & 91\%   \\
QRF                            & 26.242 & 0.000    & 0.000    & 0.000    & 2.20 & 80\%   \\
\rowcolor[HTML]{D9D9D9} 
GM-DOLL                        & 22.73  & 0.05     & 0.09     & 0.65     & 1.53 & 92\%   \\
\rowcolor[HTML]{D9D9D9} 
GM-NATGAS                      & 22.61  & 0.14     & 0.15     & 0.78     & 1.40 & 93\%   \\
GM-SAUDIPROD                  & 24.230 & 0.000    & 0.000    & 0.14     & 2.20 & 87\%   \\ \hline
                               & \multicolumn{6}{c}{$\tau=0.05$}                         \\ \hline
                               & \textit{Loss}   & \textit{UC\_pval} & \textit{CC\_pval} & \textit{DQ\_pval }& \textit{AE }  & \textit{\%Loss}\\ \hline
\rowcolor[HTML]{D9D9D9} 
DYN MIDAS-QRF                        & 34.416 & 0.106    & 0.090    & 0.046    & 1.30 &        \\
\rowcolor[HTML]{D9D9D9} 
MIDAS-QRF                            & 35.515 & 0.034    & 0.024    & 0.014    & 1.40 & 97\%   \\
\rowcolor[HTML]{D9D9D9} 
GARCH-norm                     & 35.763 & 0.852    & 0.566    & 0.995    & 1.03 & 96\%   \\
\rowcolor[HTML]{D9D9D9} 
GARCH-t                        & 35.893 & 0.580    & 0.276    & 0.997    & 1.10 & 96\%   \\
\rowcolor[HTML]{D9D9D9} 
CAViaR-SAV                     & 35.399 & 0.275    & 0.467    & 0.997    & 1.20 & 97\%   \\
CAViaR-AD                      & 42.767 & 0.034    & 0.000    & 0.000    & 1.40 & 80\%   \\
\rowcolor[HTML]{D9D9D9} 
CAViaR-AS                      & 35.540 & 0.205    & 0.108    & 0.998    & 1.23 & 97\%   \\

CAViaR-IG                      & 37.269 & 0.009    & 0.031    & 0.985    & 1.50 & 92\%   \\

QRF                            & 39.255 & 0.005    & 0.015    & 0.022    & 1.53 & 88\%   \\
\rowcolor[HTML]{D9D9D9} 
GM-DOLL                        & 35.315 & 0.463    & 0.763    & 0.999    & 1.13 & 97\%   \\
\rowcolor[HTML]{D9D9D9} 
GM-NATGAS                      & 35.095 & 0.361    & 0.281    & 0.999    & 1.17 & 98\%   \\
\rowcolor[HTML]{D9D9D9} GM-SAUDIPROD                  & 36.531 & 0.022    & 0.040    & 0.980    & 1.43 & 94\%   \\ \hline
\end{tabular}
\caption{Loss and Backtesting results of the Dynamic MIDAS-QRF for the Heating Oil Index.  The shade of grey indicate models for which the p-value of the test in greater than the $1\%$ significance level.}
\label{tab:heat-dyn}
\end{table}

\begin{figure}[H]
    \centering
    \includegraphics[width=0.93\textwidth]{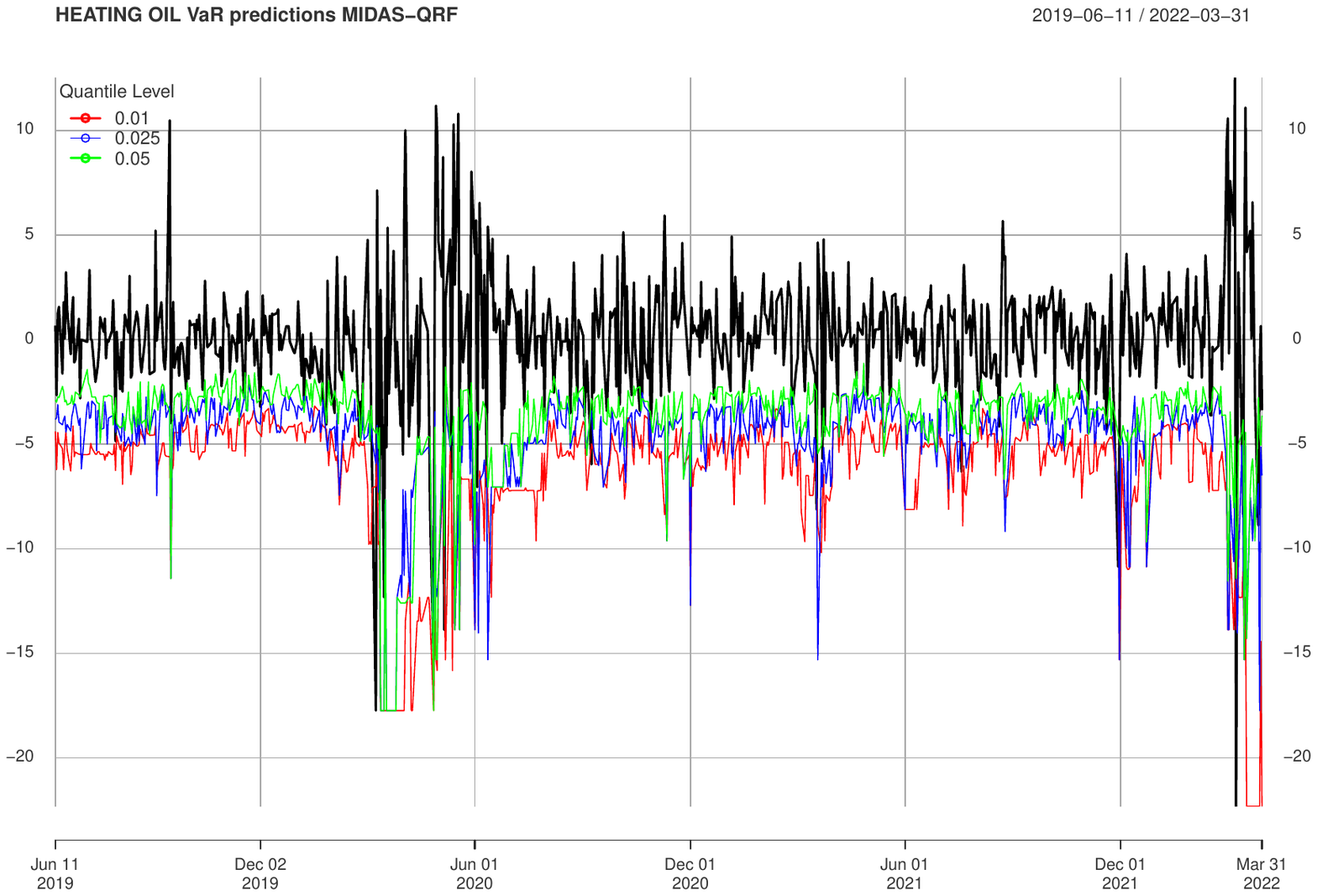}
\includegraphics[width=0.93\textwidth]{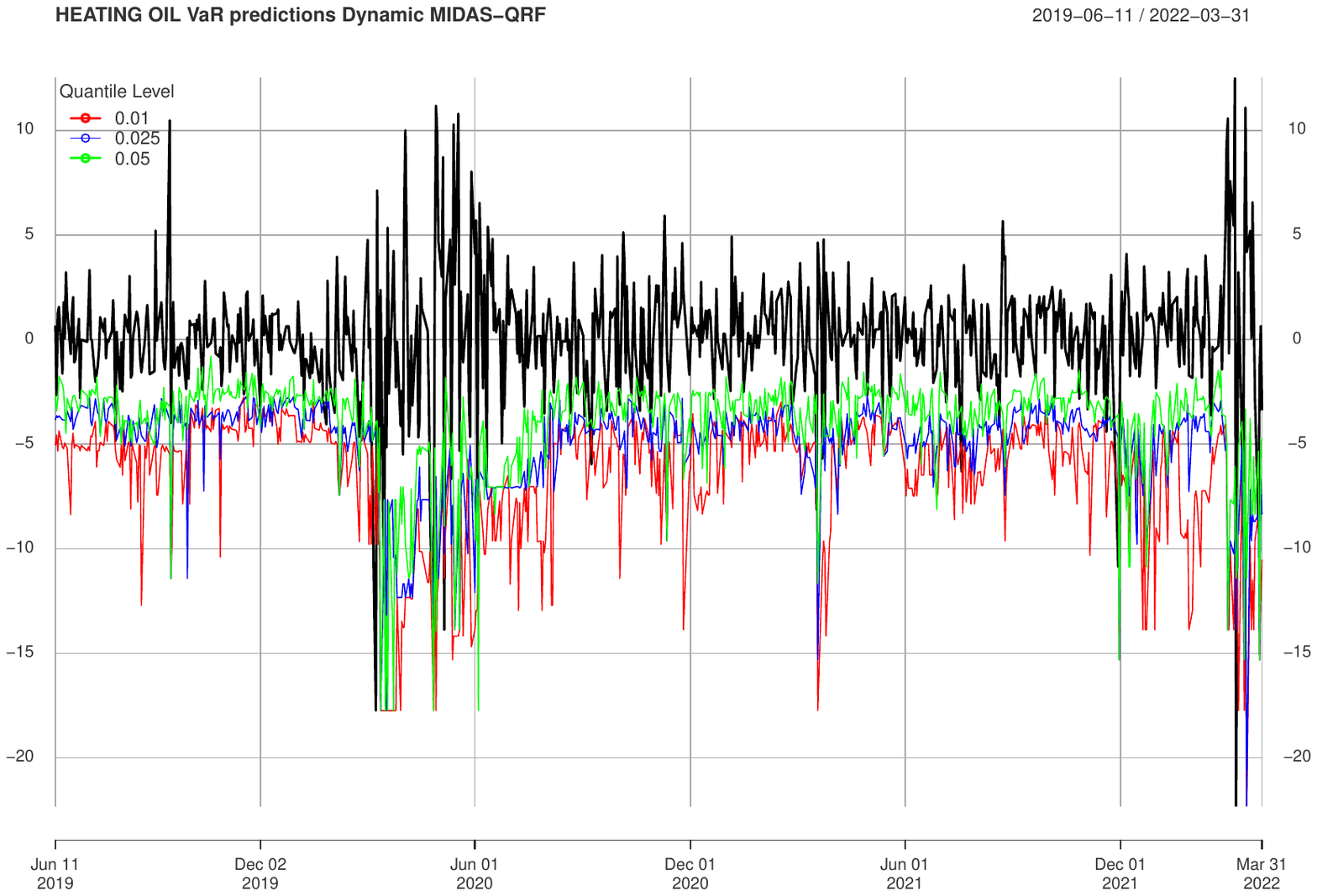}
    \caption{Heating Oil (black line) index out-of-sample predictions at quantile levels $\tau= 0.01, 0.025, 0.05$. The top panel and the bottom panel show the predictions obtained with the dynamic MIDAS-QRF model, respectively.}
    \label{fig:heat-pred}
\end{figure}

Figure \ref{fig:heat-pred} displays the predictions at the three quantile levels of the Heating Oil indexes obtained with both versions of the MIDAS-QRF model. Predictions of the remaining two indexes are represented in Figures \ref{fig:brent-pred}-\ref{fig:wti-pred}. It is worth noticing that, for all three indexes, both the MIDAS-QRF and its dynamic specification effectively manage to obtain accurate forecasts both in situations similar to the one in the training set, as well as during the unexpected high volatility periods related to the pandemic and the Russian-Ukrainian conflict.

\subsection{Variable Importance}

\begin{figure}[H]
    \centering
    \captionsetup{width=.8\linewidth}
\includegraphics[trim={0.6cm 0 0 0},clip, width=\textwidth, height=9.5cm]{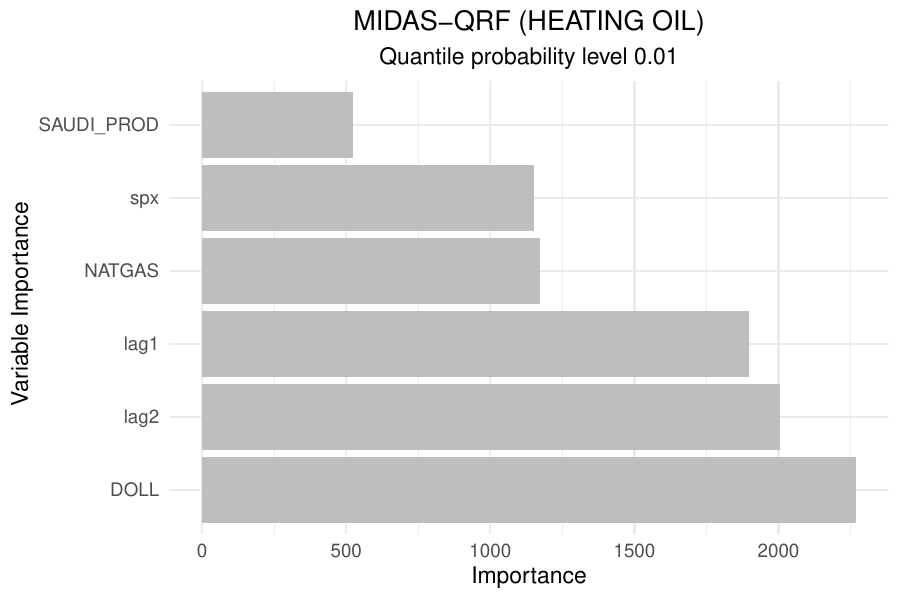}
\includegraphics[trim={0.6cm 0 0 0},clip, width=\textwidth, height=9.5cm]{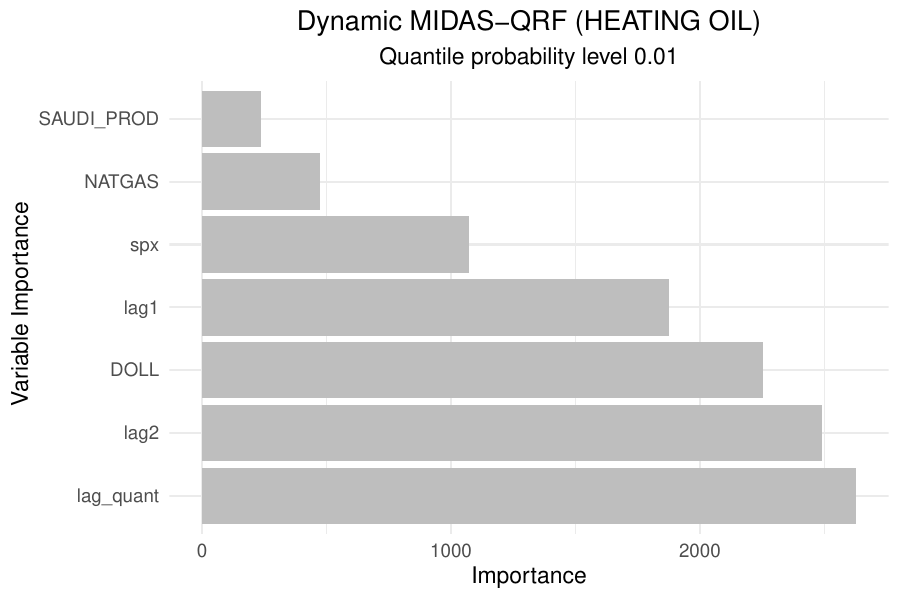}
\caption{Variable importance for the static MIDAS-QRF at $\tau=0.01$ for the Heating Oil index}
\label{fig:imp_01}
\end{figure}

\begin{figure}[H]
    \centering
    \includegraphics[trim={0.6cm 0 0 0},clip, width=\textwidth, height=9.5cm]{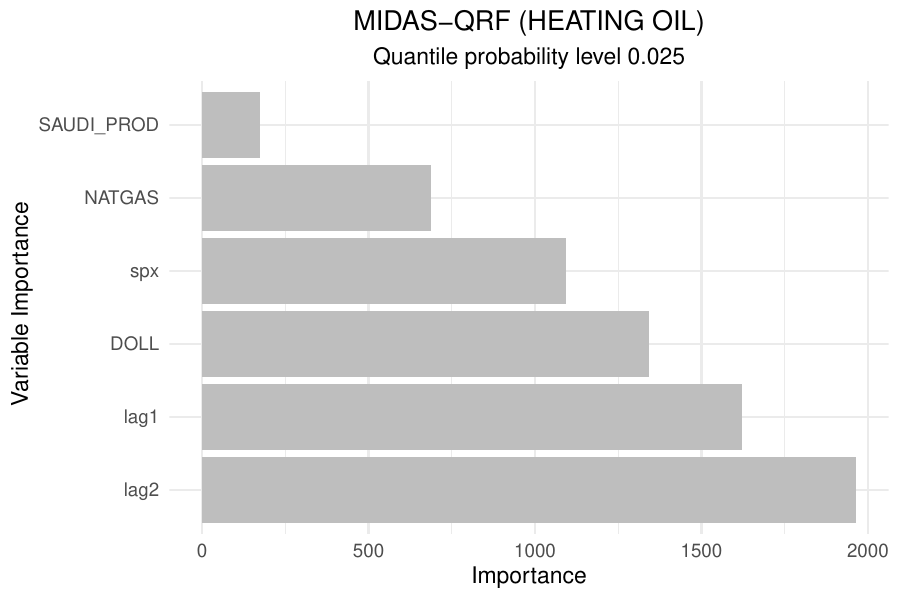}
    \includegraphics[trim={0.6cm 0 0 0},clip, width=\textwidth, height=9.5cm]{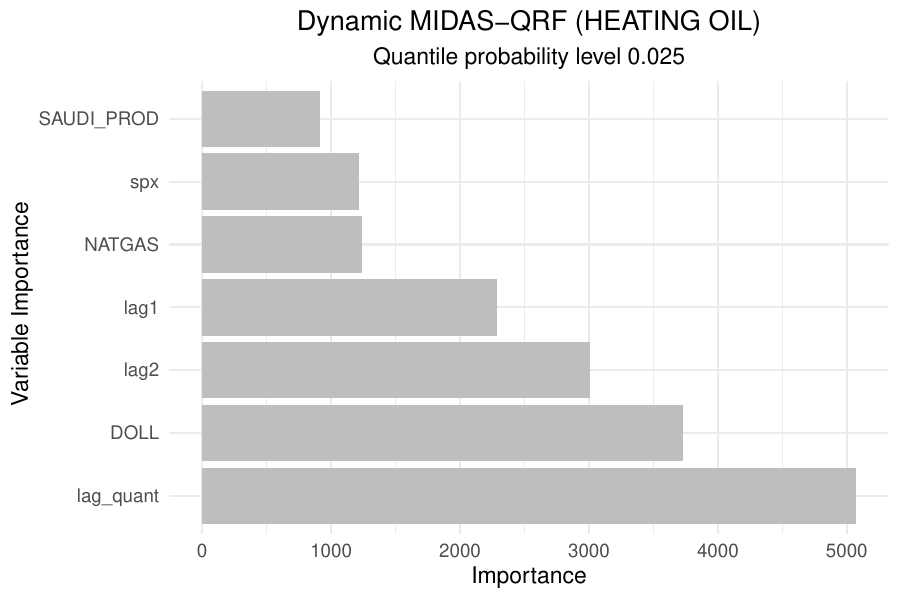}
    \caption{Variable importance for the static MIDAS-QRF at $\tau=0.025$ for the Heating Oil index}
    \label{fig:imp_025}
\end{figure}

\begin{figure}[H]
    \centering
    \includegraphics[trim={0.6cm 0 0 0},clip, width=\textwidth, height=9.5cm]{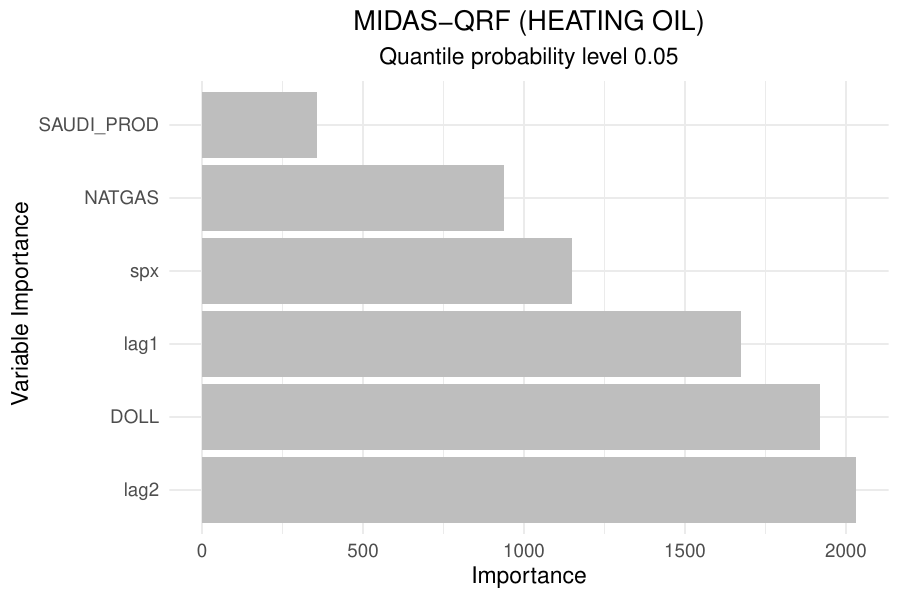}
    \includegraphics[trim={0.6cm 0 0 0},clip, width=\textwidth, height=9.5cm]{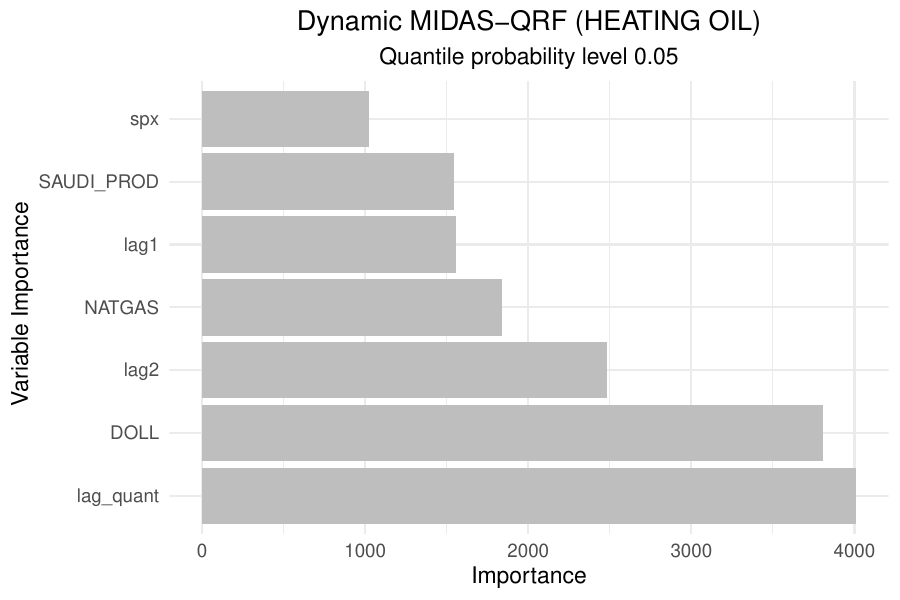}
    \caption{Variable importance for the static MIDAS-QRF at $\tau=0.05$ for the Heating Oil index}
    \label{fig:imp_05}
\end{figure}

The variable importance of the static and dynamic MIDAS-QRF is computed to evaluate the relevance of macroeconomic variables and the $lag\_quant$ variable in training the MIDAS-QRF and the dynamic MIDAS-QRF.
Figures \ref{fig:imp_01}-\ref{fig:imp_025}-\ref{fig:imp_05} depict the bar graphs showing the variable importance of the proposed models at all three quantile levels for the Heating Oil index. 

\vspace{0.15in}

\noindent The variable importance analysis reveals the significant role of macroeconomic variables in both the static and dynamic MIDAS-QRF models. For instance, in the MIDAS-QRF the variable DOLL is among the top three important variables across all quantile levels. The importance of the other two low-frequency variables changes across quantile levels, but it is always positive. 
Concerning the dynamic MIDAS-QRF instead, the most important variable at all quantile level  is $lag\_quant$, highlighting the relevant role of the dynamic component in training the proposed model. Similarly to the static MIDAS-QRF, the most important low-frequency variable is DOLL, whereas the importance of the other two low-frequency variables remains positive at all quantile levels.

\vspace{0.15in}

\noindent As well as for the Heating Oil index, also for the other two indexes, all the macroeconomic variables and the $lag\_quant$ variable play a significant role in training the two versions of the proposed model. In particular, for the Brent index all three macroeconomic variables are among the most important variables both in the static and dynamic MIDAS-QRF, especially the DOLL one.

\vspace{0.15in}

\noindent In conclusion, these results highlight the relevant role of macroeconomic variables in capturing market risk at low quantile levels and in delivering accurate forecasts. The results also show that the forecast accuracy of the MIDAS-QRF can be further improved by adding a dynamic component. In this case, empirical results show that also the dynamic MIDAS-QRF outperforms benchmark models and passes all the backtesting procedures.

 \section{Conclusions}
 \label{sec:MIDAS-QRF-conclusions}

This chapter introduces a new model called MIDAS-QRF to compute conditional quantiles in a machine learning framework to jointly account for complexity, non-linearity and mixed-frequencies in data. The proposed model relies on the MIDAS methodology of \cite{ghysels2007midas}  to exploit information coming from low-frequency variables in order to compute quantiles through the QRF algorithm of \cite{meinshausen2006quantile}, which allows to infer the entire conditional distribution of the response variable through the Random Forest algorithm \citep{breiman2001random}. 

\vspace{0.15in}

\noindent The benefits of this approach are twofold: first, it allows to model non-linear relations among variables without making any parametric assumption and to study the behaviour of the phenomena on the tails. In this sense, the MIDAS-QRF results particularly useful in settings characterised by complex relationships among variables, such as the financial and economics one.
Second, it extends the QRF algorithm by considering mixed-frequency data by introducing a feature that has never been considered in in Random Forests algorithms. 
Moreover, the employment of the PCA on the low-frequency variables to compute the MIDAS components offers a novel perspective on estimating the $\omega_2$ parameter of the MIDAS model while shrinking the computational burden of the MIDAS-QRF training.
\vspace{0.15in}

\noindent The proposed model is applied to a real financial dataset to forecast the VaR of three commodities index. The empirical findings highlight the relevant role of macroeconomic variables in capturing market risk at low quantile levels to deliver accurate forecasts. The results also show that the forecast accuracy of a MIDAS-QRF model can be further improved by adding a dynamic component to model the dynamic distribution of financial variables. In this case, also the dynamic MIDAS-QRF outperforms benchmark models and passes all the backtesting procedures.

\begingroup
\let\cleardoublepage\clearpage
\endgroup

\begingroup
\let\cleardoublepage\clearpage
\endgroup

\begingroup
\let\cleardoublepage\clearpage
\endgroup

\chapter{Finite mixtures of Quantile Regression Forests and their application to GDP growth-at-risk from climate change} \label{ch:FM-QRF}
\chaptermark{Finite Mixtures of QRF}
\section{Introduction}
\label{sec:1}

Regression models target the expected value of the conditional distribution of the outcome given a set of covariates. When the distribution of the outcome is asymmetric, modelling other location parameters, e.g. percentiles of the conditional distribution, may offer a more complete picture of the outcome of the distribution compared with models describing only its centre. 

\vspace{0.15in}

\noindent The idea of modelling location parameters has a long history in statistics. The seminal paper of \cite{koenker1978regression} is regarded as the first detailed development of Quantile Regression (QR), which represents a generalization of median regression. QR approach is particularly useful when modeling data characterised by skewness, heavy tails, outliers, truncation, censoring and heteroschedasticity. The flexibility of this model allows for a wide range of applications in fields such as economics, finance, healthcare, environmental science, and marketing; for a detailed review of the most used QR techniques, see \cite{koenker2005quantile} and \cite{koenker2017handbook}. 

\vspace{0.15in}

\noindent Recently, extensions of QR modelling have been proposed in high dimensional framework and in a non-parametric context
including the QR Neural Networks \citep{white1992nonparametric}, the QR Support Vector Machines \citep{hwang2005simple, xu2015weighted} the QR kernel based algorithms \citep{christmann2008consistency} the QR Forests (QRF) \citep{meinshausen2006quantile} and the Generalized QR Forests 
\citep{ athey2019generalized}.

\vspace{0.15in}

\noindent Empirical applications often entail dependent observations in the form of hierarchically structured data: this is the case of spatial, multilevel or longitudinal sample designs. In these contexts, when regression models are considered, the potential association between dependent observations should be taken into account in order to provide valid and efficient inferences. This is often achieved by using random of mixed effects model where subject-specific random effects are considered in the linear predictors.

\vspace{0.15in}

\noindent Throughout the statistical and econometrics literature only recently random effects models have been used to capture the dependence from a QR perspective for clustered, multilevel, spatial and repeated measurements. Applications in fields such as medicine, environmental science, finance and economics have been particularly studied: for a detailed review of these techniques see for example \citet{farcomeni2012quantile, smith2015multilevel, alfo2017finite, marino2018mixed, merlo2021forecasting, hendricks1992hierarchical,pandey1999comparative, reich2011bayesian, bassett2002portfolio, kozumi2011gibbs, bernardi2015bayesian, bernardi2018bayesian,merlo2022quantile,merlo2022quantilets}.

\vspace{0.15in}

\noindent In these models, individual-specific random effects (coefficients) result useful to describe the influence of omitted covariates on parameter estimates of the observed ones. In this context, the random effects (coefficients) are often thought of as representing (individual-specific) unobserved heterogeneity, while observed covariates describe the observed counterpart (the fixed part of the models). However, in some applications a parametric assumption on the fixed part of the model may not be appropriate, leading to inaccurate conclusions regarding the phenomenon of interest. For this reason, QR machine learning models, incorporating random effects, may be more suitable to model complex and non-linear relationships among the response variable and the covariates in case of hierarchically structured data.


\vspace{0.15in}

\noindent Only recently random effects machine learning algorithms have been introduced to model dependence in a standard regression framework; see, for example, Mixed-Effects Neural Networks \citep{xiong2019mixed}, Mixed-Effects Support Vector Machines \citep{luts2012mixed} and Mixed-Effects Random Forests \citep{hajjem2014mixed, hajjem2011mixed, sela2012re}). These models are inadequate when the main interest of the research is the conditional quantiles of the distribution of the outcome. 

\vspace{0.15in}

\noindent This chapter aims at filling this gap by introducing a new model called Finite Mixtures of Quantile Regression Forests (FM-QRF). The aim of the proposed methodology is to build a data-driven model to estimate quantiles of longitudinal data in a  non-parametric framework. To this end, the QR finite mixtures approach of \citep{merlo2022two, tian2014linear, tian2016class, alfo2017finite} is extended to the machine learning realm. The FM-QRF is based on well known random effects machine learning algorithms, but leaves the random effects distribution unspecified and estimates the fixed part of the model with a QRF. 
\vspace{0.15in}

\noindent The quantile estimates are obtained with an iterative procedure based on the Expectation Maximization-type algorithm (EM) using the Asymmetric Laplace distribution (AL) as working likelihood.
The suggested methodology may be considered as an extension to a non-linear and non-parametric framework the work of \cite{geraci2007quantile, geraci2014linear, alfo2017finite} as well as an extension to a QR framework of the  mixed-modeling approach presented in \cite{hajjem2014mixed}.

\vspace{0.15in}

\noindent The FM-QRF performance is tested with a large scale simulation study and its behaviour is compared with a set of competitor models.
The FM-QRF is empirically applied to a longitudinal dataset to assess the effects of climate-change on the distribution of future growth of GDP of 210 worldwide countries. One of the first contributions in investigating the effects of climate change on GDP growth is by
\cite{kiley2021growth}, based on the concept of Growth-at-risk (GaR) introduced in \cite{yao2001measuring}. GaR is represented by the lower quantiles of the GDP growth and measures the expected maximum economic downturn given a probability level over a certain time-period. 
\vspace{0.15in}

\noindent The aim of this chapter is to extend the findings of \cite{kiley2021growth} by employing the FM-QRF to unveil non-linear complex relationship among GDP growth and climate-related variables in a mixed-effects framework. In particular, the long-term estimate of the future GaR is obtained by considering covariates related to temperature and precipitations. 
Consistent with prior literature, the findings shown in this chapter indicate that unsustainable climate practices will have adverse impacts on most of the countries, with significant heterogeneous effects among them. It is also found that temperature and precipitations differently affect upper and lower quantiles of the GDP growth conditional distribution, and that, in contrast to previous findings based on linear approaches, precipitations also play a relevant role in affecting its tails, especially in the upper quantiles.

\vspace{0.15in}

\noindent The rest of the chapter is organized as follows. Section \ref{sec:FM-QRF-methodology} describes the methodology of the FM-QRF. Section \ref{sec:FM-QRF-simulation} shows the results of the simulation study. Section \ref{sec:FM-QRF-empirical} shows the results of the case study on climate-change related data and Section \ref{sec:FM-QRF-conclusions} contains concluding remarks and outlines possible future research agenda.

\section{Methodology}
\label{sec:FM-QRF-methodology}
This chapter introduces the FM-QRF and provides a detailed explanation of the EM algorithm used to train the proposed model.

\subsection{Finite Mixtures of Quantile Regression Forest}

Let $Y_{it}, i=1,\dots,N$, $t=1,\dots,T_i$ be the response variable for the $i$-th statistical unit observed at time $t$. Let $y_{it}$ be the realisation of the outcome variable and denote with $\mathbf{x}_{it} \in \mathbb{R}^p$ the vector of observed explanatory variables with components $x_{it,j}, j=1, \dots, p$ and $x_{it,1} \equiv 1$. For a given quantile level $\tau \in (0, 1)$, in a longitudinal and clustered data setting, denote $\mathbf{b}_{\tau}=\{b_{1, \tau},\dots,b_{N, \tau}\}$ the vector of unit- and quantile-specific random coefficients. The response $Y_{it}$ is assumed to follow an Asymmetric Laplace density \citep[ALD - ][]{yu2001bayesian}:
\begin{equation}\label{eq:ald}
f_{y|b}(y_{it}|b_{i, \tau}; \tau) = \frac{\tau(1- \tau)}{\sigma_\tau} \exp \Bigg\{ - \rho_\tau \left( \frac{y_{it} - \mu_{it,\tau}}{\sigma_\tau}\right) \Bigg\},
\end{equation}
where $\tau$, $\sigma_\tau > 0$ and $\mu_{it, \tau}$ represent, respectively, the skewness, the scale, and the location parameter of the distribution while the latter one is also the quantile of the conditional distribution. The function $\rho_\tau (u) = u (\tau - \boldsymbol{1}_{\{u < 0\}})$ represents the quantile loss function of \cite{koenker1978regression}.
As it is well known in the literature the ALD is used as a working model able to recast estimation of parameters for the linear QR model in a Maximum Likelihood framework. 

\vspace{0.15in}

\noindent In the QR framework, the location parameter $\mu_{it,\tau}$ is often modeled as:
\begin{equation}
    	\label{eq:me-lqmm}
     \mu_{it,\tau}=\mathbf{x}_{it}^{\prime}\bbeta_\tau+b_{i, \tau}  \;\;\;  \tau \in (0,1),
	\end{equation}
where the random effect $b_{i, \tau}$ is time-constant and varies across statistical units according to a distribution $f_b(\cdot)$ with support $\mathcal{B}$ where $E[b_{i, \tau}] = 0$ is used for parameter identifiability. Rather than specifying such a distribution parametrically as in, e.g.,\cite{geraci2007quantile}, here it is left unspecified and estimated directly from the observed data via a Non-Parametric Maximum Likelihood approach \cite[NPML - ][]{laird1978nonparametric}. Moreover, the model \eqref{eq:me-lqmm} is extended to a machine learning framework using a non-parametric unknown function $g_{\tau}(\mathbf{x}_{it})$ instead of the traditional linear one $\mathbf{x}_{it}^{\prime}\bbeta_\tau$.

\vspace{0.15in}

\noindent More in detail, the distribution $f_b(\cdot)$ is approximated by a discrete distribution on $K < N$ locations $\alpha_{k,\tau}$ so that:
\begin{equation}
\label{eq:bk_bi}
    {\alpha}_{k,\tau} \sim \sum_{k=1}^K \pi_{k,\tau}\delta_{{\alpha}_{k,\tau}},
\end{equation}
where the probability $\pi_{k,\tau}$ is defined as $\pi_{k,\tau} = \mathbb{P}({b}_{i, \tau} = {\alpha}_{k ,\tau})$ with $i=1,\dots, N$ and $k=1,\dots,K$ and $\delta_{{\alpha}_{k,\tau}}$ is a one-point distribution putting a unit mass at ${\alpha}_{k,\tau}$. Under this approach, for $b_{i, \tau}={\alpha}_{k,\tau}$, the location parameter of the ALD in equation \eqref{eq:me-lqmm} becomes
\begin{equation}
    	\label{eq:fmqrf}
     \mu_{itk,\tau}=g_{\tau}(\mathbf{x}_{it})+\alpha_{k, \tau}  \;\;\;  \tau \in (0,1),
	\end{equation}
where $g_\tau: \mathbb{R}^p \rightarrow \mathbb{R}$.
\vspace{0.15in}

\noindent The model likelihood becomes:
\begin{equation}\label{eq:llk2}
L({\boldsymbol{} \Phi_\tau}) = \prod_{i=1}^N  \sum_{k=1}^K \Bigg\{\prod_{t=1}^{T_i} f_{y|b}(y_{it}|b_{i, \tau}=\alpha_{k, \tau}; \tau)\Bigg\} \pi_{k, \tau},
\end{equation}
where ${\boldsymbol{ \Phi}_\tau} = \{ \sigma_\tau, g_{\tau}(\mathbf{x}_{it}), {\alpha}_{1,\tau}, \dots , {\alpha}_{K,\tau}, \pi_{1,\tau}, \dots , \pi_{K,\tau} \}$ is the vector of unknown parameters.

\vspace{0.15in}

\noindent It is worth highlighting that the proposed FM-QRF can be also considered as an extension of: (i) the QRF \citep{meinshausen2006quantile} because a random part is added to the non-parametric unknown function of QRF; (ii) the QR for longitudinal data based on latent Markov subject-specific parameters \citep{farcomeni:2012} because the linear fixed-part is replaced by the function $g_{\tau}(\mathbf{x}_{it})$. 
\vspace{0.15in}

\noindent Differently from previous contributions  \citep[see][]{geraci2007quantile, geraci2014linear,farcomeni2012quantile}, the FM-QRF relies on the NPML, which is based on a finite-mixture representation of the random part of model \eqref{eq:me-lqmm}. This approach does not require to make any a-priori assumption on the distribution of the random effects, making the FM-QRF a valid data-driven approach resistant to misspecification \citep{alfo2017finite, farcomeni2012quantile}.

\vspace{0.15in}

\noindent The simultaneous estimation of $g_{\tau}(\mathbf{x}_{it})$ and $b_{k, \tau}$ in a QR setting poses several challenges, and the next section reports the EM algorithm developed to estimate both $g_{\tau}(\mathbf{x}_{it})$ and the $\alpha_{k, \tau}$ of the proposed FM-QRF. 

\subsection{Parameters Estimation with the EM Algorithm}

Following previous contributions \citep{tian2014linear, merlo2022two, alfo2017finite}, the estimates $\widehat{g}_{\tau}(\mathbf{x}_{it})$ and $\widehat{b}_{k,\tau}$ of \eqref{eq:fmqrf} are computed iteratively: in the first step the $g_{\tau}(\mathbf{x}_{it})$ function is estimated using the algorithm developed for the QRF approach \citep{meinshausen2006quantile}, then, in the second step, given $\widehat{g}_{\tau}(\mathbf{x}_{it})$, the random-effects $b_{k, \tau}$s are obtained by maximising \eqref{eq:loglikmixture} with the EM algorithm.

\vspace{0.15in}

\noindent The complete data log-likelihood function employed in the EM algorithm is obtained starting from the finite mixture representation in \eqref{eq:llk2}, in which each observation $i$ can be considered as drawn from one of the $K$ locations. Let $w_{ik}$ be the indicator variable equal to $1$ if the $i$-th unit belongs to the $k$-th component of the finite mixture, and 0 otherwise. The EM algorithm treats as missing data the component membership $w_{ik}$. Thus the log-likelihood for the complete data is:

\begin{equation}
	\label{eq:loglikmixture}
	\ell_c(\boldsymbol{ \Phi}_\tau)=\sum_{i=1}^{N}\sum_{k=1}^{K}w_{ik, \tau}\left\{ \sum_{t=1}^{T_i} \log\left(f_{y|b}(y_{it}|b_{i, \tau}=\alpha_{k, \tau}; \tau)\right)+\log(\pi_{k,\tau}) \right\}.
\end{equation}


The algorithm for the estimation of the parameters in model \eqref{eq:fmqrf} is as it follows.
Firstly, the values of $\widehat{\alpha}_{k,\tau}^{(0)}, \hat{\sigma}^{(0)}_{\tau}, \hat{\pi}_{k, \tau}^{(0)}, \widehat{g}_{\tau}(\mathbf{x}_{it})^{(0)}$ are initialised. The value of $\widehat{g_{\tau}}(\mathbf{x}_{it})^{(0)}$ at level $\tau$ is estimated by means of the algorithm used for QRF, fitted with the training set $\mathcal{T}^{(0)}= \left\{\left(y_{it}, {\bf x}_{it}\right)\right\}_{\substack{i=1,\ldots, N\\ j=1,\ldots, T_i}}$ ignoring the clustered structure of the data. 

\vspace{0.15in}

\noindent Subsequently, given $\widehat{g}_{\tau}(\mathbf{x}_{it})^{(0)}$, in the \textbf{E-step} the estimates $\hat{w}_{ik, \tau}^{(r+1)}$, $\hat{\pi}_{k,\tau}^{(r+1)}$, $\widehat{g}_{\tau}(\mathbf{x}_{it})^{(r+1)}$ are computed. The estimate values $\hat{w}_{ik, \tau}^{(r+1)}$ and $\hat{\pi}_{k, \tau}$ are obtained with the following expressions:

\begin{equation}
	\label{eq:w_hat}
	\hat{w}_{ik,\tau}^{(r+1)}=\mathbb{E}[w_{ik,\tau}|\mu_{itk},\hat{\Phi}_\tau^{(r)}]=\frac{\prod_{t=1}^{T_i}f_{y|b}(y_{it}|b_{i, \tau}=\alpha_{k, \tau}; \tau)^{(r)}\hat{\pi}_{k, \tau}^{(r)}}{\sum_{l=1}^{K}\prod_{i=1}^{T_i}f_{y|b}(y_{it}|b_{i, \tau}=\alpha_{l, \tau}; \tau)^{(r)}\hat{\pi}_{l, \tau}^{(r)}},
\end{equation}
\begin{equation}
\label{eq:pi_hat}
\hat{\pi}_{k, \tau}^{(r+1)}=\frac{1}{N}\sum_{i=1}^N \hat{w}_{ik,\tau}^{(r+1)},
\end{equation}
where $f_{y|b}(y_{it}|b_{i, \tau}=\alpha_{k, \tau}; \tau)^{(r)}$ is the value of the ALD \eqref{eq:ald} at step $r$ related to the $t$-th measurement of the $i$-th statistical unit when considering the $k$ -th component of the finite mixture.

\vspace{0.15in}

\noindent Then, given $\hat{w}_{ik, \tau}^{(r+1)}$ and $\hat{\pi}_{k, \tau}$, $\widehat{g}_{\tau}(\mathbf{x}_{it})^{(r+1)}$ is updated as it follows. First, the quantity $y^{*(r+1)}_{it}=y_{it}-\widehat{\alpha}_{k, \tau}^{(r)}$ is computed, which represents the unknown function of the model at step $r+1$. The resulting training set $\mathcal{T}^{(r+1)}= \left\{\left(y_{it}^{*(r+1)}, {\bf x}_{it}\right)\right\}_{\substack{i=1,\ldots, N\\ j=1,\ldots, T_i}}$ is used to fit a QRF to estimate $\widehat{g}_{\tau}(\mathbf{x}_{it})^{(r+1)}$ at level $\tau$ accounting for the weight $\hat{w}_{ik, \tau}^{(r+1)}$ of each observation.

\vspace{0.15in}

\noindent Then, in the \textbf{M-step} the estimation of $\hat{\sigma}_{\tau}$ and $\widehat{\alpha}_{k, \tau}$ are obtained
by maximising the expectation $\mathbb{E}[\ell_c(\boldsymbol{\Phi}_\tau)| \mu_{itk}, \hat{\boldsymbol{\Phi}}_\tau^{(r)}]$ with respect to $\hat{\sigma}_{\tau}$ and $\widehat{\alpha}_{k, \tau}$ by numerical optimisation techniques. In particular, in this chapter the Nelder-Mead algorithm \citep{nelder1965simplex} has been implemented. Subsequently, $\widehat{\alpha}_{i, \tau}^{(r)}$ are estimated by means of \eqref{eq:bk_bi}.

\vspace{0.15in}

\noindent The E- and M-steps are alternated iteratively until convergence, that is reached when difference between the likelihood of two consecutive steps is smaller than a certain threshold. A schematic description of the algorithm is presented below.

\begin{figure}[H]
\centering
\begin{minipage}{0.87\textwidth}
\SetKwComment{Comment}{/* }{ */}
\begin{algorithm}[H]
\caption{Mixed-Effects Quantile Regression Forest}\label{tab:algorithm}
\KwData{For a fixed quantile level $\tau$, $\mathcal{T}=\left\{\left(y_{it}, {\bf x}_{it}\right)\right\}_{\substack{i=1,\ldots, N\\ j=1,\ldots, T_i}}$, ${\alpha}_{k, \tau} = \sum_{k=1}^K \pi_{k,\tau} \delta_{{b}_{k,\tau}}$.}
\KwResult{Quantile estimate $\hat{Q}^s_{\tau}(y_{it}|\mathbf{x}_{it})=\hat{g}_\tau(\mathbf{x}_{it})+\hat{\alpha}_{k,\tau}$}
\vspace{0.5cm}

$r \gets 0$ \;
$\widehat{\alpha}_{k,\tau}^{(r)} \gets 0$\;
$\hat{\sigma}_{\tau}^{(r)} \gets \sum_{s=1}^S \frac{1}{N}\sum_{i=1}^{N} \frac{1}{T_i}\sum_{t=1}^{T_i} \rho(u_{it})$\ where $u_{i,t}=y_{i,t}-\hat{Q}^s_{\tau}(y_{it}|\mathbf{x}_{it})$;
$\hat{\boldsymbol{\pi}}_{k, \tau}^{(r)} \gets K$ weights of a Gaussian quadrature\;
$\widehat{g}_{\tau}(\mathbf{x}_{it})^{(r)} \gets$ estimate of the unknown function at level $\tau$ obtained by fitting the QRF with $\mathcal{T}$\;

\vspace{0.5cm}
\textbf{E-step}: Update  estimates $\hat{w}_{ik, \tau}^{(r+1)}$ and $\widehat{g}_{\tau}(\mathbf{x}_{it})^{(r+1)}$:
\begin{itemize}
\item Update ${w_{ik, \tau}}^{(r+1)}$ with equation \eqref{eq:w_hat}
    \item Update $\widehat{g}_{\tau}(\mathbf{x}_{it})^{(r+1)}$ using a QRF fitted with  $\mathcal{T}^{(r+1)}=\left\{\left(y_{it}^{*(r+1)}, {\bf x}_{it}\right)\right\}_{\substack{i=1,\ldots, N\\ j=1,\ldots, T_i}}$ where $y^{*(r+1)}_{it}=y_{it}-\widehat{\alpha}_{k, \tau}^{(r)}$ and accounting for the weight $\hat{w}_{ik,\tau}^{(r+1)}$ of each observation.
\end{itemize} 

\vspace{0.5cm}
\textbf{M-step}: Update estimates $\widehat{\alpha}_{k, \tau}^{(r+1)}$ and $\hat{\sigma}_{\tau}^{(r+1)}$ by maximising \eqref{eq:loglikmixture}
\vspace{0.5cm} 

Keep iterating between \textbf{E-step} and \textbf{M-step} until convergence.

\end{algorithm}
\end{minipage}
\end{figure}

One of the main drawbacks of the traditional EM algorithm is that the M-step may be particularly burdensome in empirical applications involving a large set of covariates, in particular in terms of computational effort. Hence, in order to overcome this issue, the closed-form solutions approach used in \cite{merlo2022two} and \cite{tian2014linear} and reported in the next section is applied.

\subsection{Closed Form Solutions to the EM Algorithm}\label{sec:closedformsolutions}

As shown in \cite{merlo2022two} and \cite{tian2014linear}, closed form solutions to the EM algorithm are obtained considering the location-scale mixture representation of $Y_{it}$ presented in \cite{kozumi2011gibbs}:
\begin{equation}
\label{eq:hier}
    Y_{it}=\theta V_{it}+\tau \sqrt{V_{it}}Z_{it}
\end{equation}
where $V_{it}$ is an exponential random variable with realization $v_{it}$ and $Z_{it}$ a standard Normal random variable.

\vspace{0.15in}

\noindent In this setting, $f(v_{it, \tau} |y_{it}, \mu_{itk,\tau}, \sigma_{\tau})$ is a Generalized Inverse Gaussian (GIG) distribution \citep{tian2014linear, tian2016class}:\\

\begin{equation}
\label{eq:GIG}
f(v_{it, \tau} |    y_{it},  \mu_{itk,\tau}, \sigma_{\tau}) \sim \textnormal{GIG} \Bigg( \frac{1}{2}, \frac{(y_{it} - \mu_{itk, \tau})^2}{\rho^2 \sigma_{\tau}} , \frac{2\rho^2 + \theta^2}{\rho^2 \sigma_{\tau}} \Bigg).
\end{equation}
where $\theta = \frac{1-2\tau}{\tau (1-\tau)}$ and $\rho^2 = \frac{2}{\tau (1-\tau)}$.
Starting from \eqref{eq:hier} and \eqref{eq:GIG}, the complete data log-likelihood function of the EM algorithm is based on:

\begin{equation}\label{eq:ytildev}
f( y_{it}, v_{it, \tau} | \mu_{itk}, \sigma_{\tau})= \frac{1}{ \sigma_{\tau} \rho \sqrt{2 \pi \sigma_{\tau} v_{it, \tau}}}  \exp \bigg( - \frac{(   y_{it} - \mu_{itk, \tau} - \theta v_{it, \tau})^2}{2 \rho^2 \sigma_{\tau} v_{it, \tau}} -\frac{v_{it, \tau}}{\sigma_{\tau}} \bigg).
\end{equation}

From \eqref{eq:ytildev}, the complete data log-likelihood function is proportional to:

\begin{center}
\begin{align}
\begin{split}
& \ell_c ({\alpha}_{1, \tau}, ..., {\alpha}_{K, \tau}) \propto \frac{1}{2} \sum_{i=1}^N \sum_{k=1}^K \sum_{t=1}^{T_i}  w_{ik, \tau} v_{it, \tau}^{-1} (   y_{it} - g_{\tau}(\mathbf{x}_{it}) -  \alpha_{k, \tau})^2  \\
& - \theta \sum_{i=1}^N \sum_{k=1}^K \sum_{t=1}^{T_i}  w_{ik, \tau} (   y_{it} - g_\tau(\mathbf{x}_{it}) -  \alpha_{k, \tau}).
\label{eq:complhierc}
\end{split}
\end{align}
\end{center}

Estimates $\widehat{{\alpha}}_{k, \tau}$ are obtained by alternating between the E-step and the M-step.

\vspace{0.15in}

\noindent In the \textbf{E-step} the expectation of \eqref{eq:complhierc} conditional to the observed data is computed and it is proportional to:

\begin{center}
\begin{align}
\begin{split}
&\mathbb{E}[ \ell_c ({\alpha}_{1, \tau}, ..., {\alpha}_{K, \tau}) \mid {\mu_{it}, }\hat{{\boldsymbol{\Phi}}}_{\tau}^{(r)}] \propto  \\
& \frac{1}{2} \sum_{i=1}^N \sum_{k=1}^K \sum_{t=1}^{T_i} \hat{w}_{ik}^{(r+1)} \hat{v}_{it, \tau}^{(r+1)} (   y_{it} - \hat{g}_{\tau}(\mathbf{x}_{it}) -  \alpha_{k, \tau})^2 \label{eq:ecdl21} \\ 
& - \theta \sum_{i=1}^N \sum_{k=1}^K \sum_{t=1}^{T_i}  \hat{w}_{ik, \tau}^{(r+1)}  (   y_{it} - \hat{g}_{\tau}(\mathbf{x}_{it}) -  {\alpha}_{k, \tau}), 
\end{split}
\end{align}
\label{eq:ecdl22}
\end{center}

where $\hat{w}_{ik, \tau}^{(r+1)}$ is obtained as in \eqref{eq:w_hat} and the unknown function part ${g}_\tau(\mathbf{x}_{it})$ is estimated using the QRF algorithm.
 
\vspace{0.15in}

\noindent The estimates of the latent variable $\hat{v}_{it, \tau}^{(r+1)} = \mathbb{E}[V_{it, \tau}^{-1} \mid \mu_{itk}, \hat{{\boldsymbol{\Phi}}}_{\tau}^{(r)}]$ are obtained by exploiting the moment properties of the GIG distribution in \eqref{eq:GIG}:

\begin{center}
\begin{equation}\label{eq:mominv}
\hat{v}_{it, \tau}^{(r+1)} = \mathbb{E}[V_{it, \tau}^{-1} \mid \mu_{itk}, \hat{{\boldsymbol{\Phi}}}_{\tau}^{(r)}] = \frac{\sqrt{\theta^2 + 2\rho^2}}{\mid    y_{it} -   \widehat{g}_\tau(\mathbf{x}_{it})^{(r)} -  {\hat{\alpha}}_{k, \tau}^{(r)} \mid}.
\end{equation}
\end{center}

In the \textbf{M-step}, \eqref{eq:ecdl22} is maximised with respect to ${\alpha}_{1, \tau}, \dots, {\alpha}_{K, \tau}$ and the following update expression for $\widehat{{\alpha}}_{k, \tau}^{(r+1)}$ is obtained: 

\begin{center}
\begin{equation}\label{eq:b}
\widehat{{\alpha}}_{k, \tau}^{(r+1)} = \sum_{i=1}^N \sum_{t=1}^{T_i} \frac{ \hat{w}_{ik, \tau}^{(r+1)} \hat{v}_{it, \tau}^{(r+1)}  }{   \hat{w}_{ik, \tau}^{(r+1)} \big( \hat{v}_{it, \tau}^{(r+1)}  (   y_{it} -   \widehat{g}_{\tau}(\mathbf{x}_{it})^{(r)}) - \theta  \big) }.
\end{equation}
\end{center}

\section{Simulation Study}\label{sec:FM-QRF-simulation}

This section reports the results of a simulation study carried out to assess the performance of the FM-QRF. The proposed model is tested in a non-linear setting characterised by non-neglegible clustering-effects. The FM-QRF is used to predict quantiles at levels $\tau \in \{0.1, 0.5, 0.9\}$.

\vspace{0.15in}

\noindent The data are simulated under the following non-linear data generating process (DGP) \citep{hajjem2014mixed}:
$$
 y_{it} = g(\mathbf{x}_{it}) + b_i + \varepsilon_{it}
 $$
$$g(\mathbf{x}_{it}) = 2 x_{it,1} + x_{it,2}^2 + 4\cdot\mathbf{1}_{\{x_{it,3} > 0\}} + 2 x_{it,3} \log |x_{it,1}|$$

\vspace{0.15in}

\noindent The covariates are generated as $x_{it, 1}, \; x_{it,2}, \; x_{it,3} \sim \mathcal{N}(0,1)$. The random-effects parameters and the error terms are generated independently according to four DGPs with small and large proportion of random-effects variance (PREV, computed as in \cite{hajjem2014mixed}). DGPs with a large PREV are characterised by the presence of a larger proportion of total variance explained by the random effects, implying that the clustered structure of the data is more pronounced:

\begin{enumerate}
    \item \textbf{(NN-S)} $b_i \sim N(0,1), \;\; \varepsilon_{it} \sim N(0,1) $ with small PREV

     \item \textbf{(NN-L)} $b_i \sim N(0,1), \;\; \varepsilon_{it} \sim N(0,1) $ with large PREV

      \item \textbf{(TT-S)} $b_i \sim t(3), \;\; \varepsilon_{it} \sim t(3) $ with small PREV

      \item \textbf{(TT-L)} $b_i \sim t(3), \;\; \varepsilon_{it} \sim t(3) $ with large PREV
\end{enumerate}
 
 Under scenarios NN-S and NN-L the assumptions of the linear quantile mixed model (LQMM) of Gaussian random parameters hold, whereas for scenarios TT-S and TT-L these hypotheses are violated. In particular, a DGP with heavier tails represented by a Student's \textit{t} distribution with three degrees of freedom is assumed.

\vspace{0.15in}

\noindent As in \cite{hajjem2011mixed}, for each scenario are considered a training set of 500 observation for $N=100$ statistical units and $T_i=5$ measurements each, and an unbalanced test set with $T_i \in \{9, 27, 45, 63, 81\}$ for a total of 4500 observations.
Each scenario has been replicated $S = 100$ times.

\vspace{0.15in}

\noindent The average performance of the ME-QRF across the 100 replications is assessed on the test set in terms of the following three loss functions:

\begin{itemize}

\item Average Mean Absolute Error (MAE) and average Mean Squared Error (MSE) with respect to the theoretical quantile of the DGP, computed as in \cite{min2004monte}:

\begin{center}
    \begin{equation}
        MAE_{\tau}= \frac{1}{S} \sum_{s=1}^S \frac{1}{N}\sum_{i=1}^{N} \frac{1}{T_i}\sum_{t=1}^{T_i} |Q^s_{\tau}(y_{it}|\mathbf{x}_{it})-\hat{Q}^s_{\tau}(y_{it}|\mathbf{x}_{it})|
    \end{equation}
\end{center}
\begin{center}
    \begin{equation}
        MSE_{\tau}= \frac{1}{S} \sum_{s=1}^S \frac{1}{N}\sum_{i=1}^{N} \frac{1}{T_i}\sum_{t=1}^{T_i} (Q^s_{\tau}(y_{it}|\mathbf{x}_{it})-\hat{Q}^s_{\tau}(y_{it}|\mathbf{x}_{it}))^2
    \end{equation}
\end{center}

where $Q^s_{\tau}(y_{it}|\mathbf{x}_{it})$ and $\hat{Q}^s_{\tau}(y_{it}|\mathbf{x}_{it}) = \mu_{it, \tau}$ are the the theoretical and estimated conditional quantiles at level $\tau$ of the $s$-th simulated dataset, respectively.



    
    \item Average Ramp loss (RAMP) as proposed in \cite{takeuchi2006nonparametric}. This loss is used to measure the quantile property of estimator $\hat{Q}^s_{\tau}(y_{it}|\mathbf{x}_{it})$ in dividing the data so that $\tau$ percent of observations fall below $\hat{Q}^s_{\tau}(y_{it}|\mathbf{x}_{it})$ and $1-\tau$ are above:

    \begin{equation}
       RAMP_\tau= \frac{1}{S} \sum_{s=1}^S \frac{1}{N}\sum_{i=1}^{N} \frac{1}{T_i}\sum_{t=1}^{T_i}  \boldsymbol{1}_{\{u_{it} < 0\}}
    \end{equation}

where $u_{it}=y_{it}-\hat{Q}^s_{\tau}(y_{it}|\mathbf{x}_{it})$. The model satisfies the quantile property for ramp loss values close to $\tau$.
    
\end{itemize}

The performance in term of losses of the FM-QRF is compared with three benchmark models: LQMM, Quantile Regression Forest (QRF) and the Quantile Mixed Model (QMM) of \cite{merlo2022two} adapted to a no two-part model. The latter model exploits the same methodological approach of the FM-QRF in a linear setting. The number of mixture components has been set to $K=11$ with a grid search approach for all models, and the hyper-parameters of the FM-QRF (that is, number of trees, minimum number of observations in terminal nodes, number of features to consider for splitting nodes)   have been optimized by Bayesian Optimization using the \texttt{ParallelBayesOptQRF} package implemented in R \footnote{https://github.com/mila-andreani/ParallelBayesOptQRF}. The QRF has been trained by using the same hyper-parameters settings of the FM-QRF. 

\vspace{0.15in}

\noindent Results are reported in Table \ref{tab:simres}. 

\begin{table}[H]
\small
\begin{center}
{\tabcolsep=7pt\def\arraystretch{0.75}
\begin{tabular*}{\textwidth}{@{} l @{\extracolsep{\fill}}@{}cccclccc@{}}
 \midrule
\multicolumn{1}{l}{} & \textbf{} & \textbf{NN-S} &  &  &  & \textbf{NN-L} \\ \midrule
\multicolumn{1}{l}{} & \textbf{MAE} & \textbf{MSE} & \textbf{RAMP} &  & \textbf{MAE} & \textbf{MSE} & \textbf{RAMP} \\ \cmidrule(lr){2-4} \cmidrule(l){6-8} 
\multicolumn{1}{l}{$\tau=0.1$} & \multicolumn{1}{l}{} & \multicolumn{1}{l}{} & \multicolumn{1}{l}{} &  & \multicolumn{1}{l}{} & \multicolumn{1}{l}{}  \\
\textbf{FM-QRF} & 1.54 & 4.93 & 0.1 &  & \textbf{1.83} & \textbf{5.87} & 0.1 \\
\textbf{LQMM} & 2.45 & 10.20 & 0.1 &  & 1.86 & 5.89 & 0.1 \\
\textbf{QRF} & \textbf{1.50} & \textbf{4.49} & 0.1 &  & 2.64 & 10.98 & 0.1 \\
\textbf{QMM} & 4.38 & 35.34 & 0.2 &  & 4.67 & 40.62 & 0.2 \\
$\tau=0.5$ &  &  &  &  &  &  &  \\
\textbf{FM-QRF} & 1.56 & 4.41 & 0.5 &  & \textbf{1.65} & \textbf{4.62} & 0.5 \\
\textbf{LQMM} & 2.09 & 8.11 & 0.5 &  & 1.72 & 5.27 & 0.5 \\
\textbf{QRF} & \textbf{1.46} & \textbf{3.72} & 0.5 &  & 2.11 & 7.15 & 0.5 \\
\textbf{QMM} & 2.33 & 10.36 & 0.5 &  & 2.80 & 14.61 & 0.5 \\
$\tau=0.9$ &  &  &  &  &  &  &  \\
\textbf{FM-QRF} & 1.49 & 4.46 & 0.9 &  & 1.67 & 4.86 & 0.9 \\
\textbf{LQMM} & 2.15 & 7.90 & 0.9 &  & \textbf{1.57} & \textbf{4.34} & 0.9 \\
\textbf{QRF} & \textbf{1.32} & \textbf{3.57} & 0.9 &  & 2.20 & 7.93 & 0.9 \\
\textbf{QMM} & 6.39 & 79.14 & 0.7 &  & 5.16 & 61.12 & 0.8 \\ \midrule
\multicolumn{1}{l}{} & \multicolumn{1}{l}{} & \multicolumn{1}{l}{} & \multicolumn{1}{l}{} &  & \multicolumn{1}{l}{} & \multicolumn{1}{l}{} & \multicolumn{1}{l}{} \\
\multicolumn{1}{l}{} & \multicolumn{1}{l}{} & \multicolumn{1}{l}{} & \multicolumn{1}{l}{} &  & \multicolumn{1}{l}{} & \multicolumn{1}{l}{} & \multicolumn{1}{l}{} \\ \midrule
\multicolumn{1}{l}{} & \textbf{} & \textbf{TT-S} &  &  & \textbf{} & \textbf{TT-L}  \\ \cmidrule(lr){2-4} \cmidrule(l){6-8} 
\multicolumn{1}{l}{} & \textbf{MAE} & \textbf{MSE} & \textbf{RAMP} &  & \textbf{MAE} & \textbf{MSE} & \textbf{RAMP}  \\ \cmidrule(lr){2-4} \cmidrule(l){6-8} 
\multicolumn{1}{l}{$\tau=0.1$} & \multicolumn{1}{l}{} & \multicolumn{1}{l}{} & \multicolumn{1}{l}{} &  & \multicolumn{1}{l}{} & \multicolumn{1}{l}{} & \multicolumn{1}{l}{} \\
\textbf{FM-QRF} & \textbf{2.11} & \textbf{9.03} & 0.1 &  & \textbf{1.80} & 6.66 & 0.1 \\
\textbf{LQMM} & 2.57 & 11.09 & 0.1 &  & 1.94 & \textbf{6.42} & 0.1 \\
\textbf{QRF} & 2.22 & 11.09 & 0.1 &  & 2.02 & 7.88 & 0.1 \\
\textbf{QMM} & 4.94 & 49.44 & 0.2 &  & 4.11 & 33.16 & 0.2 \\
$\tau=0.5$ &  &  &  &  &  &  &  \\
\textbf{FM-QRF} & 1.62 & 5.31 & 0.5 &  & \textbf{1.43} & \textbf{4.32} & 0.5 \\
\textbf{LQMM} & 2.02 & 7.45 & 0.5 &  & 1.57 & 4.48 & 0.5 \\
\textbf{QRF} & \textbf{1.55} & \textbf{4.93} & 0.5 &  & 1.44 & 4.51 & 0.5 \\
\textbf{QMM} & 2.51 & 12.16 & 0.5 &  & 2.01 & 7.82 & 0.5 \\
$\tau=0.9$ &  &  &  &  &  &  &  \\
\textbf{FM-QRF} & \textbf{2.11} & \textbf{8.64} & 0.9 &  & \textbf{1.87} & 7.10 & 0.9 \\
\textbf{LQMM} & 2.71 & 11.53 & 0.9 &  & 2.06 & \textbf{6.81} & 0.9 \\
\textbf{QRF} & 2.22 & 9.23 & 0.9 &  & 2.06 & 8.02 & 0.9 \\
\textbf{QMM} & 7.91 & 218.196 & 0.8 &  & 5.13 & 50.79 & 0.8 \\ \bottomrule
\end{tabular*}}
\caption{Loss values for each scenario computed on the test set of the four fitted models. Bold values indicate the smallest loss.}
\label{tab:simres}
\end{center}
\end{table}

\noindent In the first scenario, in which data meet the Gaussianity assumptions of the LQMM model, the FM-QRF performance mainly depends on the relevance of the clustering effect in the simulated data. 

\vspace{0.15in}

\noindent When the clustering effect is small, the best performing model is the QRF algorithm. This is due to the fact that, in this setting, the QRF is more well specified than the LQMM, which is a linear model, and the FM-QRF model, which is designed for non-linear setting in which the clustering effect is large.
As a matter of fact, when the clustering effect is more relevant, the best performing model is the FM-QRF for almost all quantile levels.

\vspace{0.15in}

\noindent In the second scenario, the simulated data violate the gaussianity assumptions of the LQMM. 
In this case, the FM-QRF outperforms the benchmark models both when the clustering effect is small and large for almost all quantile levels.

\vspace{0.15in}

\noindent The only exception is represented by the performance of the FM-QRF in terms of MSE in the large clustering effect setting when $\tau=0.01, 0.9$.
In this case, a larger training set could lead to an improved performance of the FM-QRF in terms of MSE in predicting extreme quantile values.

\vspace{0.15in}

\noindent In conclusion, the main finding in this simulation study is that the performance of the FM-QRF improves as the clustering effect increases, and when data violate the gaussianity assumptions of the LQMM model. 


\section{Empirical Application}\label{sec:FM-QRF-empirical}

The study of the effects of climate change is a widely discussed topic in several fields, and a variety of studies have highlighted the relevant and multifaceted role of the growing frequency and intensity of natural disasters in determining economic output of countries \citep{mele2021nature, tol2018economic,palagi2022climate, peng2011toward, deschenes2007economic, dell2014we, dell2012temperature, weitzman2014fat, barro2015environmental, fankhauser2005climate}. 
\vspace{0.15in}

\noindent 
Extreme weather events, including floods, droughts, and escalating temperatures, exert direct influence on the economic output of countries, especially those heavily relying on the agricultural sector. This impact primarily stems from infrastructural damages and fluctuations in crop yields and livestock productivity, resulting in significant economic losses \citep{nelson2014climate, aydinalp2008effects, orlov2021global}.
Other studies have shown that climate change affects economic output of countries also indirectly, by altering migratory flows \citep{cattaneo2016migration, marotzke2020economic}, demography \citep{barreca2015convergence}, criminality \citep{burke2018higher} productivity and labor supply \citep{somanathan2021impact, heal2016reflections, graff2014temperature}, energy production and consumption \citep{burke2015global, burke2018large}.
These results highlight the relevance of considering climate change related risks in macroeconomic policy analyses, although evaluating the economic impact of climate change is a quite difficult task. A variety of statistical and econometric models has been proposed in literature {\citep{ kolstad2020estimating, carleton2016social, mendelsohn2006distributional, coronese2018natural, coronese2019evidence, hsiang2016climate}; in particular, several studies have exploited standard regression approaches to analyse the relation between climate-related variables and economic aggregates, such as GDP and GDP growth; see for example \cite{hsiang2016climate, dell2012temperature, kahn2021long, burke2015global, burke2018large}.
\vspace{0.15in}

\noindent Nevertheless, the effects of climate change may extend beyond the typical GDP distribution, potentially heightening the vulnerability to economic downturns, as evidenced by the lower tails of GDP growth distribution, and amplifying systemic risks through the intersection of climate change impacts and human systems \citep[e.g. international food markets, international security and countries' economic arrangements][]{king2017climate}.

\vspace{0.15in}

\noindent In order to uncover vulnerabilities and potential systemic risks overlooked by standard regression approaches, the focus of more recent analyses shifted from measures of central tendency of the GDP growth distribution \citep{kahn2021long, dell2012temperature, burke2015global, burke2018large, kalkuhl2020impact}, towards its upper and lower tails \citep{kiley2021growth, yao2001measuring, coronese2018natural, coronese2019evidence}. Indeed, analysing the relation between climate change and the tails of GDP growth provides useful information to enhance policy effectiveness, especially when the policy maker aims at preventing and mitigating the impact of extreme events, such as economic downturns or recessions, caused by climate change. Moreover, information on the tails of GDP growth can be used as a measure of economic resilience, offering insights into how well an economy can endure and recover from extreme events.
\vspace{0.15in}

\noindent One of the most used risk measures concerning the tails of the GDP growth distribution is the GDP Growth-at-Risk (GaR) \citep{yao2001measuring, adrian2022term, adrian2019vulnerable}, representing the expected maximum economic downturn given a probability level over a certain time-period. GaR may be estimated in a QR framework as the conditional quantile of the GDP growth distribution at 1\% or 5\% quantile level. 

\vspace{0.15in}

\noindent In the recent work of \cite{kahn2021long}, a linear QR parametric approach is used to predict the effects of different climate-change scenarios on future GaR of a basket of countries. The results highlight that unsustainable climate practices will have a negative effect on the GaR on the majority of countries included in the sample, increasing their risk of experiencing an extreme economic downturn.

\vspace{0.15in}

\noindent Given the relevance of the approach of \cite{kahn2021long}, the aim of this chapter is to extend their findings to a non-linear QR setting by applying the proposed FM-QRF to an unbalanced panel of 3045 country-year observations from 1995 to 2015 for 210 countries.
The outcome variable is the first difference of the natural logarithm of the yearly GDP per capita. The yearly covariates set includes the current value of: Temperature (\textit{TMP}), Precipitation (\textit{PRE}), Magnitude of precipitations greater than 20mm (\textit{r20mm}), maximum number of Consecutive Wet Days (\textit{cwd}), maximum number of Consecutive Dry Days (\textit{cdd}) and Maximum temperature of the year (\textit{txx}). All variables have been differentiated.
Also four lags of each covariate have been included in the covariates set for a total of 30 predictors.

The GDP data have been retrieved from the World Bank\footnote{https://data.worldbank.org}, and the covariates observations represent the historical values of the projected variables considered in the CMIP6 dataset\footnote{https://climateknowledgeportal.worldbank.org}. The final dataset and the \texttt{R} code used to create the dataset is available on Github \footnote{https://github.com/mila-andreani/climate-change-dataset}.
The summary statistics of the variables included in the sample are reported in Table \ref{tab:summary-FM-QRF}.

\begin{table}[]
\resizebox{\textwidth}{!}{%
\small
\begin{tabularx}{\textwidth}{*{7}{>{\centering\arraybackslash}X}}
\toprule
      & \textbf{Obs} & \textbf{Min} & \textbf{Max} & \textbf{Median} & \textbf{Mean} & \textbf{SD} \\ \midrule
\textit{GDP}   & 2782         & -0.70        & 0.63         & 0.04            & 0.04          & 0.05        \\
\textit{TMP}   & 2782         & -0.64        & 0.81         & 0.02            & 0.03          & 0.12        \\
\textit{PRE}   & 2782         & -496.06      & 484.47       & -0.08           & -0.23         & 64.17       \\
\textit{r20mm} & 2782         & -10.00       & 7.01         & 0.00            & 0.01          & 1.16        \\
\textit{cwd}  & 2782         & -22.94       & 18.00        & 0.00            & 0.00          & 2.88        \\
\textit{cdd}   & 2782         & -40.21       & 51.47        & 0.01            & 0.04          & 5.51        \\
\textit{txx}   & 2782         & -2.64        & 2.13         & 0.03            & 0.04          & 0.35        \\ \bottomrule
\end{tabularx}%
}
\caption{Summary statistics of the variables included in the sample.
The table reports the number of observations (Obs), the minimum (Min),
maximum (Max) along with the median, mean and standard deviation (SD).
}
\label{tab:summary-FM-QRF}
\end{table}

\vspace{0.15in}

\noindent The empirical analysis presented here is developed in two parts.
In the first subsection (\ref{sec:4.1}), the validity of a non-linear QR approach over a standard linear one is evaluated by means of an analysis of variance (ANOVA) test \citep{st1989analysis}, and the Variable Importance of each covariate is analysed to measure its relevance in predicting GDP growth quantiles.
\vspace{0.15in}

\noindent The second subsection (\ref{sec:4.2}) focuses on the application of the FM-QRF model to forecast quantiles of the GDP growth at five different probability levels $\tau= 0.01, 0.05, 0.5, 0.95, 0.99$  at three horizons $t=2030,2050,2100$.
The conditional lower quantile levels $0.01$ and $0.05$ measure the downside risk of GDP growth (GaR). This risk measure represents the maximum probable loss in GDP growth caused by climate change. The upper two conditional quantiles levels $0.95$ and $0.99$ represent the upside risk of GDP growth, measuring the maximum probable growth of GDP conditional on the covariates set.
\vspace{0.15in}

\noindent The conditional quantiles forecasts at different $\tau$ are computed using the projected values of the covariates from the CMIP6 dataset \citep{o2016scenario,li2021changes}. In particular, the effects of climate change are measured by comparing two alternative climate scenarios formulated in the CMIP6 experiments.
Such scenarios are denoted as Shared Socio-economic Pathways (SSPs) \citep{riahi2017shared}, and each SSP outlines the progression of climate variables conditional on the future trajectory of socio-economic factors and climate policies implemented by individual countries up to the year 2100. Five different SSP are available, and in this chapter two different SSPs are considered: SSP1 \citep{van2017energy} and SSP5 \citep{kriegler2017fossil}, which are two opposite scenarios of climate change.  
In SSP5 a future energy-intensive economy based on fossil fuels is hypothesised, whereas in SSP1 climate sustainable practices will prevail. 
Thus, the SSP5 represents the worst-case scenario in terms of climate change, and SSP1 represents the best-case one. 

\vspace{0.15in}

\noindent The aim of this analysis is to evaluate whether the fossil fuel-based economy of the SSP5 scenario will negatively affect the future GDP growth distribution at different quantile levels.
The results provide evidence that unsustainable climate policies (SSP5 scenario) will increase the GaR of the majority of the countries included in the considered sample. Moreover, differently from results obtained in the related literature \cite{dell2012temperature}, and as recently demonstrated in \cite{damania2020does}, this analysis shows that changes in precipitations-related variables due to climate change also represent a relevant risk factor undermining future economic growth.

\subsection{Preliminary Analysis}\label{sec:4.1}

This section reports the results of the preliminary analysis on the dataset of interest. Within the machine learning framework of this chapter, first the validity of a non-linear QR approach over a standard linear one is evaluated with an analysis of variance (ANOVA) test \citep{st1989analysis} to compare a spline QR model \citep{marsh2001spline, koenker1994quantile, wang1998mixed} and standard QR at five quantile levels $\tau=0.01, 0.05, 0.5, 0.95, 0.99$. 

\vspace{0.15in}

\noindent The spline QR model is an extension of the standard QR model based on piecewise-defined polynomial functions. Differently from the linear QR modeling framework, the splines model adapts to non-linear relationships with a series of knot points, where each different polynomial segment originates.
This feature makes this model particularly suitable to model complex and non-linear relationships among the variables of interest.
\vspace{0.15in}

\noindent In this section, both the  QR spline model and the linear one consider the same covariates of the FM-QRF. In particular, the QR spline model is represented by a piecewise cubic polynomial with 2 knots for each covariate.
The results of the ANOVA test, in Table \ref{tab:anova}, reveal that a non-linear approach, such as the one in FM-QRF, results more valid than the linear one.

\begin{table}[h]
    \centering
    \caption{ANOVA test results for different values of $\tau$. The 'Tn' column reports the test statistic and the 'P-Value' column reports the level of significance of the test at 5\% significance level. }
    \label{tab:anova}
    \begin{tabularx}{\textwidth}{*{6}{>{\centering\arraybackslash}X}}
        \hline
        \textbf{$\tau$} & \textbf{0.1} & \textbf{0.25} & \textbf{0.5} & \textbf{0.75} & \textbf{0.9} \\ \hline
        \textit{Tn}  & 1.513                                              & 3.585                                              & 10.686                                             & 27.329                                             & 2.2                                                \\
\textit{P-Value}                                              & 0.001**                                            & 0***                                               & 0***                                               & 0***                                               & 0***                                               \\ \hline
    \end{tabularx}
\end{table}

\vspace{0.15in}

\noindent In the second part of the preliminary analysis the relevance of each covariate in predicting the quantiles of the GDP growth distribution is assessed by extracting the Variable Importance measure from the FM-QRF.   

\vspace{0.15in}

\noindent The analysis is performed by fitting a FM-QRF for each $\tau$ and by computing the Variable Importance measure \citep{breiman1984classification} for each $p$-th covariate as follows.
First, the sum of squared residuals (SSR), denoted with $m_{p,\tau}$, is computed by using Out-Of-Bag observations of the covariate. Then, the observations are permuted and the SSR is re-calculated. The SSR after the permutation is denoted with $m^*_{p, \tau}$.
The Variable Importance of the $p$-th covariate for the $\tau$ quantile, denoted with $I_{p, \tau}$, is finally computed as:
\begin{center}
\begin{equation}
I_{p, \tau}=m_{p, \tau}-m^*_{p, \tau}.
\end{equation}
\end{center}
The higher the reduction in the SSR following the permutation, the higher the importance of the variable.
The results of variable importance are shown in figures \ref{fig:meanimp}- \ref{fig:varimp}. Figure \ref{fig:meanimp} shows the average $I_{p}$ across quantiles, whereas figures \ref{fig:varimp} show the $I_{p, \tau}$ at each quantile level for each variable.

\vspace{0.15in}

\noindent Figure \ref{fig:meanimp} shows that, at an aggregate level, the most important set of variables used to predict quantiles are the lags of the Temperature variable. The second set of variables is represented by the consecutive number of Dry Spells. This pattern is similar to the one observed at a disaggregate level, represented in Figure \ref{fig:varimp} at different quantile probability levels. 

\begin{figure}[H]
    \centering
    \includegraphics[trim=1cm 0 0 0,  width=0.9\textwidth]{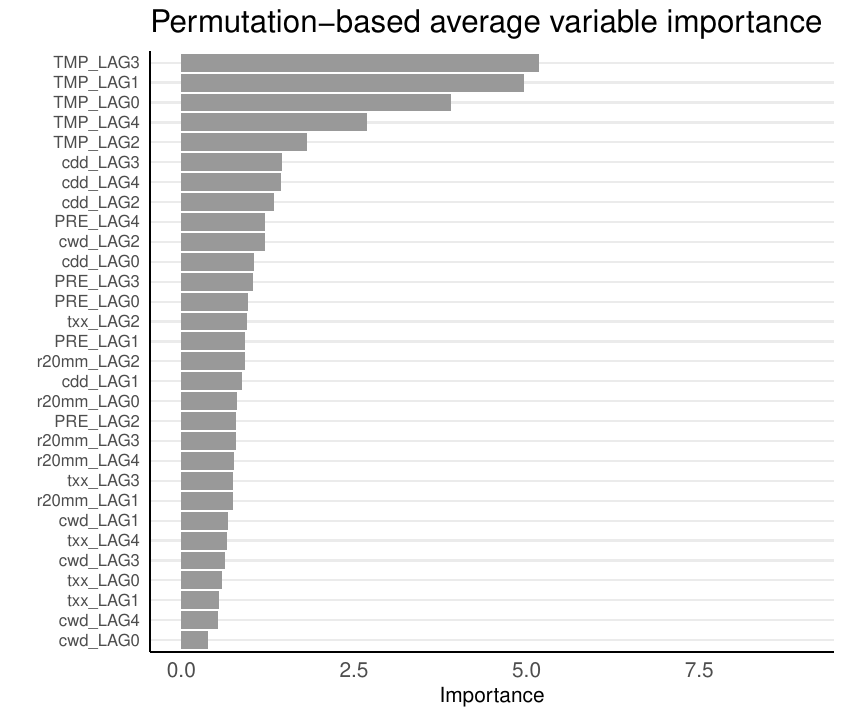}
    \caption{Average variable importance across quantiles $I_j$ for each covariate.}
    \label{fig:meanimp}
\end{figure}

\begin{figure}[H]
    \centering
    \includegraphics[trim=1cm 0 0 0, width=\textwidth]{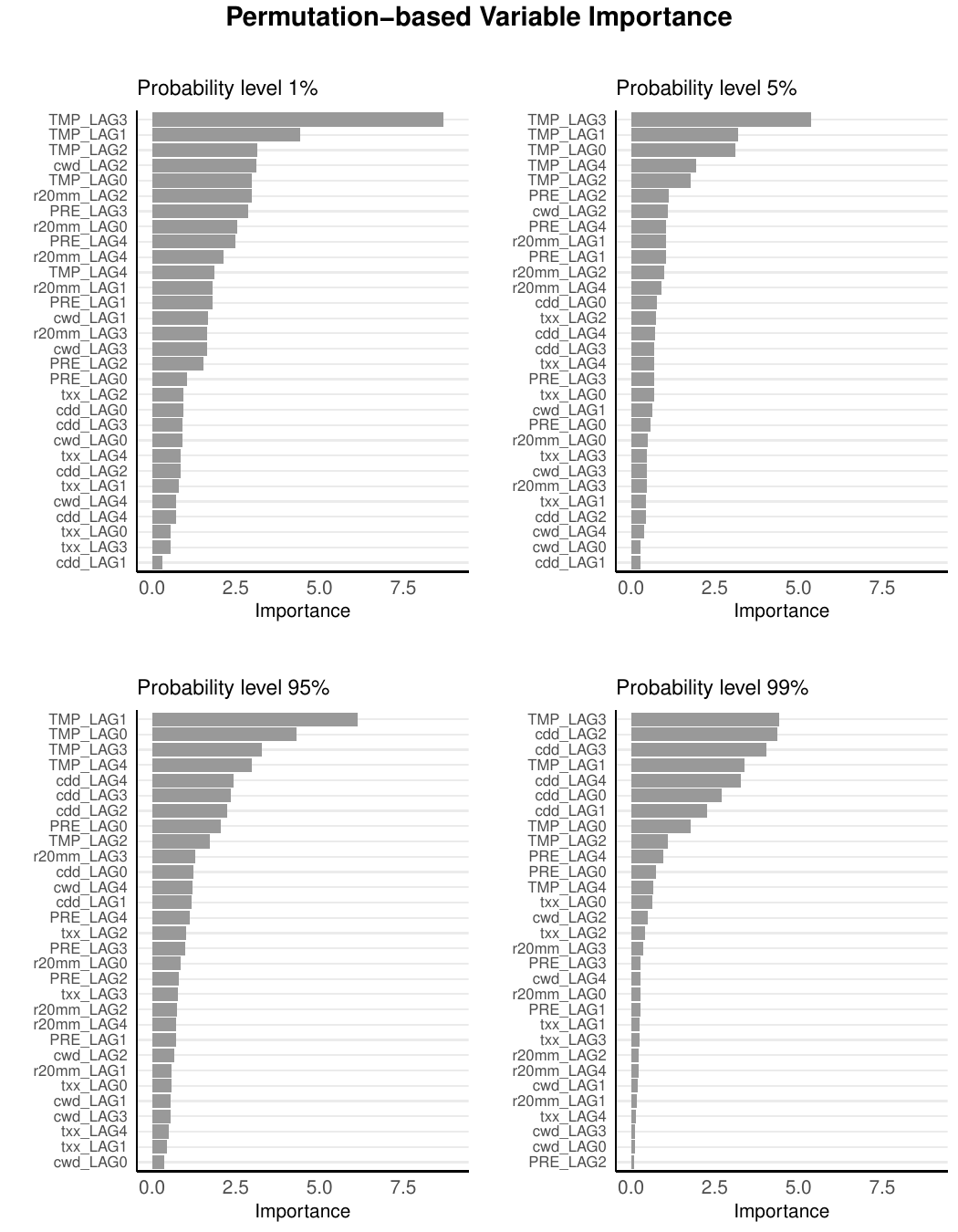}
     \caption{Permutation-based variable importance $I_{j,\tau}$ for each covariate included in the training set.}
    \label{fig:varimp}
\end{figure}

\vspace{0.15in}

\noindent These results highlight that both temperature and precipitations affect the GDP growth distribution of each country to a different extent. Although being in contrast with previous contributions \citep{dell2012temperature}, the obtained results are confirmed by more recent findings \citep{damania2020does}, which show that precipitations-related variables play a more relevant role with respect to temperature at the upper quantiles of the GDP growth distribution, representing a relevant factor influencing upside risk of economic productivity growth. 

\subsection{Projection Results}\label{sec:4.2}

This section reports the quantile estimates along with their standard errors projected in $2030,2050,2100$ under the SSP1 and SSP5 scenarios. The FM-QRF has been fitted using $K=9$ locations, identified via grid search and the computational time to fit the model is equal, on average, to 620.45 seconds on an ordinary multi-CPU server Intel Xeon with 24 cores.
\vspace{0.15in}

\noindent The mean and the bootstrap standard error of each estimate are obtained with $N=500$ replications. Table \ref{tab:avg-se} report the average mean and standard error across countries for the SSP1 and SSP5 scenarios, respectively. Country specific estimates are reported in Tables \ref{tab:se_table_SSP1} and \ref{tab:se_table_SSP5}. 

\begin{table}[]
\centering
\resizebox{\columnwidth}{!}{%
\begin{tabular}{@{}
>{\columncolor[HTML]{FFFFFF}}c 
>{\columncolor[HTML]{FFFFFF}}c 
>{\columncolor[HTML]{FFFFFF}}c 
>{\columncolor[HTML]{FFFFFF}}c 
>{\columncolor[HTML]{FFFFFF}}c 
>{\columncolor[HTML]{FFFFFF}}c 
>{\columncolor[HTML]{FFFFFF}}c 
>{\columncolor[HTML]{FFFFFF}}c 
>{\columncolor[HTML]{FFFFFF}}c 
>{\columncolor[HTML]{FFFFFF}}c 
>{\columncolor[HTML]{FFFFFF}}c 
>{\columncolor[HTML]{FFFFFF}}c 
>{\columncolor[HTML]{FFFFFF}}c 
>{\columncolor[HTML]{FFFFFF}}c 
>{\columncolor[HTML]{FFFFFF}}c 
>{\columncolor[HTML]{FFFFFF}}c 
>{\columncolor[HTML]{FFFFFF}}c 
>{\columncolor[HTML]{FFFFFF}}c 
>{\columncolor[HTML]{FFFFFF}}c @{}}
\toprule
              & \multicolumn{6}{c}{\cellcolor[HTML]{FFFFFF}\textbf{0.01}}                                                                                     & \multicolumn{6}{c}{\cellcolor[HTML]{FFFFFF}\textbf{0.5}}                                                                                   & \multicolumn{6}{c}{\cellcolor[HTML]{FFFFFF}\textbf{0.99}}                                        \\ \midrule
\textbf{year} & \textbf{2030}  & \textbf{2030} & \textbf{2050}  & \textbf{2050} & \textbf{2100}  & \multicolumn{1}{c|}{\cellcolor[HTML]{FFFFFF}\textbf{2100}} & \textbf{2030} & \textbf{2030} & \textbf{2050} & \textbf{2050} & \textbf{2100} & \multicolumn{1}{c|}{\cellcolor[HTML]{FFFFFF}\textbf{2100}} & \textbf{2030}  & \textbf{2030} & \textbf{2050}  & \textbf{2050} & \textbf{2100}  & \textbf{2100} \\
\textbf{SSP1} & \textbf{-7.68} & 0.51          & \textbf{-7.76} & 0.55          & \textbf{-7.87} & \multicolumn{1}{c|}{\cellcolor[HTML]{FFFFFF}0.61}          & \textbf{3.67} & 0.25          & \textbf{3.74} & 0.25          & \textbf{3.65} & \multicolumn{1}{c|}{\cellcolor[HTML]{FFFFFF}0.24}          & \textbf{12.1}  & 0.37          & \textbf{12.09} & 0.38          & \textbf{12.1}  & 0.37          \\
\textbf{SSP5} & \textbf{-7.55} & 0.48          & \textbf{-7.7}  & 0.52          & \textbf{-7.61} & \multicolumn{1}{c|}{\cellcolor[HTML]{FFFFFF}0.54}          & \textbf{3.75} & 0.24          & \textbf{3.75} & 0.25          & \textbf{3.75} & \multicolumn{1}{c|}{\cellcolor[HTML]{FFFFFF}0.24}          & \textbf{12.09} & 0.37          & \textbf{12.13} & 0.37          & \textbf{12.12} & 0.37          \\ \bottomrule
\end{tabular}%
}
\caption{Average mean and bootstrap standard error across countries obtained with N=500 iterations.}
\label{tab:avg-se}
\end{table}

\vspace{0.15in}

\noindent The results obtained under the two scenarios are compared by considering the difference between the values of quantiles estimated in the worst-case scenario in terms of climate policies (SSP5) and the values of quantiles estimated according to the best-case one (SSP1):
\begin{equation}
\Delta Q_{it}^{\tau}= Q_{it}^{\tau, SSP5}-Q_{it}^{\tau, SSP1},
\end{equation}

\noindent where $Q_{it}^{\tau}$ represents the estimated values of the quantile of the $i$-th country at probability level $\tau$ projected in year $t$.
Given that the main difference between the SSP1 and SSP5 is related to two opposite climate change mitigation policies scenarios, the magnitude of $\Delta Q_{it}^{\tau}$ can be attributed only to climate change and its effect on climate variables.

\vspace{0.15in}

\noindent Negative values of $\Delta Q_{it}^{\tau}$ indicate that  $Q_{it}^{\tau, SSP5} < Q_{it}^{\tau, SSP1}$. This means that under the SSP5 scenario, the GDP growth at both low and high quantile levels will be smaller due to inadequate climate-related policies.

\vspace{0.15in}

\noindent The country-specific results of this analysis are mapped in figures \ref{fig:1}-\ref{fig:5} at three different time horizons ($t=2030, 2050, 2100$). The main finding obtained by interpreting the values of $\Delta Q_{it}^{\tau}$ is that the effects of the unsustainable climate practices hypothesised in the SSP5 scenario are negative at an aggregate level, but their magnitude substantially vary across time and among countries. 

\vspace{0.15in}

\noindent For each quantile level, in $2030-2050$ the value of $\Delta Q_{it}^{\tau}$ is estimated to be negative for almost two thirds of the countries. This points out that the climate change hypothesised in the worst case scenario will increase the downside risk of the majority of the countries in the considered sample, while also representing a limiting factor to the potential country-specific GDP growth. 

\vspace{0.15in}

\noindent For instance, figure \ref{fig:4} shows the maps at 2030, 2050 and 2100 for quantile at level $\tau=0.05$. In this case, in 2030 the $\Delta Q_{i, 2030}$ assumes values ranging from a minimum of -16\% to a maximum of 7\%. This indicates that for some countries $Q_{i,2030}^{0.05, SSP5} < Q_{i,2030}^{0.05, SSP1}$, that is,  some countries will see their 5\% quantile (that represents a negative value of the GDP growth) decreasing more in the ``worst-case'' climate change scenario with respect to the ``best-case'' one. Another finding is that the effects of unsustainable climate practices change over time whether the country is considered to be 'high-income' with a moderate climate, o 'low-income' with more extreme climate conditions.

\vspace{0.15in}

\noindent By comparing the panels reported in Figures \ref{fig:1}- \ref{fig:5}, the following     results can be deducted. The impact of temperatures and precipitations from 2030 to 2050 on the United States, considered as a high-income and climate-temperate country, are modestly positive at all quantile levels. These findings indicate that high-income countries with economies reliant on energy-intensive sectors could see advantages in scenarios with limited climate policies, indicating a higher capability of high-income countries to adapt more effectively to adverse climate-change scenarios.

\vspace{0.15in}

\noindent For low-income and hot countries instead, such as countries in the African continent and India, the effects of climate-change from 2030 to 2050 are negative at all quantile levels, especially at the medium term (2050). The same holds for Russia, which is a high-income but cold country. In this case, Russia will suffer from adverse climate-change scenarios both in terms of upside and downside risk.
The results on the heterogeneity of climate change effects over time on rich/poor and hot/cold countries are confirmed by previous findings \citep{kiley2021growth}.

\vspace{0.15in}

\noindent In general, even though between 2030 and 2050 the majority of the countries will suffer from climate change in terms of GDP growth, in 2100 this heterogeneity will be slightly less pronounced, bringing the number of countries suffering from climate change to two thirds of the sample to 50\%. This might result from the  optimistic socio-demographic growth scenario hypothesised in SSP5, based on the hypothesis of large public interventions in favor of the communities against climate change. The effects of these interventions could have a positive effect in the long run on the poorest countries which, as shown in the literature, are those which will suffer most from the effects of climate change, precisely due to their socio-economic conditions. 

\vspace{0.15in}

\noindent The socio-economic development of these countries would thus ensure that the effects of climate change, even though still existing, will be mitigated by the presence of a growing number of mitigation policies and public intervention.

\begin{landscape}
\begin{figure}[H]
    \centering
    \includegraphics[height=13cm]{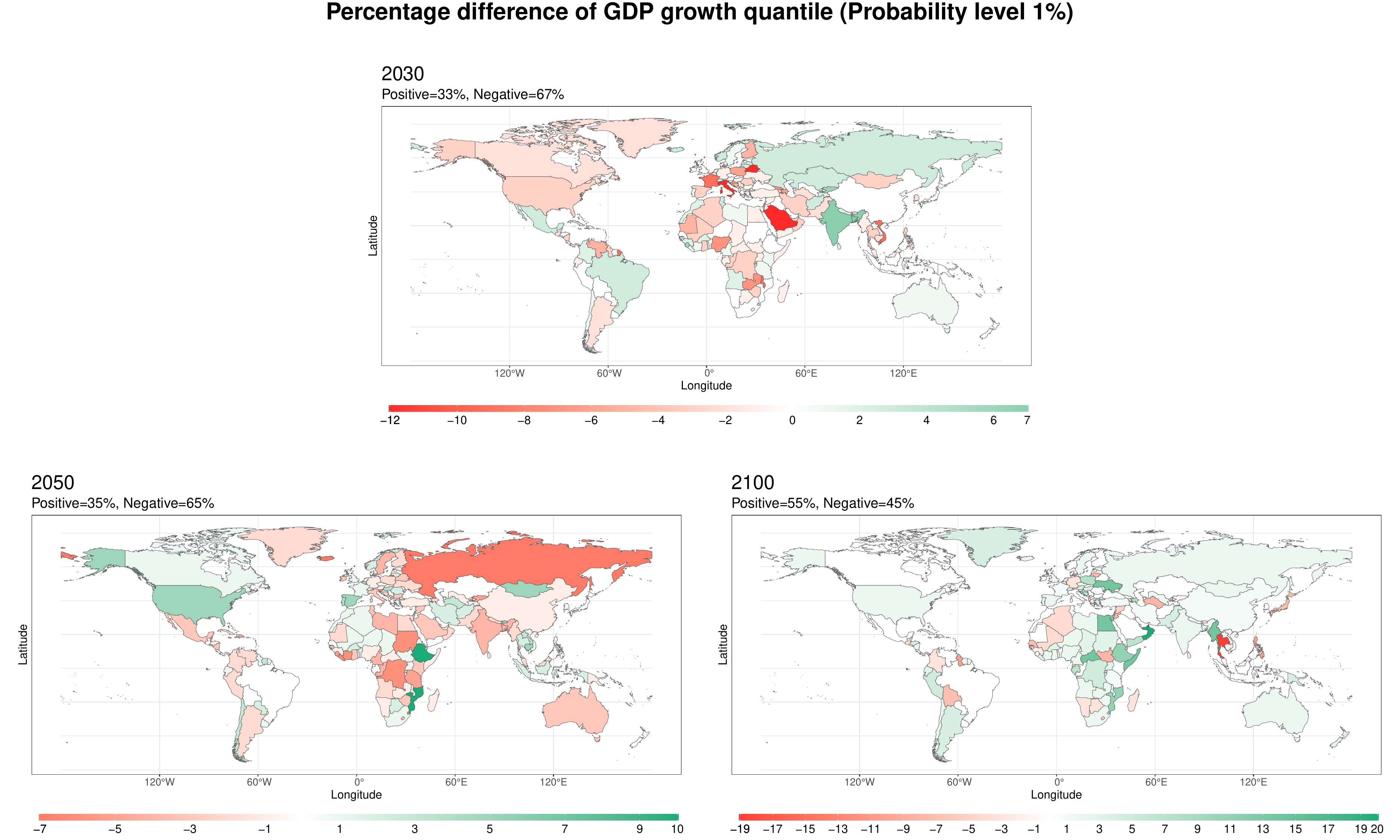}
   \caption{Map showing $\Delta Q_{it}^{\tau}$ in years $t=2030,2050,2100$ for quantile at probability level $\tau=0.01$.}
    \label{fig:1}
\end{figure}
\end{landscape}

\begin{landscape}
\begin{figure}[H]
    \centering
    \includegraphics[height=13cm]{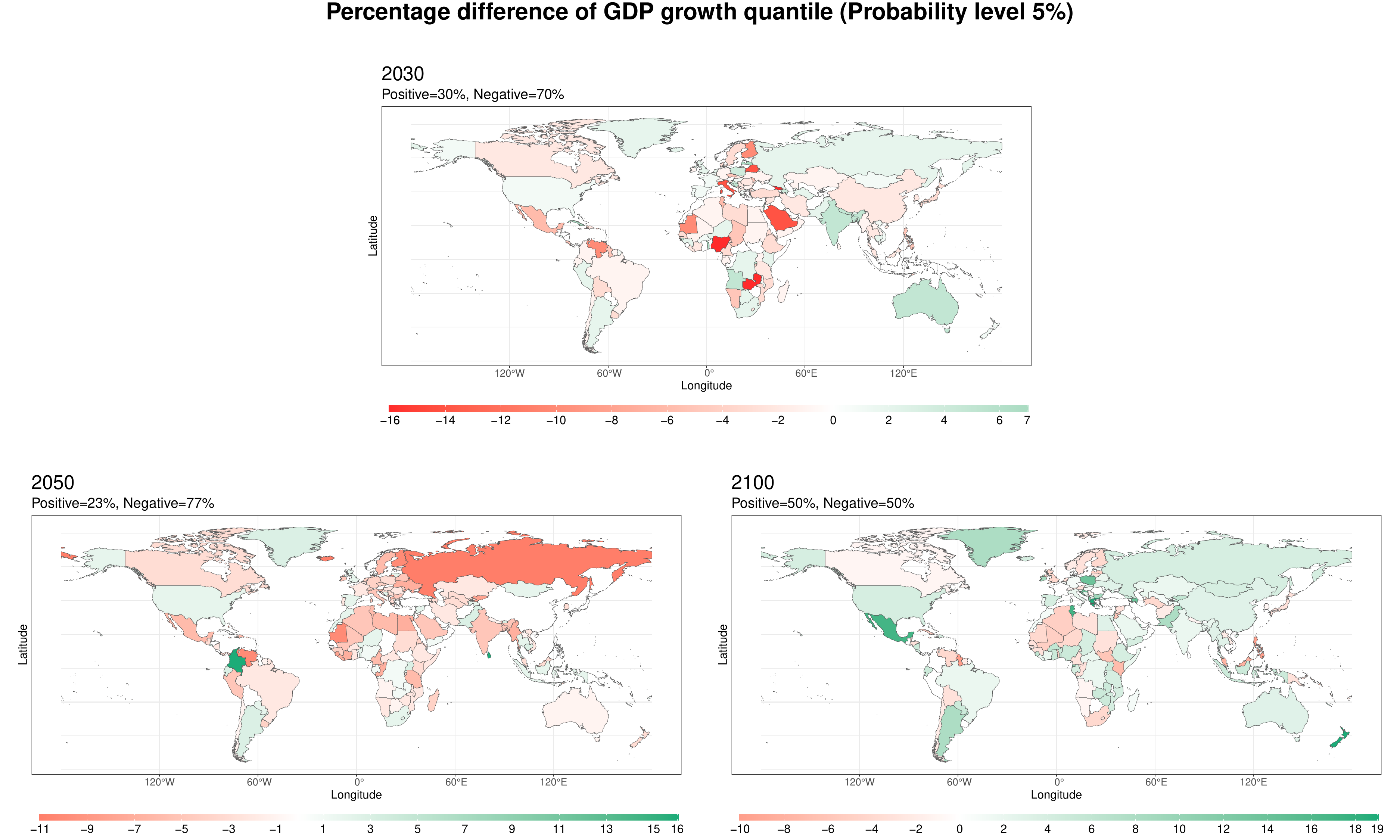}
   \caption{Map showing $\Delta Q_{it}^{\tau}$ in years $t=2030,2050,2100$ for quantile at probability level $\tau=0.05$.}
        \label{fig:2}
\end{figure}
\end{landscape}

\begin{landscape}
\begin{figure}[h]
    \centering
    \includegraphics[height=13cm]{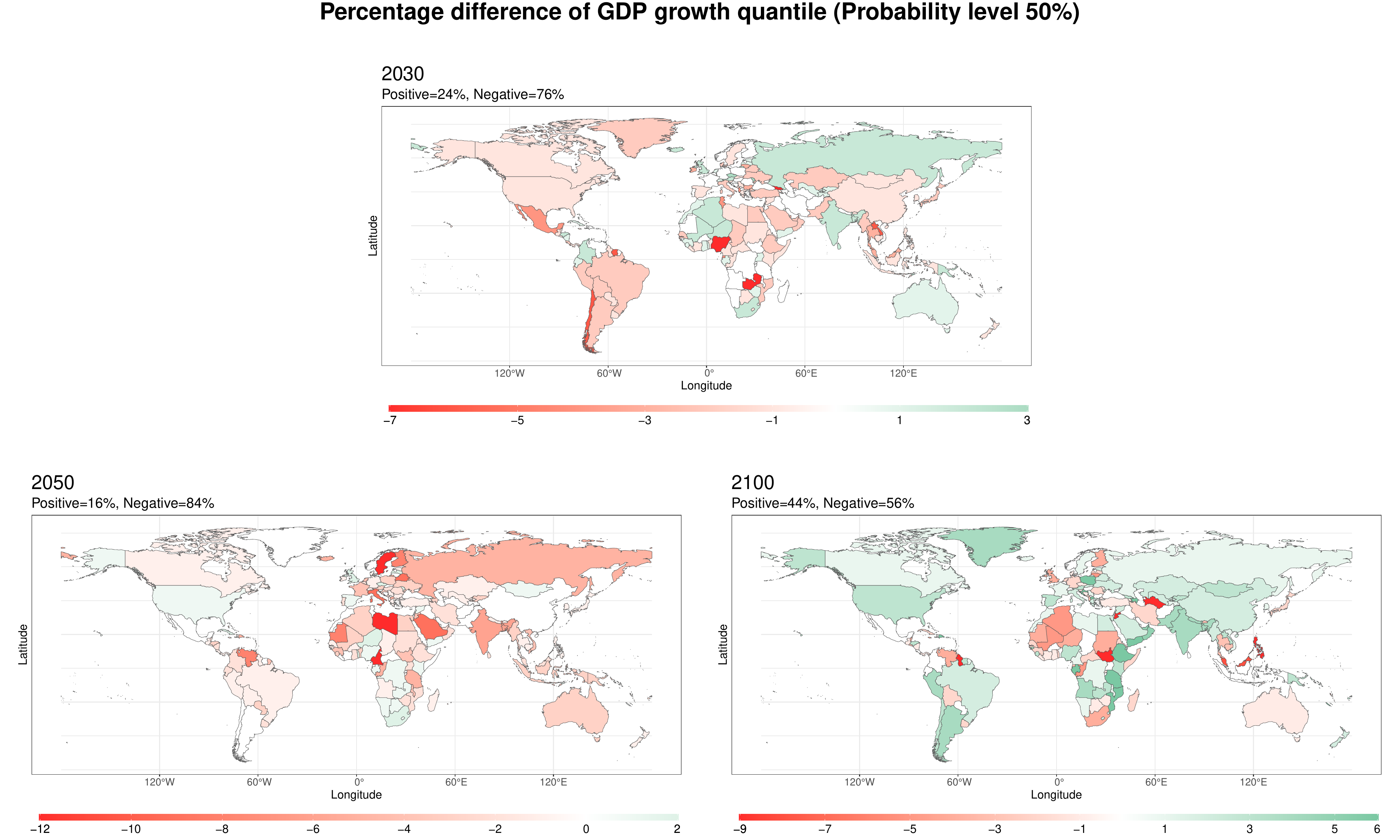}
    \caption{Map showing $\Delta Q_{it}^{\tau}$ in years $t=2030,2050,2100$ for quantile at probability level $\tau=0.50$.}
    \label{fig:3}
\end{figure}
\end{landscape}

\begin{landscape}
\begin{figure}[H]
    \centering
    \includegraphics[height=13cm]{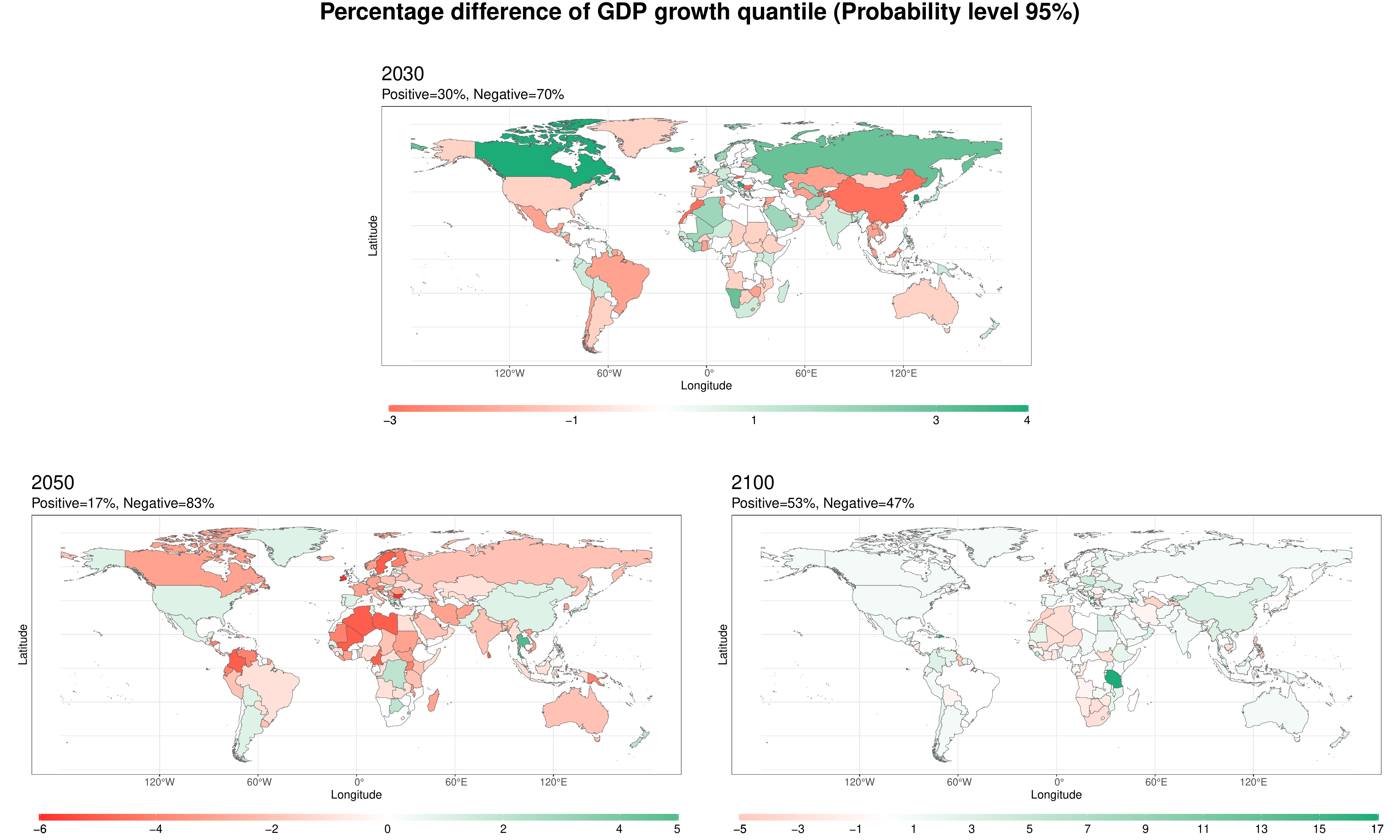}
     \caption{Map showing $\Delta Q_{it}^{\tau}$ in years $t=2030,2050,2100$ for quantile at probability level $\tau=0.95$.}
    \label{fig:4}
\end{figure}
\end{landscape}

\begin{landscape}
\begin{figure}[H]
    \centering
    \includegraphics[height=13cm]{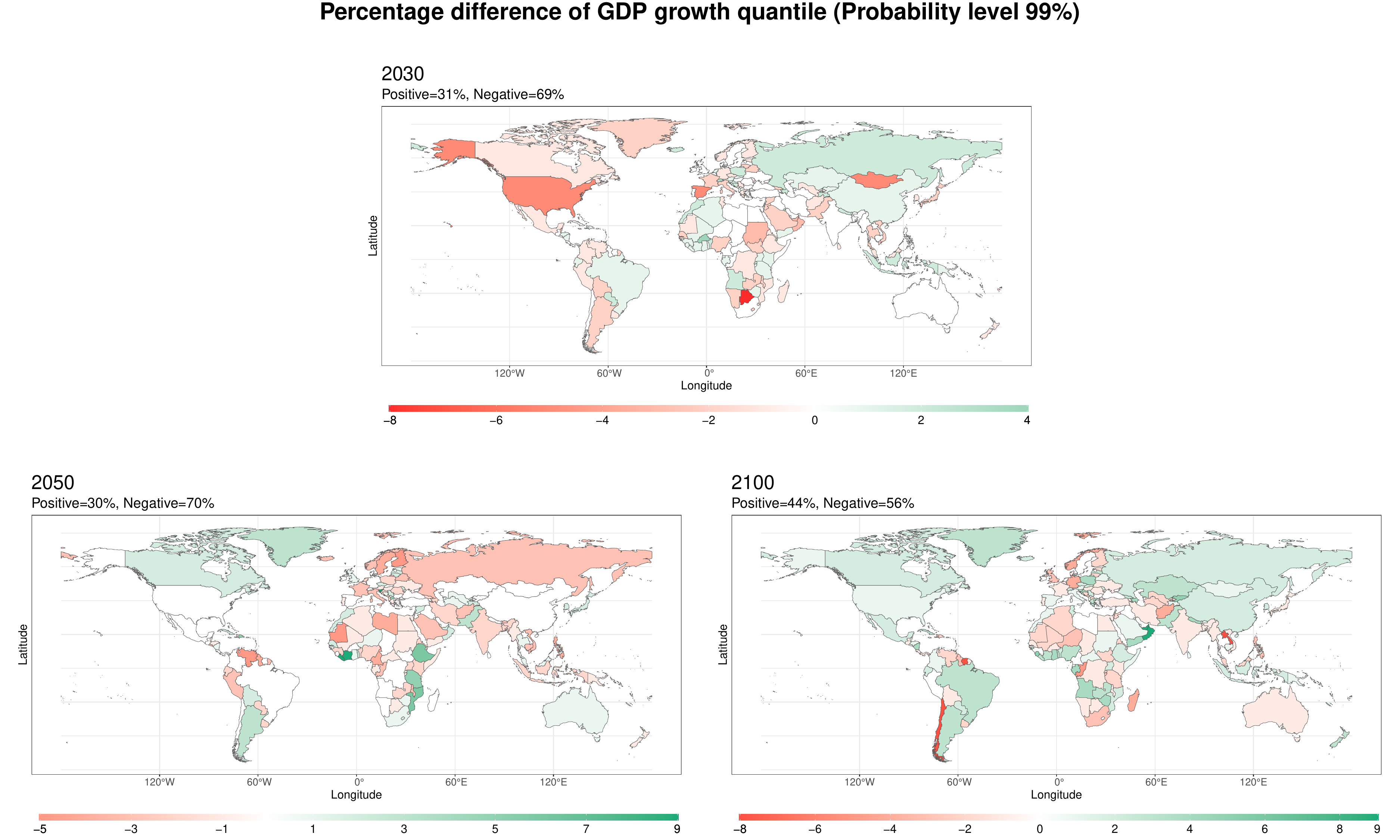}
    \caption{Map showing $\Delta Q_{it}^{\tau}$ in years $t=2030,2050,2100$ for quantile at probability level $\tau=0.99$.}
    \label{fig:5}
\end{figure}
\end{landscape}

\section{Conclusions}\label{sec:FM-QRF-conclusions}

This chapter introduces the FM-QRF, a novel model to estimate quantiles of longitudinal data in a non-linear mixed-effects framework by means of QRF and the NPML approach.

\vspace{0.15in}

\noindent A large scale simulation study shows that the FM-QRF outperforms other benchmark models in a non-linear setting. The model is applied empirically to study the long-term effects of climate change on GDP growth-at-risk, unveiling the relevant effects of future changes in temperature and precipitations on the tails of the GDP growth distribution.
\vspace{0.15in}

\noindent The non-linear approach presented in this chapter offers several advantages when assessing the economic effects of climate change compared to a standard linear approach, as the FM-QRF allows to model the complex non-linear relationships between climate-related variables and GDP growth without any a-priori distributional assumption on the form these relationships, on the outcome variable and the random effects parameters. The empirical results presented in this chapter point out that these features make the FM-QRF well-suited for evaluating the future economic effects of climate change, allowing to deliver more accurate forecasts with respect to standard linear approaches.


\vspace{0.15in}

\noindent Possible future developments of the FM-QRF model concern the inclusion of mixed-frequency covariates in order to consider variables observed at a higher frequency with respect to the outcome, that may include important information for understanding the phenomenon under consideration. The analysis of the results of the FM-QRF could be further enriched by including the study of the presence of heterogeneity among statistical units in terms of individual intercepts. Moreover, future model extension could concern the inclusion of time-varying random parameters by exploiting a Markovian structure as proposed in \cite{marino2018mixed, merlo2022quantilets, merlo2022quantile}.


\chapter{The Impact of the COVID-19 Pandemic on Risk Factors for Children's Mental Health: Evidence from the UK Household Longitudinal Study}\label{ch:SDQ}
\chaptermark{The impact of COVID-19 Pandemic}

\section{Introduction}

The recent COVID-19 pandemic sensibly impacted communities, healthcare systems and society, affecting the physical and mental health of people of all ages. Among these broader effects, social isolation and increased household stress particularly affected children, and part of the extensive literature on the pandemic has focused on investigating the effects of such disruptions on children's mental health \citep{de2020covid, ma2021prevalence, adegboye2021understanding,kauhanen2023systematic}.

\vspace{0.15in}

\noindent In this context, a variety of studies have used longitudinal data concerning the "SDQ score", a widely used measure in pediatric psychological research computed with answers to the Strengths and Difficulties Questionnaire (SDQ) \citep{panagi2024mental, miall2023inequalities, merlo2022quantile, ravens2021quality, bignardi2021longitudinal, waite2021did}. This metric has been introduced in \cite{goodman1997strengths} to evaluate children's emotional, social and behavioural characteristics. High SDQ scores indicate a higher prevalence of psychological issues related to conduct or peer-related issues, hyperactivity, and emotional symptoms, including anxiety and depression.
\vspace{0.15in}

\noindent Under a methodological point of view, given the longitudinal structure of the these data, part of the above-mentioned studies have implemented standard linear random-effects or mixed-effects models, which allow to estimate the expected value of the conditional distribution of the response variable in a linear modeling framework. However, in many empirical applications, the relationship between the outcome and the covariates is non-linear, and the outcome's distribution often violates the gaussianity assumptions of random-effects models. This is the case of the SDQ distribution, which is typically asymmetric. Moreover, results of previous contributions point out that the effect of risk factors such as socio-economic disadvantage and maternal depression changes across the SDQ conditional distribution, with a more significant impact on the right tail, which is related to abnormal levels of behavioural issues \citep{davis2010socioeconomic, flouri2008psychopathology, flouri2010modeling, merlo2022quantile, tzavidis2016longitudinal}.

\vspace{0.15in}

\noindent These findings highlight that a more complete picture of the SDQ conditional distribution is necessary to gain a deeper understanding of the phenomenon of interest. In this context, contributions such as \cite{tzavidis2016longitudinal} and \cite{merlo2022quantile} implemented QR mixed-effects models to obtain more robust and accurate results. These approaches allow to infer the entire conditional distribution of the outcome by modeling location parameters beyond the conditional expected value, such as quantiles, in a longitudinal data setting.

\vspace{0.15in}

\noindent To overcome the drawbacks of parametric formulations of these models, QR has been also extended to the non-parametric setting by developing QR machine learning algorithms, such as QR Neural Networks \citep{white1992nonparametric}, QR Support Vector Machines \citep{hwang2005simple, xu2015weighted}, QR kernel based algorithms \citep{christmann2008consistency}, QR Forests (QRF) \citep{meinshausen2006quantile} and Generalized QR Forests 
\citep{ athey2019generalized}. However, these models cannot handle longitudinal data in a mixed-effects framework, and for this reason the analysis carried out in this chapter employs the FM-QRF algorithm of Chapter \ref{ch:FM-QRF}. 

\vspace{0.15in}

\noindent The FM-QRF extends the traditional QR model to a non-linear setting and it also enhances the QRF algorithm by introducing an additional parameter represented by the random effects part of mixed-effects model, which allows to model unobserved heterogeneity across statistical units. These unique features make the FM-QRF well-suited for modeling phenomena characterized by complex non-linear relationships among the variables of interest, especially in empirical studies involving repeated measurements and longitudinal data, such as SDQ score data.

\vspace{0.15in}

\noindent Thanks to this approach, this study contributes to the strand of literature on children's mental health by uncovering relationships between children's mental health and risk factors during the pandemic that have not have been captured by the standard linear models applied in previous contributions \citep{walton2010contextual,goodnight2012quasi,bradley2002socioeconomic}.

\vspace{0.15in}

\noindent To this end, the data used in this study concern the SDQ scores of children that participated in at the UK Household Longitudinal Study (UKHLS), the largest household panel survey carried out in the United Kingdom since 2009. Data related to this study have been widely used in the literature concerning the effects of the pandemic on mental health \citep{miall2023inequalities,bayrakdar2023inequalities,metherell2022digital,daly2022psychological,thorn2022education,reimers2022primary,mendolia2022have}, and the aim of this study is to use the SDQ score as outcome variable and a set of variables selected according to previous studies findings as set of covariates.

\vspace{0.15in}

\noindent To compare the set of risk factors before and after the pandemic, the FM-QRF is trained with two datasets: the first contains observations from the pre-pandemic period (2016-2019), and the other from the pandemic period (2020-2021). Then, quantiles at five different probability levels are estimated for each period and the Variable Importance measure is extracted to select the most significant risk factors for children's mental health in the two periods of interest.

\vspace{0.15in}

\noindent The results from the pre-pandemic and pandemic periods are compared to identify patterns and shifts in the most important risk factors. Additionally, these findings are compared with the set of statistically significant variables obtained using the LQMM model.

\vspace{0.15in}

\noindent The empirical findings obtained with the FM-QRF indicate that the key risk factors vary based on the quantile level, justifying the adoption of a QR approach over a standard regression method, and that they differ between the pre-pandemic and pandemic periods. Results also show that relying solely on the LQMM model provides only a partial understanding of the phenomena under investigation, highlighting the relevance of a non-linear approach. In this sense, the FM-QRF proves to be a valuable choice to gain a deeper understanding of the complex relationship between the COVID-19 pandemic and children's mental well-being.

\vspace{0.15in}

\noindent The rest of the chapter is organized as follows. Section \ref{sec:SDQ-data} presents the data used in this empirical study, Section \ref{sec:SDQ-empirical} reports the results obtained with the FM-QRF and the comparison with the LQMM model and Section  \ref{sec:SDQ-conclusions} concludes.

\section{The Data}
\label{sec:SDQ-data}

The data used in this study concern the SDQ score as outcome variable and a set of  covariates selected according to previous results findings.

\vspace{0.15in}

\noindent In particular, the SDQ score data have been retrieved by the UKHLS, the largest household panel survey carried out in the United Kingdom since 2009 involving 40,000 households. The aim of the UKHLS is to capture a broad range of information concerning economic circumstances, employment, education, health, and mental well-being before and after the pandemic. The longitudinal structure of this dataset provides a unique set of information to understand the long-term impact of a variety of factors on households.

\vspace{0.15in}

\noindent The UKHLS collects also data of individual responses from both adults, youngsters and children. In the latter case, data concerning the assessment of children mental health are collected through the SDQ. The focus of the analysis presented in this chapter is the "total difficulties score" (\textit{chsdqtd}) which ranges from 0 to 40 and represents the sum of 25 scores related to five domains comprising emotional symptoms, peer problems, conduct problems, hyperactivity and prosocial behaviour. The response for each item is evaluated on a 3-points scale, where 0 represents a "not true" answer, 1 is given if the respondent partially agrees with the question statement, and 2 for a "true" answer. High scores indicate a higher prevalence of psychological issues related to conduct or peer-related issues, hyperactivity, inattention, and emotional symptoms, including anxiety and depression.

\vspace{0.15in}

\noindent The covariates have been selected based on previous studies on children's mental health risk factors, such as family poverty, ethnicity, and living area. In particular, the set of covariates is composed of variables whose effects might have changed between before and during the pandemic: social benefit income (\textit{fihhmnsben}), income from investements (\textit{fihhmninv}),  household size (\textit{hhsize}) and number of bedrooms in the house (\textit{hsbeds}), number of employed people in the household (\textit{nemp}) and employed people in the household that are not being paid (\textit{nue}), living area (urban or rural) (\textit{urban}), being up to date with bills payments (\textit{xphsdba}) and internet access (\textit{pcnet}).
The belonging to an ethnic minority (\textit{emboost}) is considered as an additional time-invariant variable.

\vspace{0.15in}

\noindent The aim of this chapter is to compare how the main risk factors for children's mental health changed before and during the pandemic period.
Thus, the dataset comprises observations from 2016 to 2019 for the pre-pandemic period and from 2020 to 2021 for the pandemic period.
As stated above, the outcome variable is the SDQ total score, and after eliminating statistical units with missing data, the final sample including both the pre-pandemic and pandemic periods is composed of 2401 children for a total of $n=3101$ observations. A total of 1729 and 1028 children are included in the pre-pandemic and pandemic sample, respectively. The barplot showing the number of children interviewed 1, 2, 3 or 4
times during sample period is represented in Figure \ref{fig:n_children}.

\vspace{0.15in}

\noindent The summary statistics reported in Tables \ref{tab:summary-PRECOVID} and \ref{tab:summary-COVID}, related to the pre-COVID and COVID periods respectively, highlight the asymmetry of the SDQ unconditional distribution, whose mean and median sensibly differ especially in the pre-COVID period.

\begin{table}[]
\centering
\resizebox{\textwidth}{!}{%
\begin{tabular}{@{}
>{\columncolor[HTML]{FFFFFF}}l 
>{\columncolor[HTML]{FFFFFF}}l 
>{\columncolor[HTML]{FFFFFF}}l 
>{\columncolor[HTML]{FFFFFF}}l 
>{\columncolor[HTML]{FFFFFF}}l 
>{\columncolor[HTML]{FFFFFF}}l 
>{\columncolor[HTML]{FFFFFF}}l 
>{\columncolor[HTML]{FFFFFF}}l @{}}
\toprule
\textit{}               & \textbf{Obs} & \textbf{Min} & \textbf{Max} & \textbf{Median} & \textbf{Mean} & \textbf{St.Dev} & \textbf{Null} \\ \midrule
\textit{chsdqtd}    & 2006         &              & 37.00        & 8.00            & 8.51          & 5.94            & 8856          \\
\textit{urban}      & 2006         &              & 2.00         & 1.00            & 1.23          & 0.42            & 14            \\
\textit{hhsize}     & 2006         &              & 13.00        & 4.00            & 4.32          & 1.18            & 8833          \\
\textit{hsbeds}     & 2006         &              & 8.00         & 3.00            & 3.29          & 0.87            & 8835          \\
\textit{fihhmnsben} & 2006         &              & 6096.14      & 235.67          & 605.84        & 758.22          & 0             \\
\textit{fihhmninv}  & 2006         &              & 19416.67     & 0.00            & 173.01        & 845.31          & 8             \\
\textit{nemp}       & 2006         &              & 7.00         & 2.00            & 1.63          & 0.71            & 0             \\
\textit{pcnet}      & 2006         &              & 2.00         & 1.00            & 1.01          & 0.10            & 6             \\
\textit{nue}        & 2006         &              & 6.00         & 0.00            & 0.45          & 0.72            & 0             \\
\textit{xphsdba}    & 2006         &              & 3.00         & 1.00            & 1.08          & 0.28            & 22            \\
\textit{emboost}    & 2006         &              & 1.00         & 0.00            & 0.10          & 0.30            & 0             \\  \bottomrule
\end{tabular}%
}
\caption{Summary statistics related to the pre-COVID period of the variables included in the sample after the data cleaning procedure. The Null column reports the number of null values (not applicable or missing items) before the data cleaning procedure.}
\label{tab:summary-PRECOVID}
\end{table}
\begin{table}[]
\centering
\resizebox{\textwidth}{!}{%
\begin{tabular}{@{}
>{\columncolor[HTML]{FFFFFF}}l 
>{\columncolor[HTML]{FFFFFF}}l 
>{\columncolor[HTML]{FFFFFF}}l 
>{\columncolor[HTML]{FFFFFF}}l 
>{\columncolor[HTML]{FFFFFF}}l 
>{\columncolor[HTML]{FFFFFF}}l 
>{\columncolor[HTML]{FFFFFF}}l 
>{\columncolor[HTML]{FFFFFF}}l @{}}
\toprule
\textit{}               & \textbf{Obs} & \textbf{Min} & \textbf{Max} & \textbf{Median} & \textbf{Mean} & \textbf{St.Dev} & \textbf{Null} \\ \midrule
\textit{chsdqtd}    & 1095         & 0            & 32.00        & 7               & 8.68          & 6.01            & 9155          \\
\textit{urban}      & 1095         & 1            & 2.00         & 1               & 1.23          & 0.42            & 74            \\
\textit{hhsize}     & 1095         & 2            & 12.00        & 4               & 4.36          & 1.21            & 9137          \\
\textit{hsbeds}     & 1095         & 0            & 8.00         & 3               & 3.33          & 0.92            & 9141          \\
\textit{fihhmnsben} & 1095         & 0            & 4901.66      & 208             & 568.05        & 760.12          & 0             \\
\textit{fihhmninv}  & 1095         & 0            & 9683.33      & 0               & 211.61        & 819.50          & 7             \\
\textit{nemp}       & 1095         & 0            & 7.00         & 2               & 1.62          & 0.72            & 0             \\
\textit{pcnet}      & 1095         & 1            & 2.00         & 1               & 1.01          & 0.07            & 12            \\
\textit{nue}        & 1095         & 0            & 6.00         & 0               & 0.50          & 0.83            & 0             \\
\textit{xphsdba}    & 1095         & 1            & 3.00         & 1               & 1.11          & 0.33            & 24            \\
\textit{emboost}    & 1095         & 0            & 1.00         & 0               & 0.10          & 0.30            & 0             \\ 
\bottomrule
\end{tabular}%
}
\caption{Summary statistics related to the COVID period of the variables included in the sample after the data cleaning procedure. The Null column reports the number of null values (not applicable or missing items) before the data cleaning procedure.}
\label{tab:summary-COVID}
\end{table}

\vspace{0.15in}

\noindent The features of the SDQ conditional distribution given the above-mentioned set of covariates are also investigated by fitting a model for the SDQ score with random-effects at child level. The Q-Q plot shown in Figure \ref{fig:qqplot} highlights a severe departure of the model's residuals from the Gaussian distribution, which is one of the key assumption of the linear random-effects model. Thus, it would be useful to estimate more robust measures of central tendency and conditional quantiles to better estimate the relation between the SDQ score and the covariates.

\begin{figure}[H]
    \centering
\includegraphics[width=0.7\textwidth]{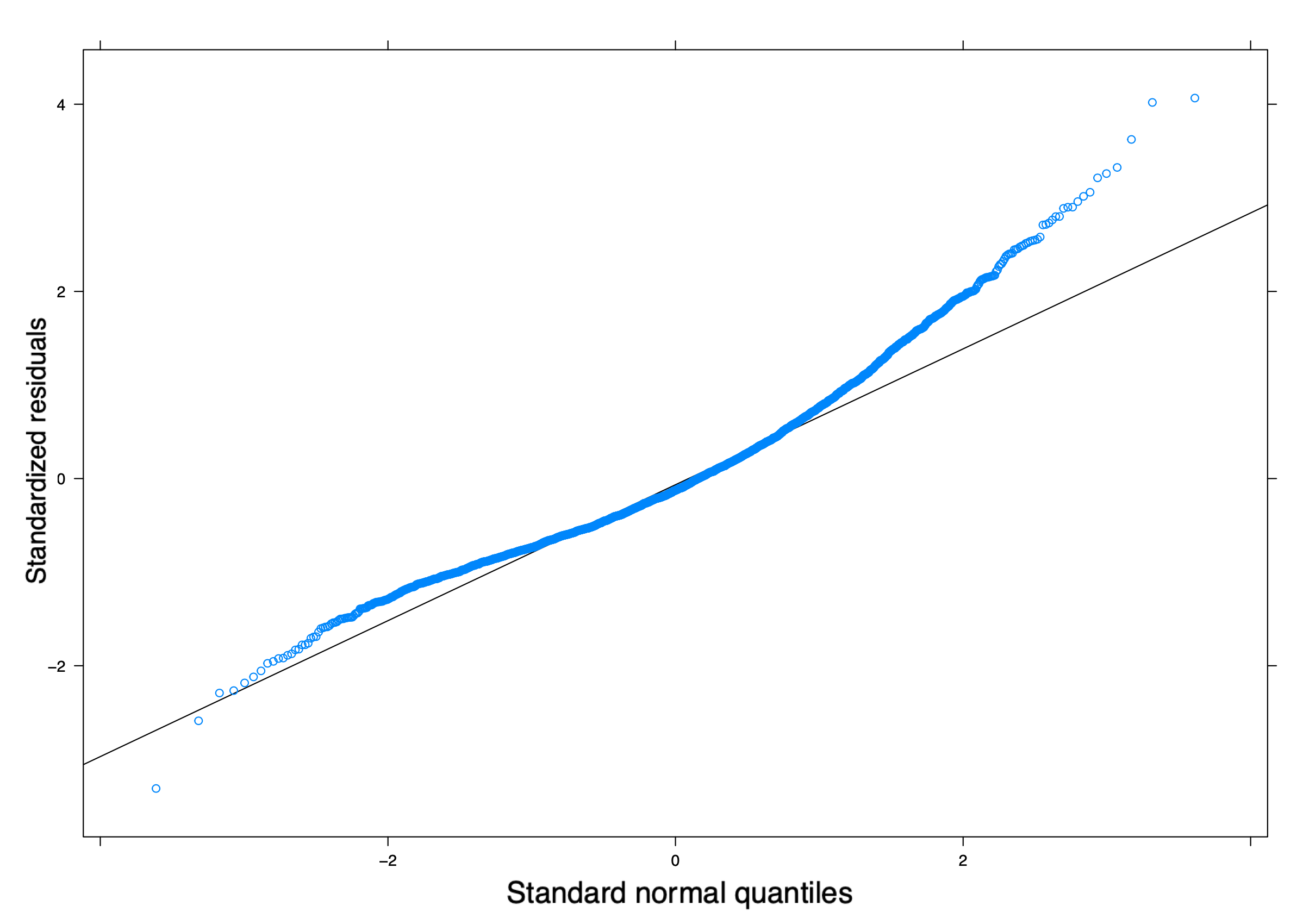}
    \caption{Normal probability plot of the linear mixed model residuals for SDQ total score with  with random-effects specified at child level}
    \label{fig:qqplot}
\end{figure}

\vspace{0.15in}

\noindent The validity of a non-linear QR approach over a standard linear one is assessed with an analysis of variance (ANOVA) test \citep{st1989analysis} to compare a spline QR model and standard QR at five quantile levels $\tau=0.1, 0.25, 0.5, 0.75, 0.9$. 
The spline QR model is an additive model represented by piecewise-defined polynomial functions. This approach allows to model non-linear relationships by means of knot points, at which each different polynomial segment originates. In particular, in this chapter the the QR spline model is represented by
a piecewise cubic polynomial with 3 knots for each covariate.

\vspace{0.15in}

\noindent The results obtained from the ANOVA test in Table \ref{tab:anova} reveal that, with the dataset of interest, a non-linear approach results more valid than the linear one. This finding supports the adoption of a non-linear QR approach, such as the FM-QRF, as a preferred choice over standard linear QR model, especially in scenarios where the underlying data exhibits non-linear relationships.

\begin{table}[htbp]
    \centering
    \caption{ANOVA test results for different values of $\tau$. The 'Tn' column reports the test statistic and the 'P-Value' column reports the level of significance of the test at 5\% significance level.}
    \label{tab:anova}
    \begin{tabularx}{\textwidth}{*{7}{>{\centering\arraybackslash}X}}
        \hline
        \textbf{$\tau$} & \textbf{0.1} & \textbf{0.25} & \textbf{0.5} & \textbf{0.75} & \textbf{0.9} \\ \hline
        \textit{Tn} & 3.471 & 5.478 & 19.392 & 6.536 & 8.895 \\ 
        \textit{P-Value} & 0*** & 0*** & 0*** & 0*** & 0*** \\ \hline
    \end{tabularx}
\end{table}

\section{Analysis of the UK Household Longitudinal Study Data} \label{sec:SDQ-empirical}

This section reports the results of the analysis of the SDQ total score dataset for the selected UKHLS sample of children. Given previous literature results, the analysis concerns the risk factors for children’s emotional and behavioural problems related to family poverty, ethnicity, overcrowding in the household and internet access. 
\vspace{0.15in}

\noindent In particular, the analysis considers covariates related to such risk factors whose effects might have changed between before and during the pandemic: social benefit income, household size and number of bedrooms in the house, number of employed people in the household and employed people in the household that are not being paid, living area (urban or rural), being up to date with bills payments and internet access.
The belonging to an ethnic minority is included as time-invariant variable.

\vspace{0.15in}

\noindent The FM-QRF described in Chapter \ref{ch:FM-QRF} is used to model quantiles at five different levels $\tau=0.1, 0.25, 0.5, 0.75, 0.9$:

\begin{equation}
\hat{Q}_{it,\tau}=\widehat{g_{\tau}(\mathbf{x}_{it})}+\hat{\alpha}_{k, \tau}  \;\;\;  \tau \in (0,1),
	\end{equation}
where $g_\tau: \mathbb{R}^p \rightarrow \mathbb{R}$. In this case, $\hat{Q}_{it,\tau}$ is the estimated quantile for the $i-th$ individual at level $\tau$ and time $t$, $g_{\tau}(\mathbf{x}_{it})$ is the fixed-effects part of the model estimated with a QRF approach and $\alpha_{k, \tau}$ is the random effects part estimated with the finite mixture approach described in Section \ref{sec:FM-QRF-methodology}.

\noindent In this empirical application, the number of mixture components $K$
has been set to 10 via grid search.

\noindent For the sake of clarity, this chapter reports only results for $\tau=0.1, 0.5, 0.9$. Results for the remaining quantile levels, which are similar to the ones reported in this section, are available upon request.

\subsection{Risk Factors Analysis}

The risk factor analysis is run by extracting the Variable Importance measure from the FM-QRF in order to evaluate which covariate has a more relevant role in predicting the SDQ total score quantiles. Figures \ref{fig:quantile01}-\ref{fig:quantile09} report the bar graphs in which each bar represents the variable importance of each covariate at quantile levels $\tau=0.1, 0.5, 0.9$ before and during the COVID-19 pandemic. The variables have been ordered in each graph according to the average variable importance in both periods. This allows to understand both the ranking of the variables in each separate period and the overall ranking at quantile level. The variable importance values for the quantile levels $ \tau=0.25, 0.75$ are shown in Figures \ref{fig:quantile025}-\ref{fig:quantile075}.

\begin{figure}[H]
    \centering
\includegraphics[width=0.8\textwidth]{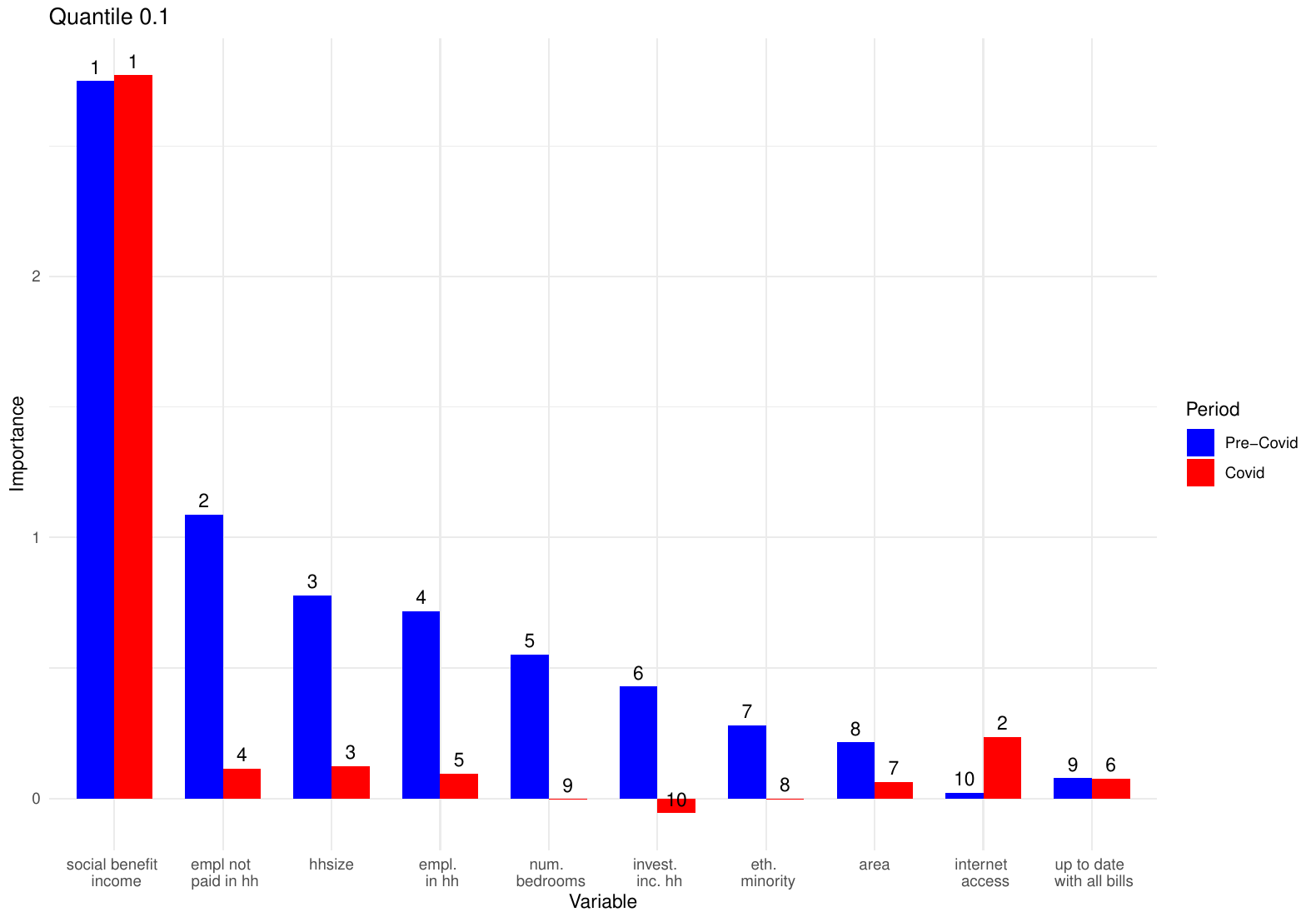}
    \caption{Bar plot showing the Variable Importance extracted from the FM-QRF for each covariate at quantile level $\tau=0.1$. The blue bars represent the Variable Importance in the Pre-pandemic period and the red bars are related to the pandemic period. Numbers at the top of the bars indicate the ranking position of each variable in terms of Variable Importance.}
    \label{fig:quantile01}
\end{figure}

\begin{figure}[H]
    \centering
    \includegraphics[width=0.8\textwidth]{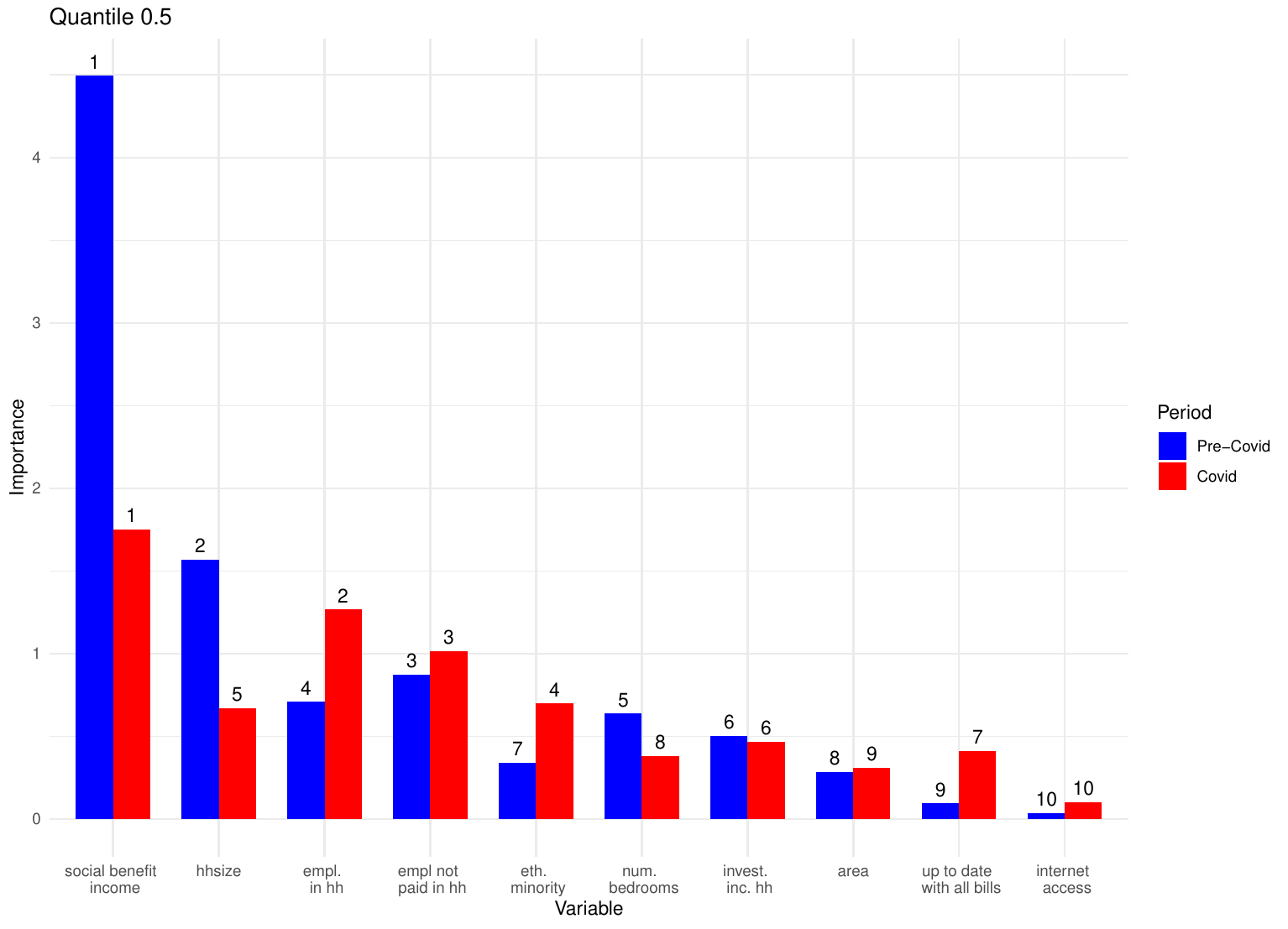}
 \caption{Bar plot showing the Variable Importance extracted from the FM-QRF for each covariate at quantile level $\tau=0.5$. The blue bars represent the Variable Importance in the Pre-pandemic period and the red bars are related to the pandemic period. Numbers at the top of the bars indicate the ranking position of each variable in terms of Variable Importance.}
    \label{fig:quantile05}
\end{figure}

\begin{figure}[H]
    \centering
    \includegraphics[width=0.8\textwidth]{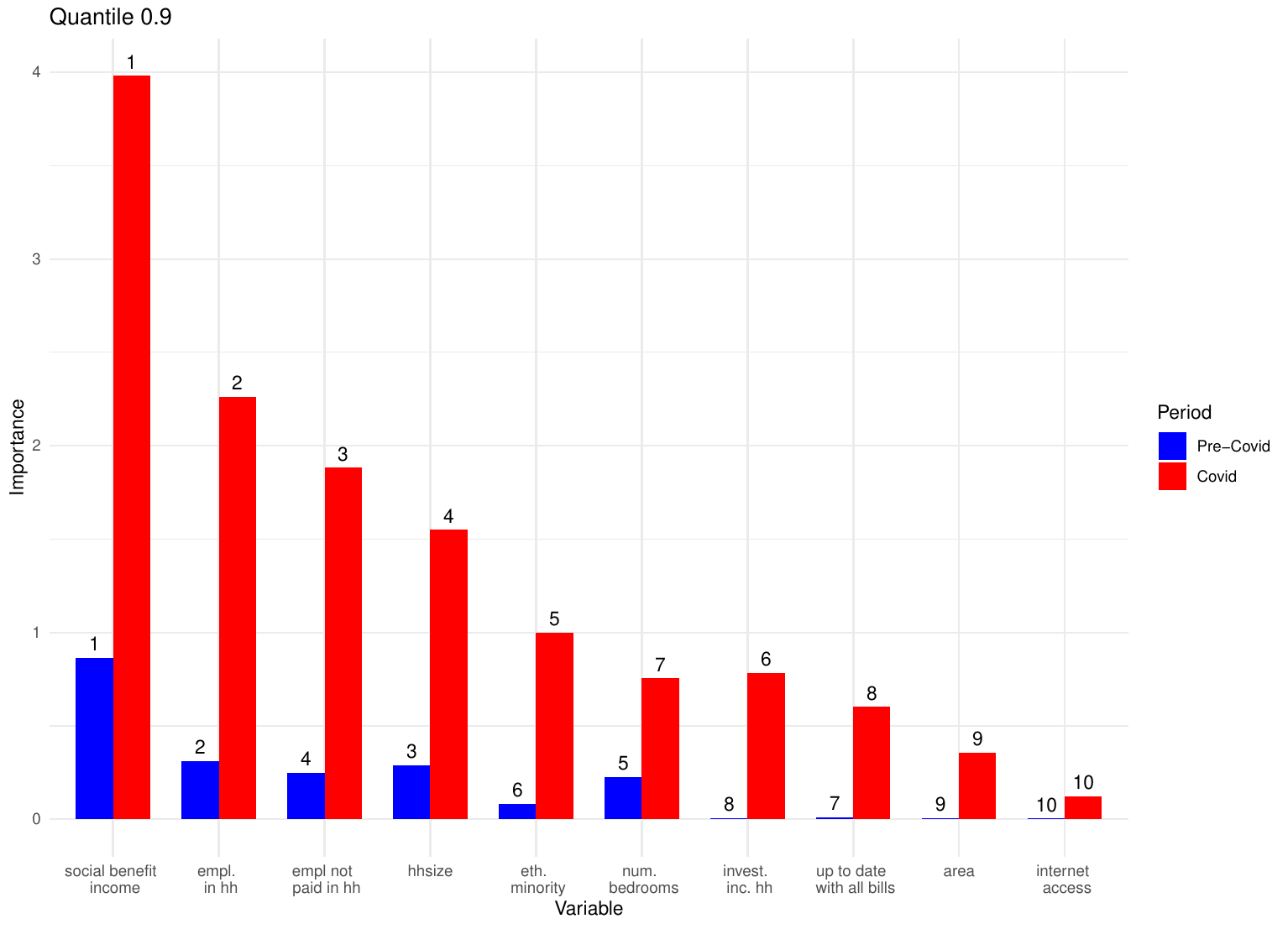}
     \caption{Bar plot showing the Variable Importance extracted from the FM-QRF for each covariate at quantile level $\tau=0.9$. The blue bars represent the Variable Importance in the pre-pandemic period and the red bars are related to the pandemic period. Numbers at the top of the bars indicate the ranking position of each variable in terms of Variable Importance.}
    \label{fig:quantile09}
\end{figure}

\vspace{0.15in}

\noindent At level $\tau=0.01$, the three most important variables in the pre-pandemic period are social benefit income, number of employed people not paid in the household, both proxies for family poverty, and household size.
During the COVID pandemic, social benefit income remains the most important variable, whereas the second most important one shifts from being the number of employed people not paid in the household to internet access, which is among the less important variables before the pandemic. This result highlights how during the pandemic the importance of internet access significantly increased for children with low levels of behavioural issues.

\vspace{0.15in}

\noindent At the median level $\tau=0.5$, before the pandemic the most important variable is social benefit income, the second important variable is household size and the third one is number of employed people not paid in the household. After the pandemic, the variable importance order changes only at the second place, which is covered by the number of employed people in the household.

\vspace{0.15in}

\noindent At the higher quantile level $\tau=0.9$, the first three most important variables before the pandemic are social benefit income, number of employed people in the household and household size. During the pandemic, the third most important variable is the number of employed people not being paid. Even if the order is quite similar before and during the pandemic, the main difference between these two periods is that the relevance of the three most important variables is sensibly higher during the pandemic. Being these three variables proxies for the level of family poverty, this result indicates that this factor became more important after the pandemic for children with a high grade of behavioural issues.

\vspace{0.15in}

\noindent In conclusion, in both periods, the most important variable for children with both high and low levels of behavioural issues is social benefit income. The result of this analysis highlights that during the pandemic the importance of internet access significantly increased during the pandemic in maintaining low levels of behavioural issues, whereas the relevance of family poverty increased for children with high levels of behavioural issues.
Moreover, it is also worth noticing that the variable importance changes across quantile levels, corroborating the use of the QR approach proposed in this chapter.

\subsection{Comparison with the LQMM Results}

This section reports the results obtained by fitting the LQMM model of Equation \eqref{eq:me-lqmm} with random intercept at the five quantile levels $\tau=0.1, 0.25, 0.5, 0.75, 0.9$. The aim of this analysis is to investigate whether the non-parametric approach of the FM-QRF might represent an additional valuable approach to investigate the risk factors determining children behavioural issues. 

\vspace{0.15in}

\noindent To this end, the sets of significant variables obtained with the LQMM models fitted for each quantile level are compared with the set of most important variables obtained from the FM-QRF using the Variable Importance measure. 
\vspace{0.15in}

\noindent The LQMM results coefficients for the pre-pandemic and pandemic period are reported in Table \ref{tab:lqmm_summary}.

\vspace{0.15in}

\noindent At quantile level $\tau=0.1$, in the pre-pandemic period the set of significant variables comprises variables proxies of the overcrowding in the household (number of beds and household size), the number of employed people in the household and ethnic minority. During the pandemic, the set of significant variables gets larger and includes also social benefit income and investment income. Similarly to results shown in previous contributions, the significant variables with the higher coefficient during the pandemic period are the employment variable and the ethnic minority dummy variable.

\vspace{0.15in}

\noindent At quantile level $\tau=0.5$, the set of significant variables remains almost the same in the pre-pandemic and pandemic periods. In the pre-pandemic period the significant variables are household size, number of beds, social benefit income, employed people in the household and ethnic minority. The main difference with the pandemic period is that during the COVID-19 pandemic the set of significant variables includes the investment income and does not include social benefit income and number of beds. 

\vspace{0.15in}

\noindent At quantile level $\tau=0.9$, in the pre-pandemic period the significant variables are social benefit income, number of employed people and ethnic minority. During the pandemic, the set of significant variables includes also household size and number of employed people not paid.

\vspace{0.15in}

\noindent Additionally, the pseudo-$R^2$ measure is computed for both models to evaluate their goodness of fit. This measure has been proposed in \cite{koenker1999goodness} and implemented in the quantile regression literature (see, for instance, \cite{bianchi2018estimation, borgoni2024semiparametric}) as a valid alternative to the standard $R^2$. In particular, the pseudo-$R^2$ is computed as:

\begin{equation}
R^2_{\rho}(\tau)=1-\frac{\sum_{i=1}^{n}\rho_{\tau}(e_{it})}{\sum_{i=1}^{n}\rho_{\tau}(\Tilde{e}_{it})}
\end{equation}

\noindent where $\rho_{\tau}(\cdot)$ is the quantile loss function of \cite{koenker1978regression}, $e_{it}$ are the standardized residuals of the full model trained with the whole set of covariates and
$\Tilde{e}_{it}$ are the standardized residuals under the null model, which considers only the intercept (the coefficients of the covariates are set to 0). The results concerning the goodness of fit of the LQMM and FM-QRF model in the pre-COVID and COVID periods are reported separately in Tables \ref{tab:r2-precovid} and \ref{tab:r2-covid}.

\begin{table}[]
\centering
\resizebox{\columnwidth}{!}{%
\begin{tabularx}{\textwidth}{*{6}{>{\centering\arraybackslash}X}}
\toprule
\multicolumn{6}{c}{\textbf{Pre-Pandemic}}                                  \\ \midrule
$\tau$       & \textbf{0.10} & \textbf{0.25} & \textbf{0.50} & \textbf{0.75} & \textbf{0.90} \\
\textbf{LQMM}   & 2.85        & 3.58        & 4.70        & 6.64        & 8.41        \\
\textbf{FM-QRF} & 6.56        & 21.96       & 13.29       & 41.88       & 66.47       \\ \bottomrule
\end{tabularx}%
}
\caption{Pseudo-$R^2$ values related to the pre-pandemic period for the LQMM and the FM-QRF models. Values are expressed in percentages.}
\label{tab:r2-precovid}
\end{table}

\begin{table}[]
\centering
\resizebox{\columnwidth}{!}{%
\begin{tabularx}{\textwidth}{*{6}{>{\centering\arraybackslash}X}}
\toprule
\multicolumn{6}{c}{\cellcolor[HTML]{FFFFFF}\textbf{Pandemic}}                                  \\ \midrule
$\tau $  & \textbf{0.10} & \textbf{0.25} & \textbf{0.50} & \textbf{0.75} & \textbf{0.90} \\
\textbf{LQMM}   & 0.46        & 1.96        & 2.08        & 4.74        & 10.58       \\
\textbf{FM-QRF} & 0.86        & 16.64       & 8.57        & 71.82       & 35.07       \\ \bottomrule
\end{tabularx}%
}
\caption{Pseudo-$R^2$ values related to the pandemic period for the LQMM and the FM-QRF models. Values are expressed in percentages.}
\label{tab:r2-covid}
\end{table}

\noindent The values of the pseudo-$R^2$ highlight that the FM-QRF has a greater goodness of fit with respect to the LQMM at all quantile levels both in the pre-COVID and COVID periods. Moreover, as already noted in \cite{} for other kind of models, the pseudo-$R^2$ increases with the quantile level of interest for both the FM-QRF and the LQMM.

\vspace{0.15in}

\noindent From these results a variety of conclusions can be drawn. 
First, similarly to the FM-QRF analysis results, the significant variables set and the related coefficient values change across quantile levels. This result further justify the relevance of a QR approach. However, the ethnic minority variable and number of employed people in the household are significant at all quantile levels in the pandemic period and their coefficient values are similar across quantiles.

\vspace{0.15in}

\noindent Second, the set of significant variables obtained with the LQMM approach does not always coincide with the set of most important variables obtained with the FM-QRF.

\vspace{0.15in}

\noindent For instance, at quantile level $\tau=0.1$, the internet access variable is the second most important variable in the FM-QRF analysis, whereas in the LQMM analysis it is not significant neither in the pre-pandemic period nor during the pandemic. 

\vspace{0.15in}

\noindent These results highlight how a non-parametric approach might be useful to uncover meaningful non-linear relationships among variables that are being overlooked with a linear approach.

\begin{table}[H]
\centering
\caption{LQMM results coefficients for the pre-pandemic and the pandemic period at five quantile levels $\tau=0.1, 0.5, 0.9$. The symbol '***' denotes significance at 1\% level and '**'  significance at 5\%.}
\label{tab:lqmm_summary}
\resizebox{\textwidth}{!}{%
\begin{tabular}{@{}ccccclcccccl@{}}
\toprule
 & \multicolumn{5}{c}{\textbf{Pre-Pandemic}} &  & \multicolumn{5}{c}{\textbf{Pandemic}} \\ \midrule
 & \textit{Variable} & \textit{Estimate} & \textit{Std. Error} & \multicolumn{2}{c}{\textit{P-Value}} &  & \textit{Variable} & \textit{Estimate} & \textit{Std. Error} & \multicolumn{2}{c}{\textit{P-Value}} \\ \cmidrule(lr){2-6} \cmidrule(l){8-12} 
\multirow{11}{*}{\textbf{0.1}} & \cellcolor[HTML]{D9D9D9} Intercept & 11.153 & (1.681) & 0.000& *** &  & Intercept & 5.447 & (4.517) & 0.234 &  \\
 & area & -0.036 & (0.173) & 0.838 &  &  & area & 0.055 & (0.136) & 0.690 &  \\
 & \cellcolor[HTML]{D9D9D9} hhsize & -0.887 & (0.155) & 0.000& *** &  & \cellcolor[HTML]{D9D9D9} hhsize & -0.966 & (0.25) & 0.000& *** \\
 & \cellcolor[HTML]{D9D9D9} n\_beds & -0.759 & (0.145) & 0.000& *** &  & \cellcolor[HTML]{D9D9D9} n\_beds & -0.437 & (0.209) & 0.042 & ** \\
 & soc\_ben\_inc & -0.001 & (0.001) & 0.106 &  &  & \cellcolor[HTML]{D9D9D9} soc\_ben\_inc & -0.002 & (0.001) & 0.021 & ** \\
 & invest\_inc & -0.001 & (0.001) & 0.254 &  &  & \cellcolor[HTML]{D9D9D9} invest\_inc & -0.002 & (0.001) & 0.002 & *** \\
 & empl\_not\_paid & -0.241 & (0.272) & 0.380 &  &  & empl\_not\_paid & 0.530 & (0.484) & 0.278 &  \\
 & internet\_access & -0.387 & (1.204) & 0.750 &  &  & internet\_access & 4.197 & (4.503) & 0.356 &  \\
 & \cellcolor[HTML]{D9D9D9} empl & 0.726 & (0.284) & 0.014 & ** &  & \cellcolor[HTML]{D9D9D9} empl & 1.337 & (0.413) & 0.002 & *** \\
 & bills & 0.325 & (0.499) & 0.519 &  &  & bills & 0.011 & (0.59) & 0.985 &  \\
 & \cellcolor[HTML]{D9D9D9} eth\_min & -1.696 & (0.288) & 0.000& *** &  & \cellcolor[HTML]{D9D9D9} eth\_min & -1.663 & (0.683) & 0.019 & ** \\
 \midrule
\multirow{11}{*}{\textbf{0.5}} & \cellcolor[HTML]{D9D9D9} Intercept & 11.236 & (1.778) & 0.000& *** &  & Intercept & 5.517 & (5.37) & 0.309 &  \\
 & area & 0.065 & (0.183) & 0.723 &  &  & area & 0.138 & (0.126) & 0.279 &  \\
 & \cellcolor[HTML]{D9D9D9} hhsize & -0.575 & (0.154) & 0.000& *** &  & \cellcolor[HTML]{D9D9D9} hhsize & -0.699 & (0.247) & 0.007 & *** \\
 & \cellcolor[HTML]{D9D9D9} n\_beds & -0.495 & (0.146) & 0.001 & *** &  & n\_beds & -0.204 & (0.237) & 0.392 &  \\
 & \cellcolor[HTML]{D9D9D9} soc\_ben\_inc & 0.001 & (0.000) & 0.000& *** &  & soc\_ben\_inc & 0.001 & (0.000) & 0.145 &  \\
 & invest\_inc & 0.000& (0.000) & 0.135 &  &  & \cellcolor[HTML]{D9D9D9} invest\_inc & 0.000& (0.000) & 0.026 & ** \\
 & empl\_not\_paid & -0.088 & (0.249) & 0.724 &  &  & empl\_not\_paid & 0.656 & (0.418) & 0.122 &  \\
 & internet\_access & -0.303 & (1.483) & 0.839 &  &  & internet\_access & 4.267 & (5.078) & 0.405 &  \\
 & \cellcolor[HTML]{D9D9D9} empl & 0.735 & (0.325) & 0.028 & ** &  & \cellcolor[HTML]{D9D9D9} empl & 1.346 & (0.386) & 0.001 & *** \\
 & bills & 0.409 & (0.41) & 0.324 &  &  & bills & 0.084 & (0.743) & 0.910 &  \\
 & \cellcolor[HTML]{D9D9D9} eth\_min & -1.693 & (0.4) & 0.000& *** &  & \cellcolor[HTML]{D9D9D9} eth\_min & -1.662 & (0.591) & 0.007 & *** \\
\midrule
\multirow{11}{*}{\textbf{0.9}} & \cellcolor[HTML]{D9D9D9} Intercept & 11.285 & (1.531) & 0.000& *** &  & Intercept & 5.542 & (5.511) & 0.320 &  \\
 & area & 0.129 & (0.22) & 0.560 &  &  & area & 0.168 & (0.15) & 0.268 &  \\
 & hhsize & -0.385 & (0.217) & 0.083 &  &  & \cellcolor[HTML]{D9D9D9} hhsize & -0.601 & (0.21) & 0.006 & *** \\
 & n\_beds & -0.335 & (0.196) & 0.093 &  &  & n\_beds & -0.122 & (0.203) & 0.550 &  \\
 & \cellcolor[HTML]{D9D9D9} soc\_ben\_inc & 0.007 & (0.001) & 0.000& *** &  & \cellcolor[HTML]{D9D9D9} soc\_ben\_inc & 0.008 & (0.001) & 0.000& *** \\
 & invest\_inc & 0.001 & (0.001) & 0.417 &  &  & invest\_inc & 0.001 & (0.001) & 0.151 &  \\
 & empl\_not\_paid & 0.006 & (0.283) & 0.983 &  &  & \cellcolor[HTML]{D9D9D9} empl\_not\_paid & 0.704 & (0.345) & 0.047 & ** \\
 & internet\_access & -0.253 & (1.433) & 0.860 &  &  & internet\_access & 4.292 & (5.184) & 0.412 &  \\
 & \cellcolor[HTML]{D9D9D9} empl & 0.740 & (0.325) & 0.027 & ** &  & \cellcolor[HTML]{D9D9D9} empl & 1.350 & (0.385) & 0.001 & *** \\
 & bills & 0.458 & (0.403) & 0.261 &  &  & bills & 0.110 & (0.727) & 0.880 &  \\
 & \cellcolor[HTML]{D9D9D9} eth\_min & -1.690 & (0.365) & 0.000& *** &  & \cellcolor[HTML]{D9D9D9} eth\_min & -1.660 & (0.662) & 0.015 & ** \\ \cmidrule(l){2-12} 
\end{tabular}%
}
\end{table}

\section{Conclusions}
\label{sec:SDQ-conclusions}

The COVID-19 pandemic has significantly affected different aspects of society, including children's mental health. This chapter investigates the risk factors driving children's mental health before and during the pandemic using data from the UK Household Longitudinal Study. The analysis employs the novel machine learning algorithm FM-QRF of Chapter \ref{ch:FM-QRF} to model the complex relationship between pandemic-related factors and children's mental health, measured with the well-known SDQ score.

\vspace{0.15in}

\noindent The empirical findings reveal that the drivers of children's mental health differ between those with low and high SDQ scores, and that these drivers also vary before and during the pandemic. Moreover, by studying the SDQ conditional distribution at different quantile levels, this study provides a deeper understanding of the impact of the pandemic on children's mental health outcomes with respect to standard linear approached.

\vspace{0.15in}

\noindent Key findings indicate that social benefit income variable remains a crucial factor across different quantile levels in both pre-pandemic and pandemic periods. Additionally, the importance of internet access significantly increased during the pandemic, especially for children with lower levels of behavioral issues. The study also highlights the higher relevance of family poverty during the pandemic for children with higher levels of behavioral issues.

\vspace{0.15in}

\noindent The comparison with the LQMM model results highlights the relevance of the non-parametric approach employed in this study. As a matter of fact, the FM-QRF  offers an additional approach to gain a more comprehensive understanding of the complex dynamics affecting children's mental health, since it reveals non-linear relationships among variables that are overlooked by linear models, such as the LQMM.

\vspace{0.15in}

\noindent In conclusion, this research contributes to the growing body of literature addressing the impact of the COVID-19 pandemic on children's mental health. The use of non-linear approaches like FM-QRF enhances the depth of the analysis, providing valuable insights for policymakers, educators, and healthcare professionals working to support and improve children's well-being.


\begingroup
\bibliographystyle{apalike} 
\bibliography{PhDThesisBib}

\appendix
\chapter{Mixed-Frequency Quantile
Regression Forests}\label{app:MIDAS-QRF}

This section reports additional Figures and Tables from Chapter \ref{ch:MIDAS-QRF}.

\begin{table}[h]
\centering
\begin{tabularx}{\textwidth}{*{8}{>{\centering\arraybackslash}X}}
\toprule
            & \textit{Obs}  & \textit{Min}     & \textit{Max}    & \textit{Mean}  & \textit{SD }   & \textit{Skew.}  & \textit{Kurt.}  \\ \midrule
WTI         & 2053 & -28.138 & 42.583 & 0.042 & 3.275 & 1.177  & 39.354 \\
BRENT       & 2053 & -25.639 & 41.202 & 0.036 & 2.922 & 1.107  & 39.192 \\
HEAT. & 2053 & -22.314 & 14.862 & 0.022 & 2.511 & -0.416 & 12.046 \\ 
SP500 & 2053 & -12.765 & 8.968 & 0.047 & 1.095 & -1.039 & 23.955  \\

DOLL & 92 &  4.429 & 4.747 & 4.599 & 0.090 & -0.439 & 1.638 \\

NATGAS & 31 &   -0.219 & 0.605 & 0.158 & 0.166 & 0.284 & 3.158 \\

SAUDI\_P. & 31 &   -1.313 & 1.775 & 0.0242 & 0.714 &  0.350 & 2.746 \\

\bottomrule
\end{tabularx}
\caption{Summary statistics of the variables included in the sample. The table reports the number of observations (Obs), the minimum (Min), maximum (Max) along with the mean, standard deviation (SD), skewness and excess kurtosis.}
\label{tab:sum-stats}
\end{table}

\begin{figure}[b]
    \centering
\includegraphics[width=\textwidth]{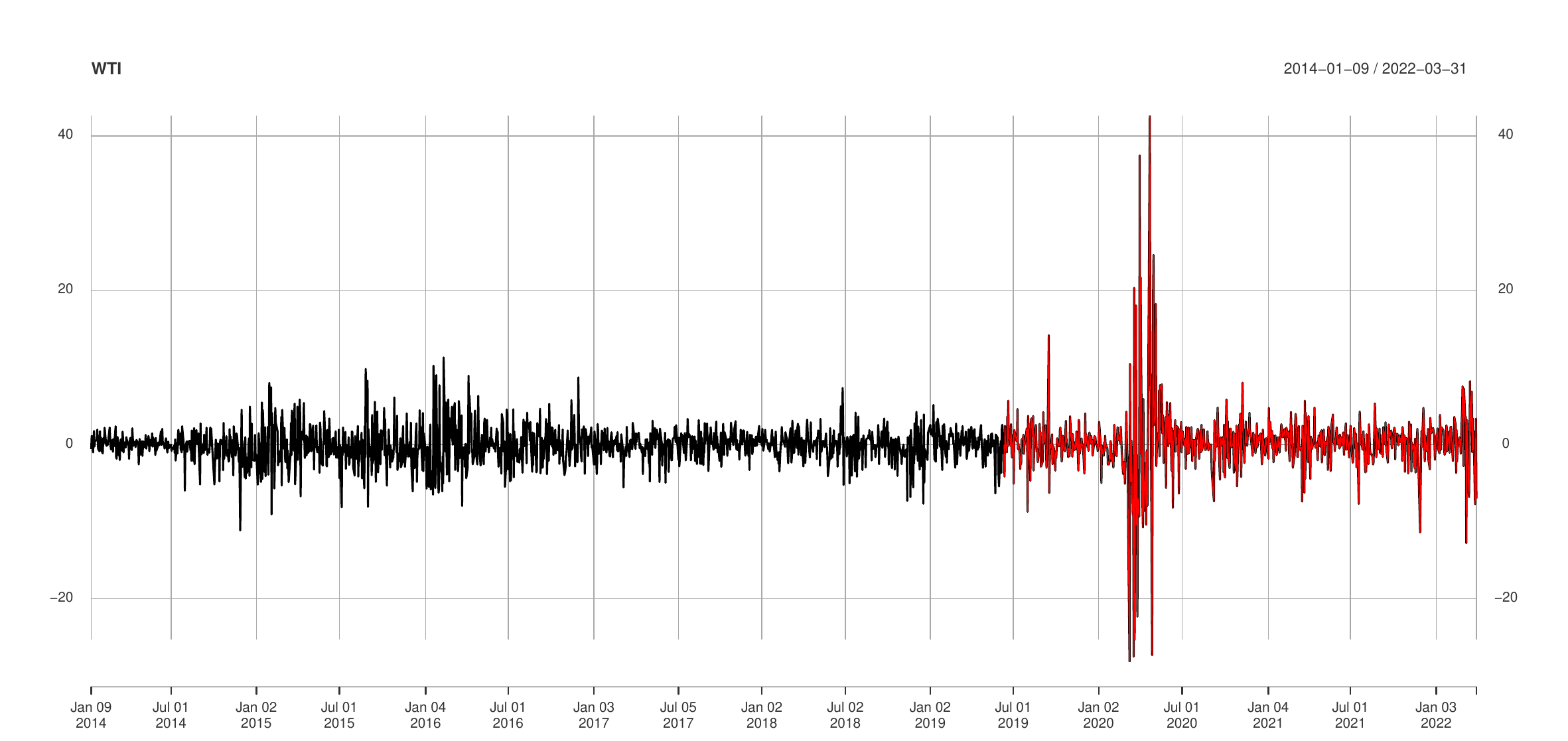}
    \caption{WTI index time series. the black line represents the training set and the red line the out-of-sample period.}
    \label{fig:wti}
\end{figure}

\begin{figure}[h]
    \centering \includegraphics[width=\textwidth]{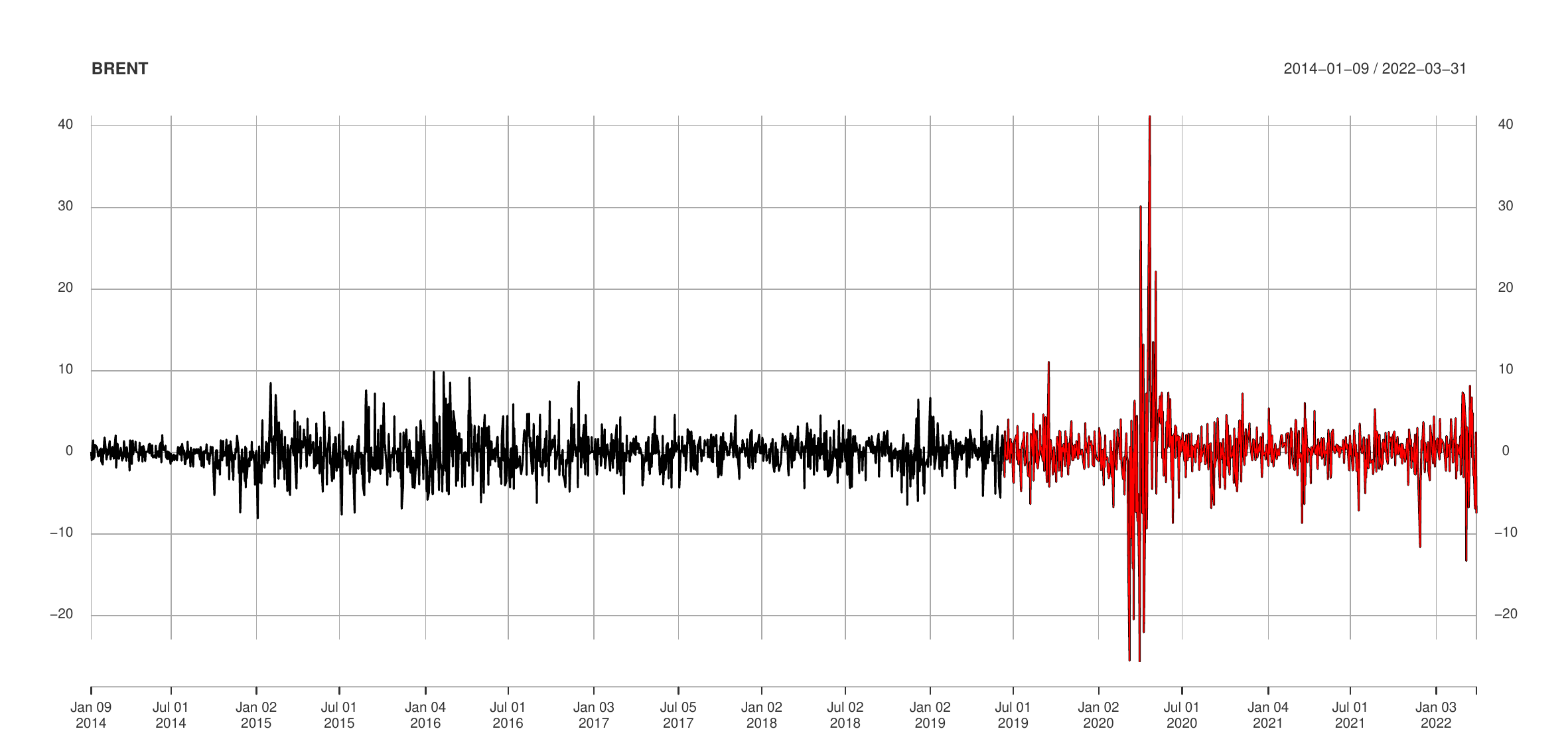}
    \caption{Brent index time series. the black line represents the training set and the red line the out-of-sample period.}
    \label{fig:brent}
\end{figure}

\begin{figure}[h]
    \centering
    \includegraphics[width=\textwidth]{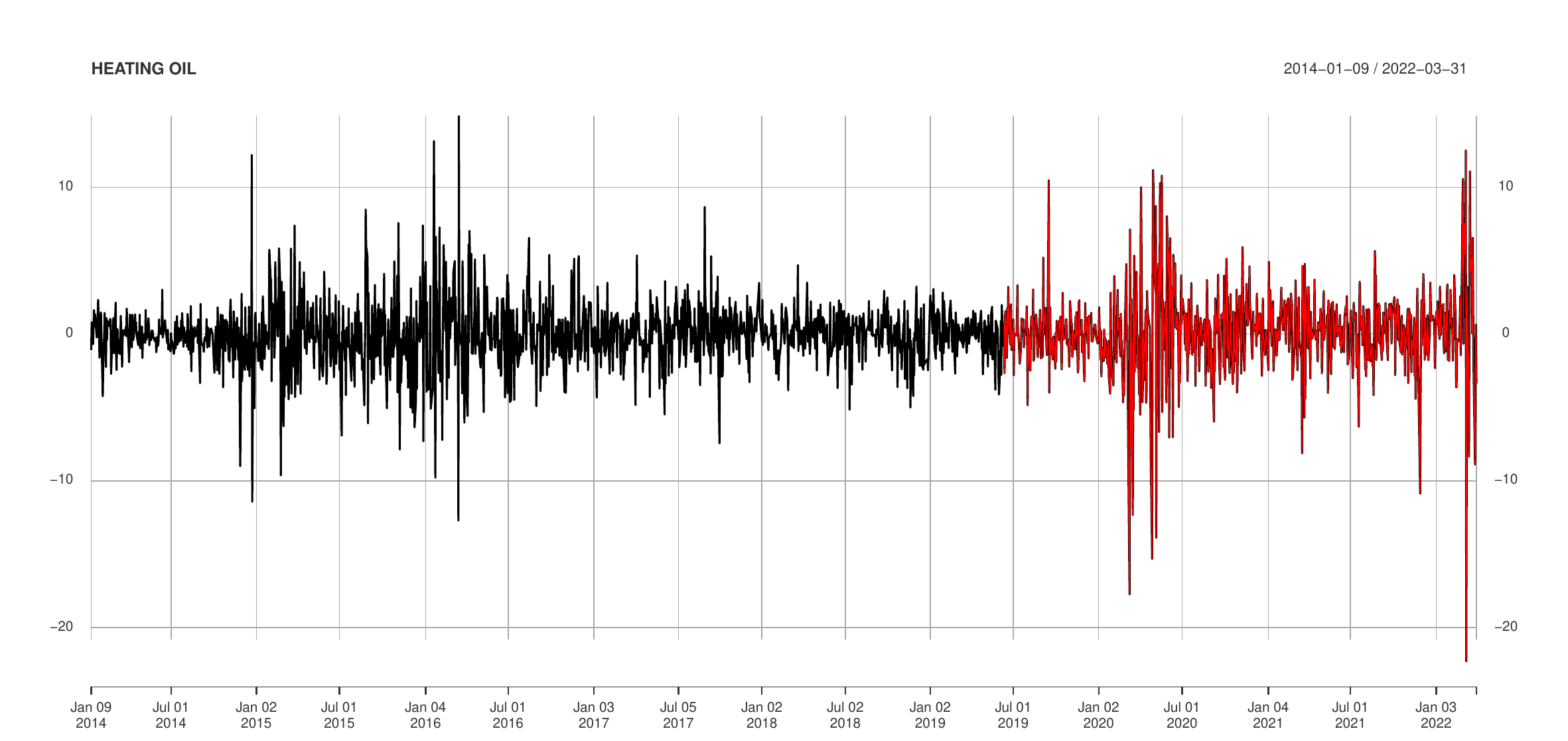}
    \caption{Heating oil index time series. the black line represents the training set and the red line the out-of-sample period.}
    \label{fig:heat}
\end{figure}

\begin{figure}[h]
    \centering
    \includegraphics[width=\textwidth]{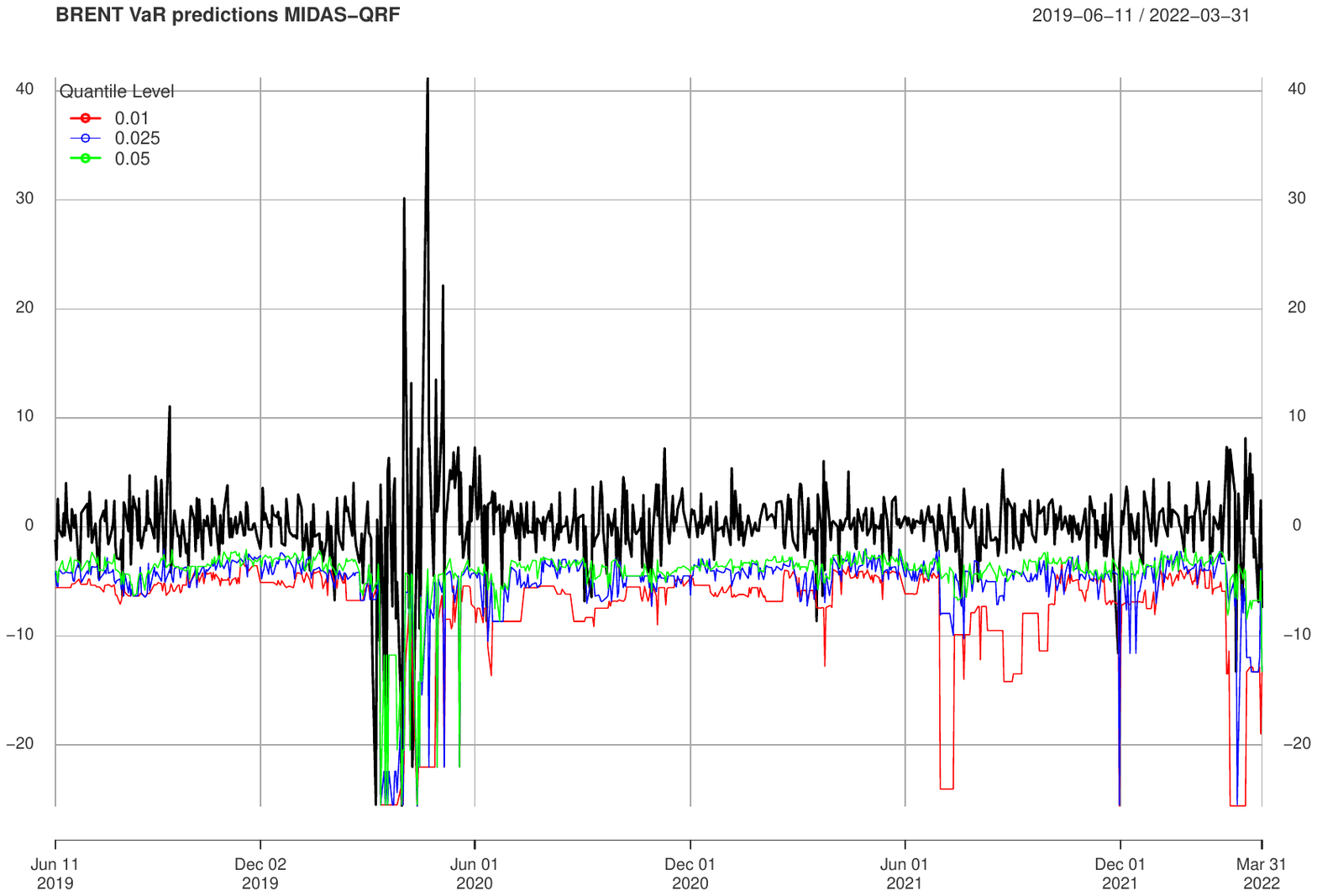}
\includegraphics[width=\textwidth]{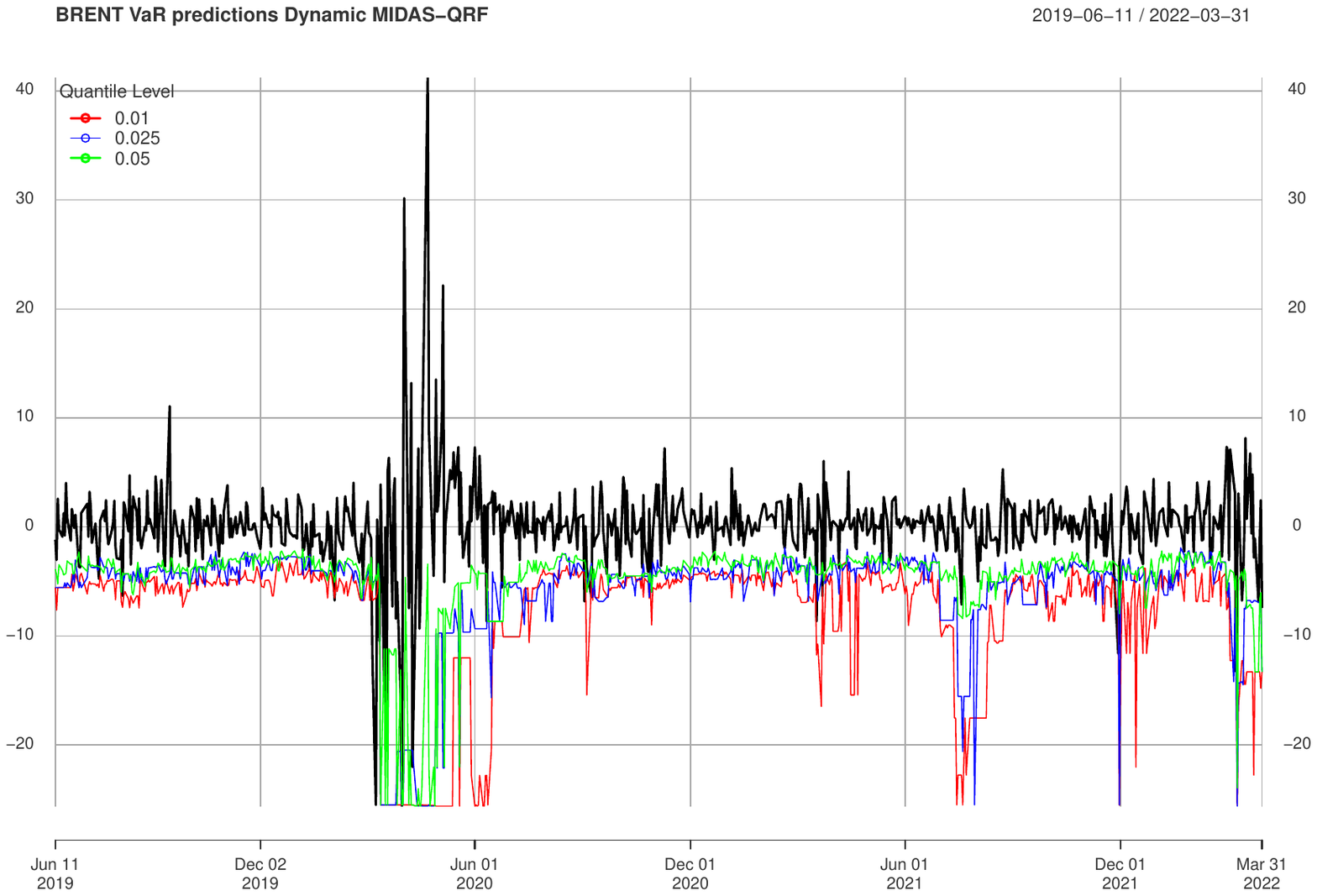}
    \caption{Brent index (black line) out-of-sample predictions at quantile levels $\tau= 0.01, 0.025, 0.05$. The top panel and the bottom panel show the predictions obtained with the dynamic MIDAS-QRF model, respectively.}
    \label{fig:brent-pred}
\end{figure}

\begin{figure}[h]
    \centering
    \includegraphics[width=\textwidth]{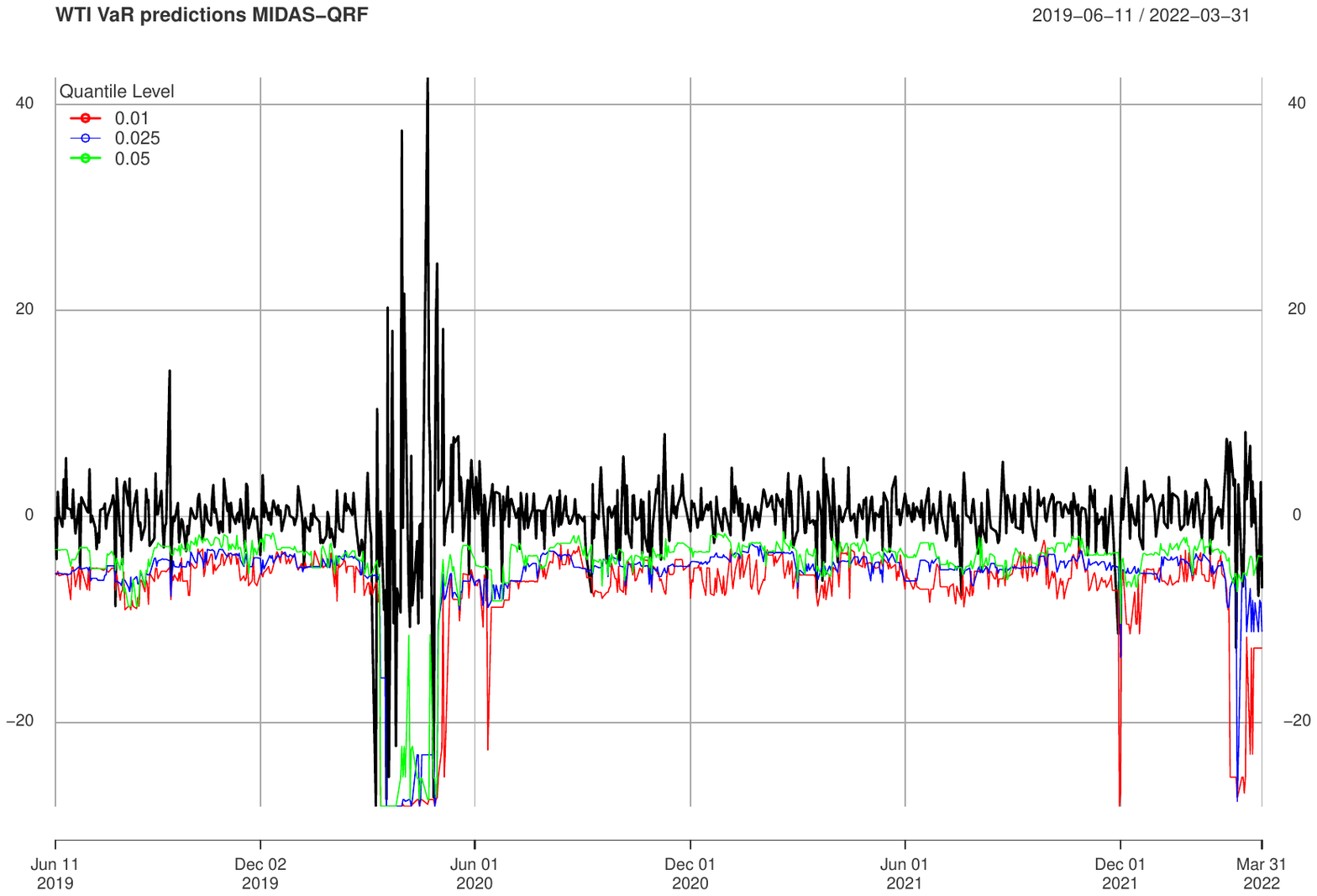}
\includegraphics[width=\textwidth]{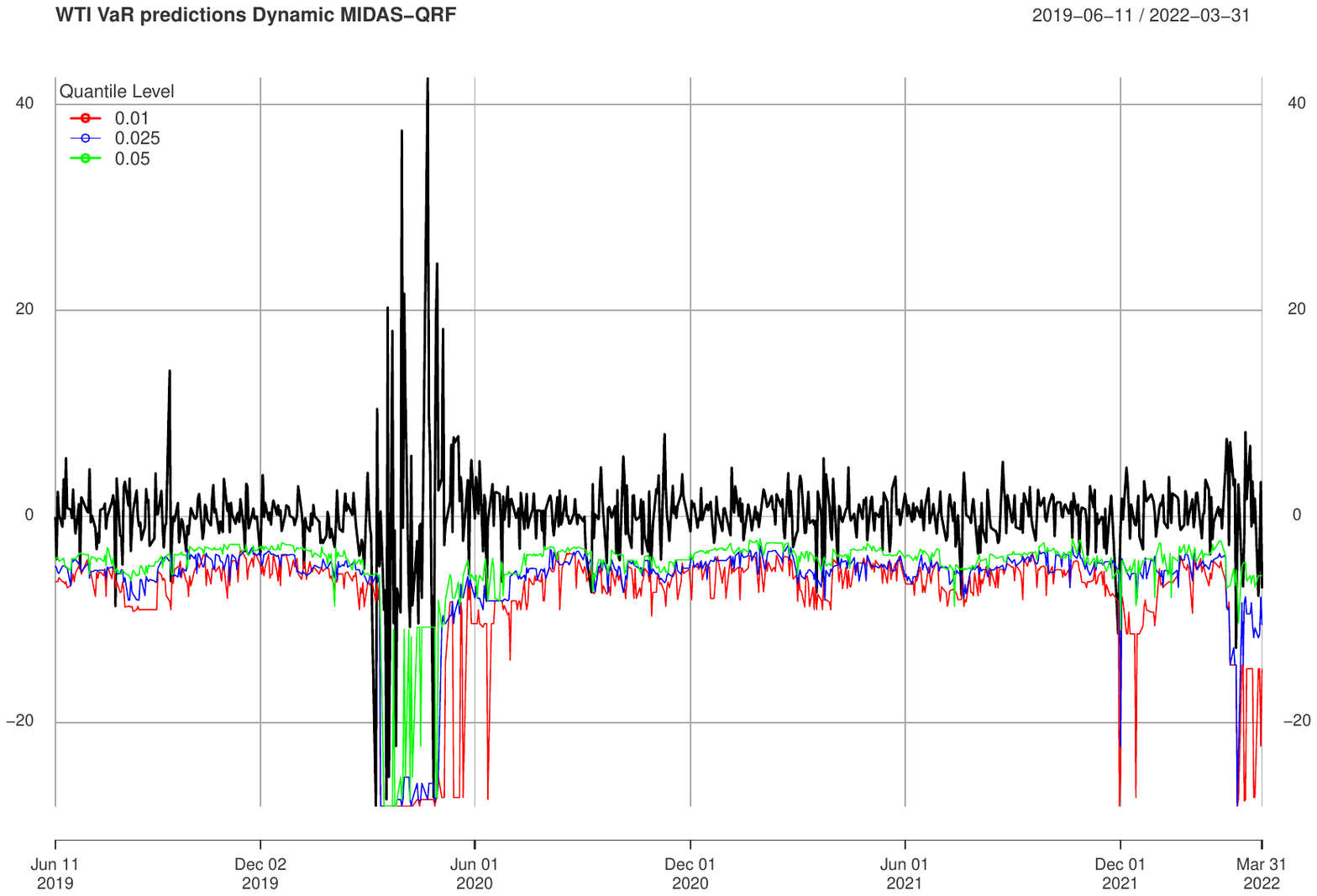}
    \caption{WTI index (black line) out-of-sample predictions at quantile levels $\tau= 0.01, 0.025, 0.05$. The top panel and the bottom panel show the predictions obtained with the dynamic MIDAS-QRF model, respectively.}
    \label{fig:wti-pred}
\end{figure}

\begin{table}[]
\centering
\resizebox{\columnwidth}{!}{%
\begin{tabularx}{\textwidth}{*{5}{>{\centering\arraybackslash}X}}
\toprule
\multicolumn{5}{c}{\textbf{WTI}}                                                                                               \\ \midrule
                                  & $\tau$         & \textit{\textbf{0.01}} & \textit{\textbf{0.025}} & \textit{\textbf{0.05}} \\
\multirow{2}{*}{\textit{Static}}  & \textit{0.01}  & \multicolumn{1}{c}{-}  & 0.21                    & 0.04                   \\
                                  & \textit{0.025} &                        & \multicolumn{1}{c}{-}   & 0.12                   \\ \midrule
\multirow{2}{*}{\textit{Dynamic}} & \textit{0.01}  & \multicolumn{1}{c}{-}  & 0.12                    & 0.01                   \\
                                  & \textit{0.025} &                        & \multicolumn{1}{c}{-}   & 0.07                   \\ \midrule
\multicolumn{5}{c}{\textbf{BRENT}}                                                                                             \\ \midrule
                                  & $\tau$         & \textit{\textbf{0.01}} & \textit{\textbf{0.025}} & \textit{\textbf{0.05}} \\
\multirow{2}{*}{\textit{Static}}  & \textit{0.01}  & \multicolumn{1}{c}{-}  & 0.08                    & 0.007                  \\
                                  & \textit{0.025} &                        & \multicolumn{1}{c}{-}   & 0.16                   \\ \midrule
\multirow{2}{*}{\textit{Dynamic}} & \textit{0.01}  & \multicolumn{1}{c}{-}  & 0.158                   & 0.01                   \\
                                  & \textit{0.025} &                        & \multicolumn{1}{c}{-}   & 0.24                   \\ \midrule
\multicolumn{5}{c}{\textbf{HEATING OIL}}                                                                                       \\ \midrule
                                  & $\tau$         & \textit{\textbf{0.01}} & \textit{\textbf{0.025}} & \textit{\textbf{0.05}} \\
\multirow{2}{*}{\textit{Static}}  & \textit{0.01}  & \multicolumn{1}{c}{-}  & 0.08                    & 0.01                   \\
                                  & \textit{0.025} &                        & \multicolumn{1}{c}{-}   & 0.11                   \\ \midrule
\multirow{2}{*}{\textit{Dynamic}} & \textit{0.01}  & \multicolumn{1}{c}{-}  & 0.14                    & 0.02                   \\
                                  & \textit{0.025} &                        & \multicolumn{1}{c}{-}   & 0.11                   \\ \bottomrule
\end{tabularx}%
}
\caption{Ratio between the number of times quantiles computed at level $\tau$ indicated in the rows are higher than those computed at level $\tau$ in the columns.}
\label{tab:quants}
\end{table}

\chapter{Finite mixtures of Quantile
Regression Forests and their
application to GDP growth-at-risk
from climate change}

This section reports the mean (in bold) and the standard error of the quantile estimates at level $\tau=0.01, 0.5, 0.99$ for each country for the SSP1 and SSP5 scenario. The estimates are obtained via bootstrapping with 500 iterations. Each table reports the values for the years 2030, 2050, 2100.

\begin{landscape}
    
\footnotesize
\begin{longtable}[c]{
>{\columncolor[HTML]{D4D4D4}}l llllllllllllllllll}
\hline
\cellcolor[HTML]{B0B3B2}                     & \multicolumn{6}{c}{\cellcolor[HTML]{B0B3B2}\textbf{0.01}}                                                                                                                         & \multicolumn{6}{c}{\cellcolor[HTML]{B0B3B2}\textbf{0.5}}                                                                                                                          & \multicolumn{6}{c}{\cellcolor[HTML]{B0B3B2}\textbf{0.99}}                                                                                                                         \\ \hline
\endfirsthead
\multicolumn{19}{c}%
{{\bfseries Table \thetable\ continued from previous page}} \\
\hline
\cellcolor[HTML]{B0B3B2}                     & \multicolumn{6}{c}{\cellcolor[HTML]{B0B3B2}\textbf{0.01}}                                                                                                                         & \multicolumn{6}{c}{\cellcolor[HTML]{B0B3B2}\textbf{0.5}}                                                                                                                          & \multicolumn{6}{c}{\cellcolor[HTML]{B0B3B2}\textbf{0.99}}                                                                                                                         \\ \hline
\endhead
\hline
\endfoot
\endlastfoot
\multicolumn{1}{c}{\cellcolor[HTML]{B0B3B2}} & \multicolumn{2}{c}{\cellcolor[HTML]{B0B3B2}\textbf{2030}} & \multicolumn{2}{c}{\cellcolor[HTML]{B0B3B2}\textbf{2050}} & \multicolumn{2}{c}{\cellcolor[HTML]{B0B3B2}\textbf{2100}} & \multicolumn{2}{c}{\cellcolor[HTML]{B0B3B2}\textbf{2030}} & \multicolumn{2}{c}{\cellcolor[HTML]{B0B3B2}\textbf{2050}} & \multicolumn{2}{c}{\cellcolor[HTML]{B0B3B2}\textbf{2100}} & \multicolumn{2}{c}{\cellcolor[HTML]{B0B3B2}\textbf{2030}} & \multicolumn{2}{c}{\cellcolor[HTML]{B0B3B2}\textbf{2050}} & \multicolumn{2}{c}{\cellcolor[HTML]{B0B3B2}\textbf{2100}} \\
\textbf{AFG}                                 & \textbf{-4.56}                    & 0.39                  & \textbf{-4.32}                    & 0.4                   & \textbf{-4.83}                    & 0.51                  & \textbf{6.51}                    & 0.22                   & \textbf{6.44}                    & 0.24                   & \textbf{6.38}                    & 0.22                   & \textbf{21.25}                   & 0.72                   & \textbf{20.99}                   & 0.66                   & \textbf{21.1}                    & 0.69                   \\
\textbf{ALB}                                 & \textbf{-4.49}                    & 0.53                  & \textbf{-4.51}                    & 0.53                  & \textbf{-3.88}                    & 0.38                  & \textbf{3.99}                    & 0.2                    & \textbf{3.99}                    & 0.16                   & \textbf{4.26}                    & 0.17                   & \textbf{9.33}                    & 0.32                   & \textbf{9.7}                     & 0.35                   & \textbf{9.77}                    & 0.32                   \\
\end{longtable}
\end{landscape}

\begin{landscape}
\footnotesize
\begin{longtable}[c]{
>{\columncolor[HTML]{D4D4D4}}l llllllllllllllllll}
\hline
\cellcolor[HTML]{B0B3B2}                     & \multicolumn{6}{c}{\cellcolor[HTML]{B0B3B2}\textbf{0.01}}                                                                                                                         & \multicolumn{6}{c}{\cellcolor[HTML]{B0B3B2}\textbf{0.5}}                                                                                                                          & \multicolumn{6}{c}{\cellcolor[HTML]{B0B3B2}\textbf{0.99}}                                                                                                                         \\ \hline
\endfirsthead
\multicolumn{19}{c}%
{{\bfseries Table \thetable\ continued from previous page}} \\
\hline
\cellcolor[HTML]{B0B3B2}                     & \multicolumn{6}{c}{\cellcolor[HTML]{B0B3B2}\textbf{0.01}}                                                                                                                         & \multicolumn{6}{c}{\cellcolor[HTML]{B0B3B2}\textbf{0.5}}                                                                                                                          & \multicolumn{6}{c}{\cellcolor[HTML]{B0B3B2}\textbf{0.99}}                                                                                                                         \\ \hline
\endhead
\hline
\endfoot
\endlastfoot
\multicolumn{1}{c}{\cellcolor[HTML]{B0B3B2}} & \multicolumn{2}{c}{\cellcolor[HTML]{B0B3B2}\textbf{2030}} & \multicolumn{2}{c}{\cellcolor[HTML]{B0B3B2}\textbf{2050}} & \multicolumn{2}{c}{\cellcolor[HTML]{B0B3B2}\textbf{2100}} & \multicolumn{2}{c}{\cellcolor[HTML]{B0B3B2}\textbf{2030}} & \multicolumn{2}{c}{\cellcolor[HTML]{B0B3B2}\textbf{2050}} & \multicolumn{2}{c}{\cellcolor[HTML]{B0B3B2}\textbf{2100}} & \multicolumn{2}{c}{\cellcolor[HTML]{B0B3B2}\textbf{2030}} & \multicolumn{2}{c}{\cellcolor[HTML]{B0B3B2}\textbf{2050}} & \multicolumn{2}{c}{\cellcolor[HTML]{B0B3B2}\textbf{2100}} \\
\textbf{AFG}                                 & \textbf{-4.54}                    & 0.46                  & \textbf{-4.73}                    & 0.35                  & \textbf{-4.28}                    & 0.36                  & \textbf{6.49}                    & 0.23                   & \textbf{6.26}                    & 0.24                   & \textbf{6.27}                    & 0.24                   & \textbf{21.11}                   & 0.7                    & \textbf{21.76}                   & 0.77                   & \textbf{21.21}                   & 0.76                   \\
\textbf{ALB}                                 & \textbf{-3.96}                    & 0.4                   & \textbf{-4.42}                    & 0.47                  & \textbf{-5.01}                    & 1                     & \textbf{4.41}                    & 0.15                   & \textbf{4.2}                     & 0.17                   & \textbf{4.4}                     & 0.16                   & \textbf{9.68}                    & 0.33                   & \textbf{9.57}                    & 0.32                   & \textbf{9.49}                    & 0.33                   \\
\end{longtable}
\end{landscape}

\chapter{The Impact of the COVID-19 Pandemic on Risk Factors for Children’s Mental Health}\label{app:SDQ}

This section reports additional Figures from Chapter \ref{ch:SDQ}.

\begin{figure}[h]
    \centering
\includegraphics[width=0.75\textwidth]{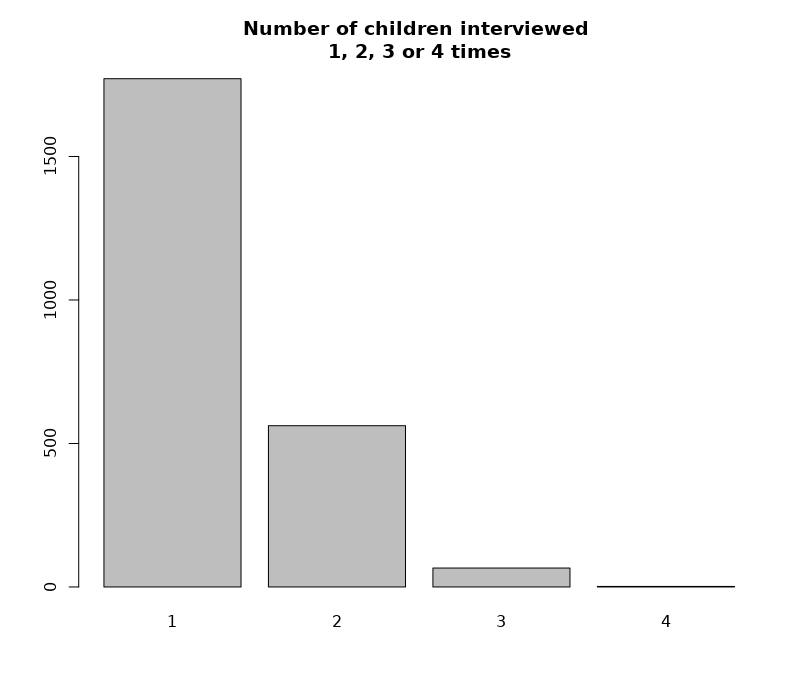}
    \caption{Bar plot showing the number of children interviewed 1, 2, 3 or 4 times during the sample period. }
    \label{fig:n_children}
\end{figure}

\begin{figure}[h]
    \centering
\includegraphics[width=0.93\textwidth]{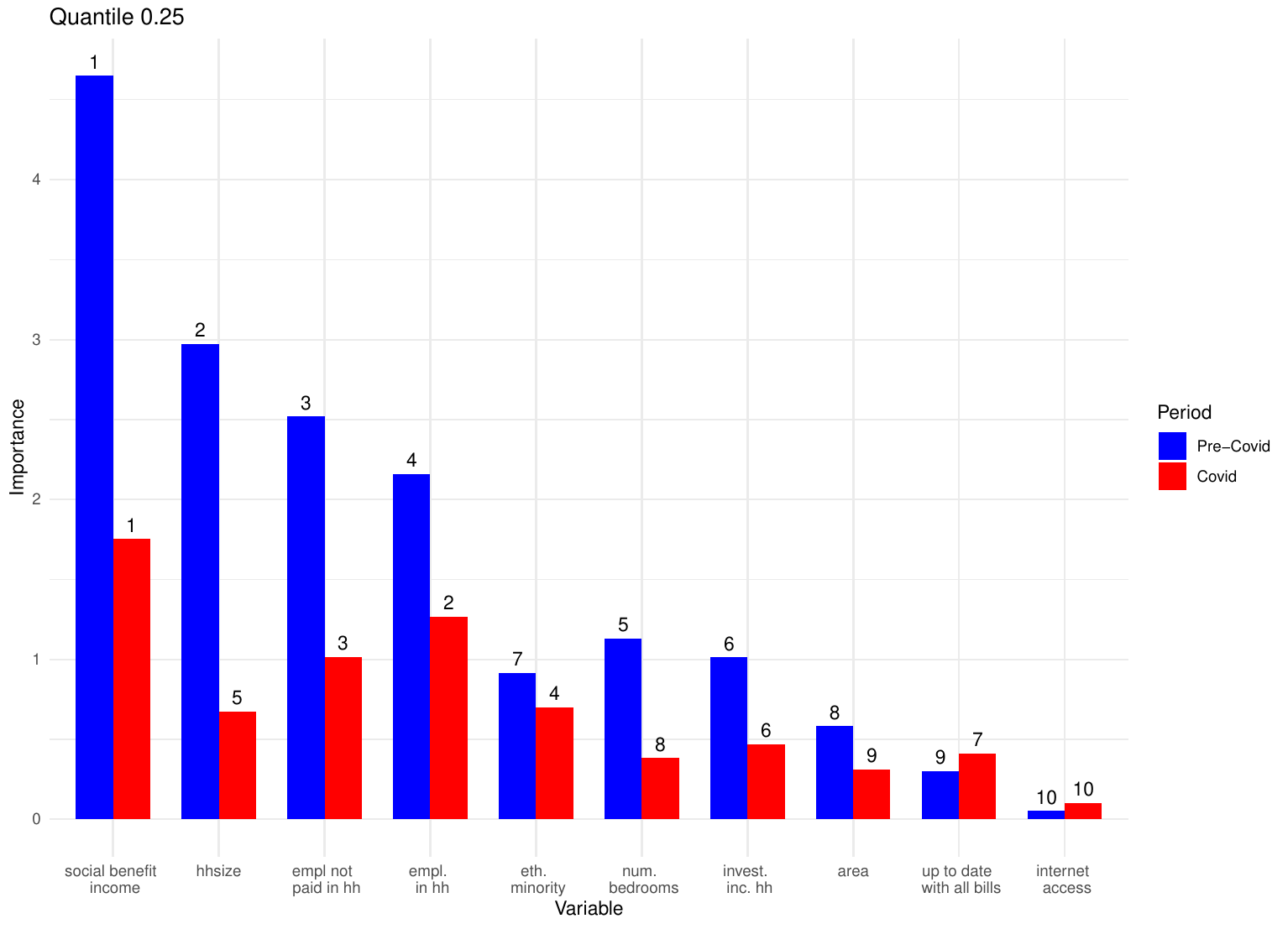}
    \caption{Bar plot showing the Variable Importance extracted from the FM-QRF for each covariate at quantile level $\tau=0.25$. }
    \label{fig:quantile025}
\end{figure}

\begin{figure}[h]
    \centering
\includegraphics[width=0.93\textwidth]{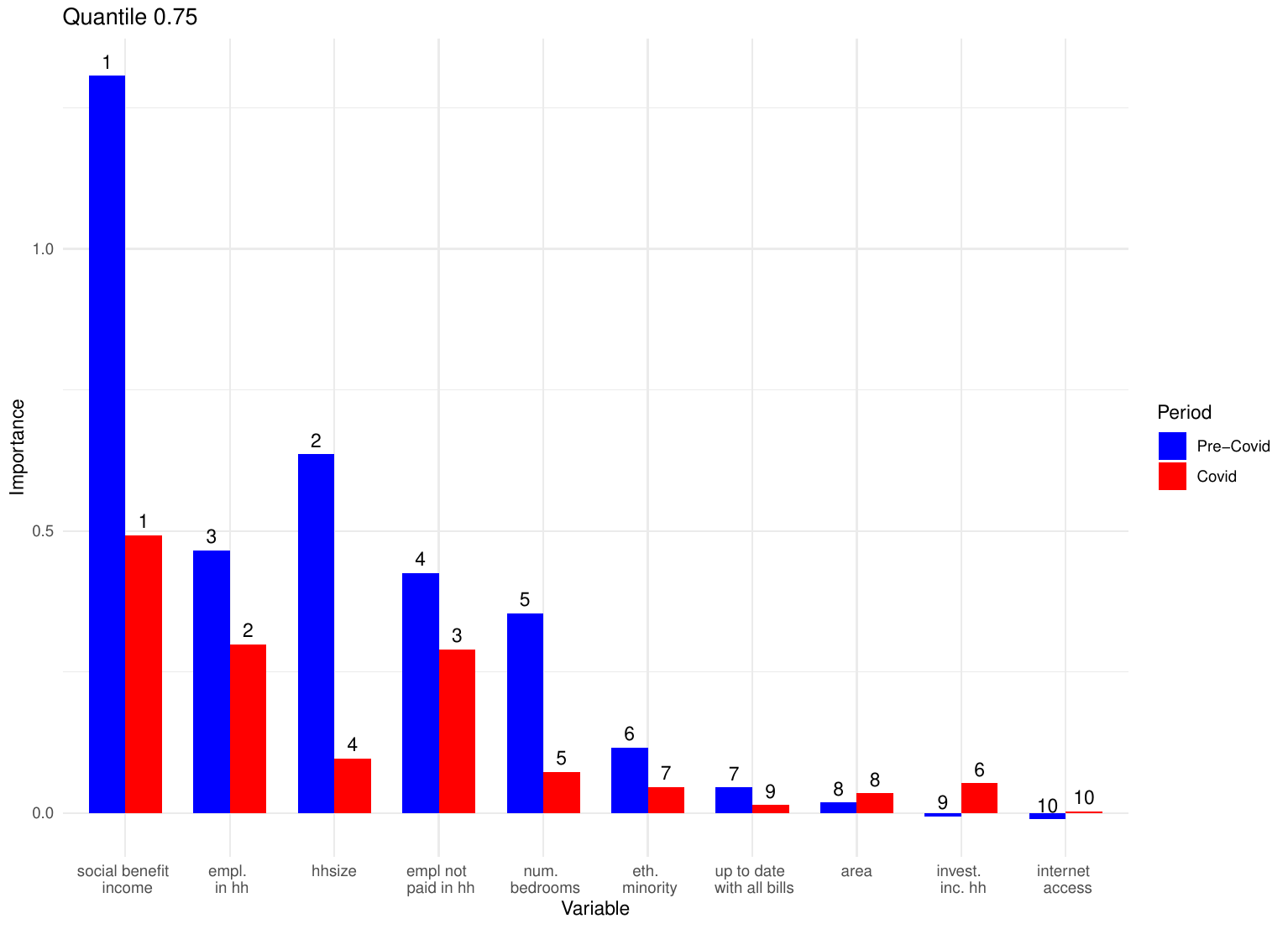}
    \caption{Bar plot showing the Variable Importance extracted from the FM-QRF for each covariate at quantile level $\tau=0.75$.}
    \label{fig:quantile075}
\end{figure}

\end{document}